\documentclass[letterpaper,twocolumn,fleqn]{article} 

\usepackage{ist}
\usepackage{hyperref}
\usepackage{amsmath}
\usepackage{amsfonts}
\usepackage{subfig}

\providecommand{\tbpic}[1]{\begin{minipage}{.18\textwidth}
\includegraphics[width=\linewidth]{#1}\end{minipage}}

\title{Interactive Illumination Invariance}

\author{Han Gong, Graham Finlayson\\University of East Anglia, UK}

\date{} 

\begin{document} 

\maketitle 

\thispagestyle{empty} 


\begin{abstract}
Illumination effects cause problems for many computer vision algorithms. We present a user-friendly interactive system for robust illumination-invariant image generation. Compared with the previous automated illumination-invariant image derivation approaches, our system enables users to specify a particular kind of illumination variation for removal. The derivation of illumination-invariant image is guided by the user input. The input is a stroke that defines an area covering a set of pixels whose intensities are influenced predominately by the illumination variation. This additional flexibility enhances the robustness for processing non-linearly rendered images and the images of the scenes where their illumination variations are difficult to estimate automatically.
Finally, we present some evaluation results of our method.
\end{abstract}

\section{Introduction}
\label{sec:intro}
Illumination conditions affect many computer vision algorithms from fundamental level to high level. For instance, unwanted shadows can cause artefacts in image segmentation, tracking, and object recognition. This is due to the difficulty in distinguishing the different image edges caused by illumination discontinuity, surface material~\cite{gershon1986ambient}, occlusion~\cite{sundberg2011occlusion}, and object boundary~\cite{dollar2006supervised}. An illumination-invariant image is therefore in great demand in both computer vision and computer graphics communities.

A shadow is an area where light from a light source is blocked by an obstacle. This is perhaps the most important phenomenon caused by illumination. There has been much previous work~\cite{finlayson2006removal,finlayson2009entropy,guo2013paired,su2010three,gong2013user,gong2014interactive} focusing on removing shadows from single images. These approaches aim at restoring an image with predominant shadows removed. In this restored image, most surface texture information is intact. The minor shadows are usually kept and they are regarded as the surface textures. The diffuse (extremely soft) shadows are also not removed because these approaches rely on image segmentation or image edge detection but the soft shadow boundaries are not easily detectable. Even when the soft shadow edges are detected, it is still very difficult to restore the penumbra information as the surface textures are the strong noise that affect illumination estimation. In fact, the recovery of penumbra is the most difficult part of this problem. Only a few of them~\cite{su2010three,gong2014interactive} can preserve the penumbra texture for a wide range of different shadows, such as broken shadows, coloured shadows, and soft shadows. The recovery process is also computation-costly that even some fast methods~\cite{su2010three,gong2014interactive} can still take several seconds to solve, by using an average machine.

Intrinsic image decomposition divides a single image into two components: a shading component and a reflectance component which is independent of illumination. The original single image is the product of these two components. These approaches~\cite{barron2012color,rother2011recovering,bousseau2009user} commonly make two assumptions: (1) Neighbouring pixels have the same reflectance if their chromaticities are similar; (2) Intensity discontinuities in the luminance of an image are caused by sharp reflectance changes and the illumination change is only smooth. The decomposed components are useful for image editing, such as re-colouring and surface material replacement. Most of them require a sliding window to process a local image block and an optimisation energy function to enforce the smoothness of decomposition results for different local image blocks. The optimisation process is usually very computational-costly as it requires the adjustment for every single pixel in each iteration. The resulting reflectance component can often contain visible illumination traces.

Illumination-invariant image computation is related to the intrinsic image derivation. An image is derived that depends only on reflectance. An important illuminant invariant image is derived in work~\cite{finlayson2001color} where it was shown that there existed a projection direction in a log RGB image that depended only on reflectance. Initially, the calibration direction was found using a calibration procedure but \cite{finlayson2004intrinsic} adopts an entropy minimisation procedure to find the best projection direction. The intuition here is that the same surface in a log RGB space (and assuming Planckian Illumination) will as the illumination changes produces values that fall on a line. Different surfaces induce different parallel line. The projection direction, orthogonal to the direction of variance, was shown to minimise entropy. Unfortunately, for the entropy minimisation strategy to work the input should be raw linear images. This is a serious limitation - one which we address here - because many of the applications of intrinsic images are for rendered, non-linear, images (e.g. you want to remove a shadow on a family photo). The entropy minimisation strategy also does not guarantee the best invariant direction when the predominant 2D projection direction is not caused by the prevailing illumination. The derivation process can also remove important texture information when the intensity variation caused by texture is similar to the estimated illumination variation. When the illumination invariant direction is known, these algorithms are very fast (usually only take some milliseconds). Although the invariant methods \cite{finlayson2004intrinsic} and \cite{finlayson2006removal} at a first step produce a single linear combination of log chromaticity values that is invariant to illumination, this can be converted into a pseudo colour chromaticity image~\cite{drew2003recovery}.

This paper is concerned with a novel interactive approach for illumination-invariant image derivation. Instead of finding the invariant direction by either camera calibration~\cite{finlayson2006removal} or entropy minimisation~\cite{finlayson2004intrinsic}, we require the user to supply some easy hints (usually only one single stroke) to guide the derivation of the illumination variation direction. The required user input is a simple stroke that indicates an area covering a set of pixels whose intensity variations are predominately caused by a variation in illumination. This additional user task is easy and quick to perform. It also makes the illumination-invariant derivation steerable for the best result. Our method works in a very simple way: Given a segment of an image defined by a single input stroke, the pixels of that segment are divided into two groups according to the features of their intensities and coordinates. To get this illumination invariant direction, all the RGB intensity values of the image are first converted to 2D log geometric mean chromaticites by projecting their 3D log geometric mean chromaticities onto a plane~\cite{finlayson2004intrinsic}. It is assumed that the intensity variation of the pixels in this segment is mainly caused by the illumination. On this 2D chromaticty plane, the intensity difference between these two groups thus represents the illumination direction. As the given rough segment can contain noise and outliers (usually caused by surface texture or surface material change), some outlier amendment steps are also applied to resolve this issue. To get the 1D illumination-invariant image, our method projects the 2D chromaticities onto a line orthogonal to the illumination invariant direction vector. The 2D illumination-invariant image can be obtained by applying a 1D-to-3D projection to the projected 1D illumination-invariant values.

This paper is organised as follows. Firstly, we describe our user interaction design and the pre-processing step for user input. Secondly, we describe our approach for illumination-invariant image generation guided by user input. Finally, some evaluation results of our interactive illumination-invariant image are shown.

\section{User Interaction And Pre-Processing}
\label{sec:ui}
In this section, we describe our user interaction design and its algorithm.

\subsection{User Interaction}
\begin{figure*}[ht!]
  \centering
  \includegraphics[width=\linewidth]{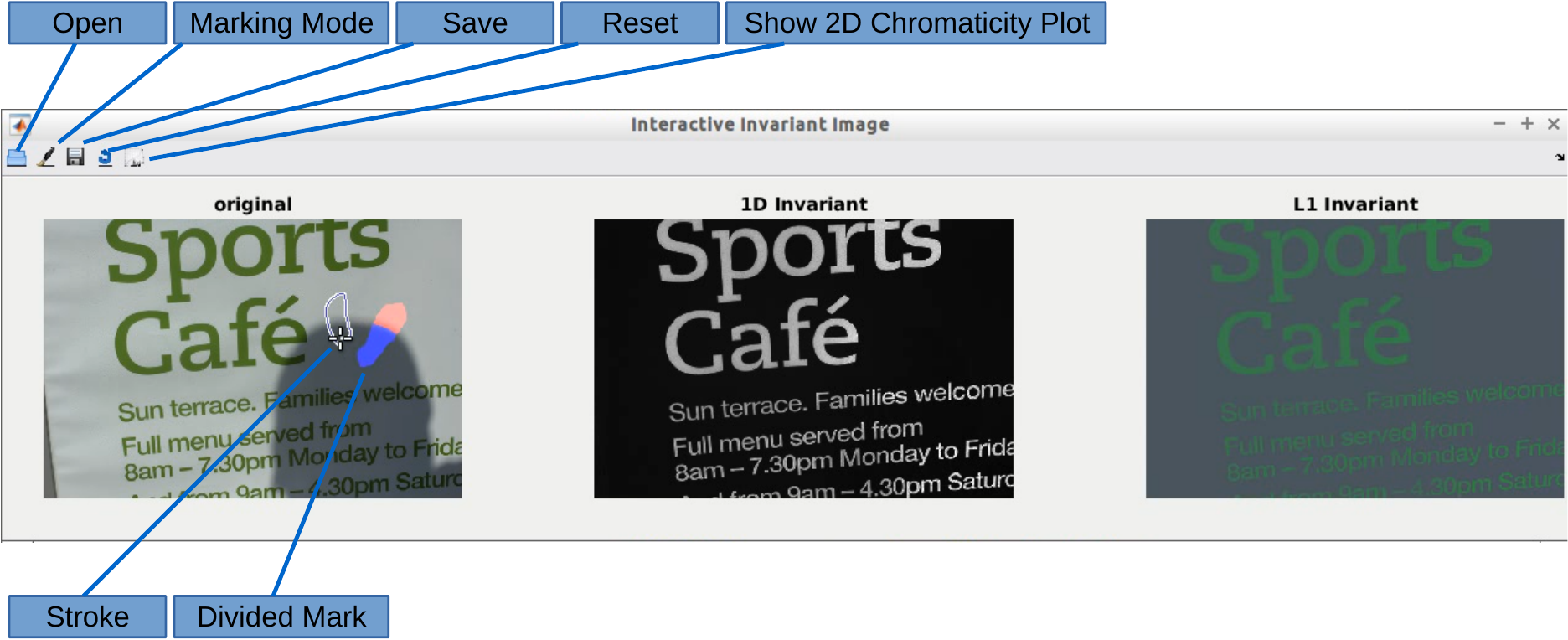}
  \caption{Graphical user interface: (1) open -- open an image file; (2) marking mode -- choose either to draw or to erase marks; (3) save -- save the illumination-invariant images; (4) reset -- clear all marks; (5) stroke -- a stroke enclosing an area of mark; (6) divided mark -- the mark being divided into a lit part (in red) and a shadow part (in blue). The three images shown in the interface are: left -- original RGB image; middle -- 1D illumination-invariant image; right -- L1 illumination-invariant log chromaticity image. The input image is taken from a shadow removal dataset~\cite{gong2014interactive}.}
  \label{fig:ui}
  \vspace{10pt}
\end{figure*}

Fig.~\ref{fig:ui} shows the prototype of our user interface and the description of its controls.
The required mark is a segment in a shape that is similar to an ellipse, e.g., the divided mark in blue and red in Fig.~\ref{fig:ui}. The direction of its semi-major axis indicates the potentially strongest intensity gradient change direction. This requirement is easy to fulfil and it gives an additional cue to make our user input pre-processing more resistant to strong surface textures. Our system also allows users to supply multiple marks to improve the result. The proposed drawing actions have two types:\newline
\textbf{Mark Addition}
Mark addition is equivalent to adding or sometimes merging a new part to the previous marks. This is done by applying a logical \emph{or} image operation as follows:
\begin{equation}
    \label{equ:addshape}
    N_{n} = N \vee N_{a}
\end{equation}
where $N$ refers to the current mask of mark, $N_{a}$ is the additional mask of mark drawn by user, $N_{n}$ is the updated mask of mark.\newline
\textbf{Mark Subtraction}
Mark subtraction is similar to a eraser which is used to remove the existing marks. Similar to Eq.~\ref{equ:addshape}, mark subtraction is done by applying two logical image operations as follows:
\begin{equation}
    N_{n} = N \wedge \neg N_{a}
\end{equation}

\subsection{Pre-Processing}
Given the marked pixels of an image, our pre-processing step divides them into two groups. The marked pixels' RGB intensities and pixel coordinates are used as the features for division. For each marked illuminant change, we find the RGB that is in the shadow and the one outside the shadow by a K-Means++ clustering~\cite{arthur2007k}. The marked pixels' 2D pixel coordinates are not directly used as the features. Instead, they are processed using PCA~\cite{wold1987principal} and only their scores of the first principle component are used as the feature of location. We further normalise the scores of the first principle such that their values are in the range $[0, 1]$. The intuition is that the vector of the first principle dimension is close to the major semi-axis of an ellipse. The direction of this semi-axis indicates the direction of the strongest illumination change. The RGB intensities are also normalised to the range $[0, 1]$. Therefore, our final feature space has 4 dimensions: three for RGB intensities and one for pixel location. Our main configuration for this K-Means++ clustering are: (1) Distance Measure: Squared Euclidean distance; (2) Maximum number of iteration: 100. The cluster with the lowest mean for its RGB intensity is considered as a shadow cluster and vice versa. For each cluster, we find the median RGB intensities.
\section{Illumination-Invariant Image}
\label{sec:main}
\begin{figure}[htb!]
  \centering
  \includegraphics[width=0.9\linewidth]{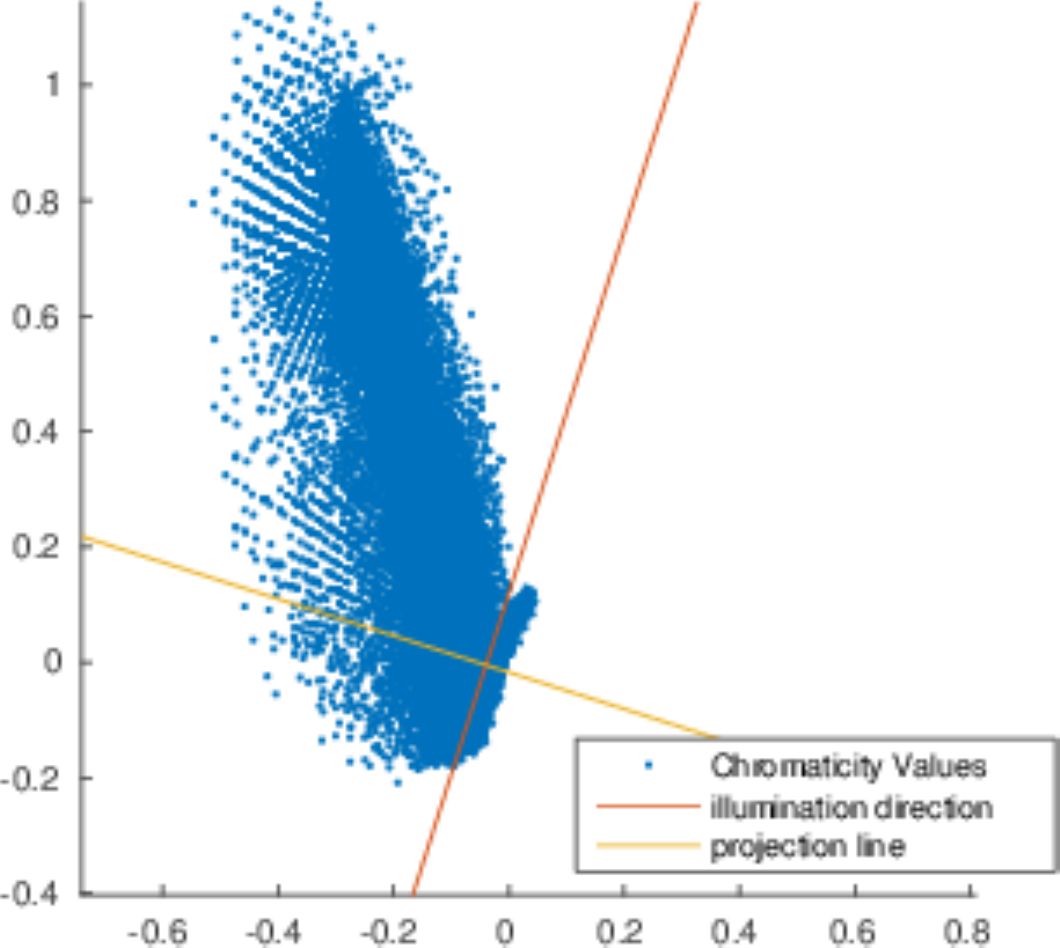}
  \caption{Example of illumination-invariant projection in a 2D log chromaticity space defined in \cite{finlayson2004intrinsic}. The log chromaticity points are projected on to the projection line (yellow). The data displayed in this figure corresponds to the image data in Fig.~\ref{fig:ui}.}
  \label{fig:2Dex}
  \vspace{10pt}
\end{figure}

Using the algorithm described in \cite{finlayson2004intrinsic}, the RGB intensities of the original image and the two single median RGB intensities (analysed from the user input) are first projected onto a 2D log chromaticity plane such that each pixel has two log chromaticity values. The user input provides an important cue for finding the accurate projection direction under such conditions. Given the converted 2D log chromaticity vectors $C^{L}$ (lit) and $C^{S}$ (shadow) for the two single median RGB intensities, the normalised illumination direction vector $P^{\bot}$ is
\begin{equation}
    \label{eq:illu_vec}
    P^{\bot} = \frac{C^{L}-C^{S}}{|C^{L}-C^{S}|}.
\end{equation}
The projection (i.e. illumination-invariant) vector $P$ is orthogonal to $P^{\bot}$ and can be computed by rotating $P^{\bot}$ by 90 degrees clockwise as the follows
\begin{equation}
    \label{eq:inv_vec}
    P = \begin{bmatrix}
0 & -1 \\
1 & 0 \\
\end{bmatrix}P^{\bot}.
\end{equation}
An example of this projection is shown in Fig.~\ref{fig:2Dex}.
Given a pixel's 2D log chromaticity vector and the illumination-invariant vector $P$, its 1D illumination-invariant value and 3D L1 log chromaticity value can be computed using the projection algorithm described in \cite{finlayson2004intrinsic}. 
An example of this illumination-invariant image is shown in Fig~\ref{fig:ui}.

\section{Evaluation}
\label{sec:eval}
In this section, we first show some typical examples with a result comparison highlighting the advantage of using user interaction. Finally, some failure cases are shown. 
\subsection{Results}
\begin{figure*}
    \resizebox{\linewidth}{!}{\begin{tabular}{|c|c|c|c|c|c|c|}
\hline
No. & Original & Our 1D Invariant & Our L1 Invariant & 1D Invariant by \cite{finlayson2004intrinsic} & L1 Invariant by \cite{finlayson2004intrinsic}\\ \hline
1&\tbpic{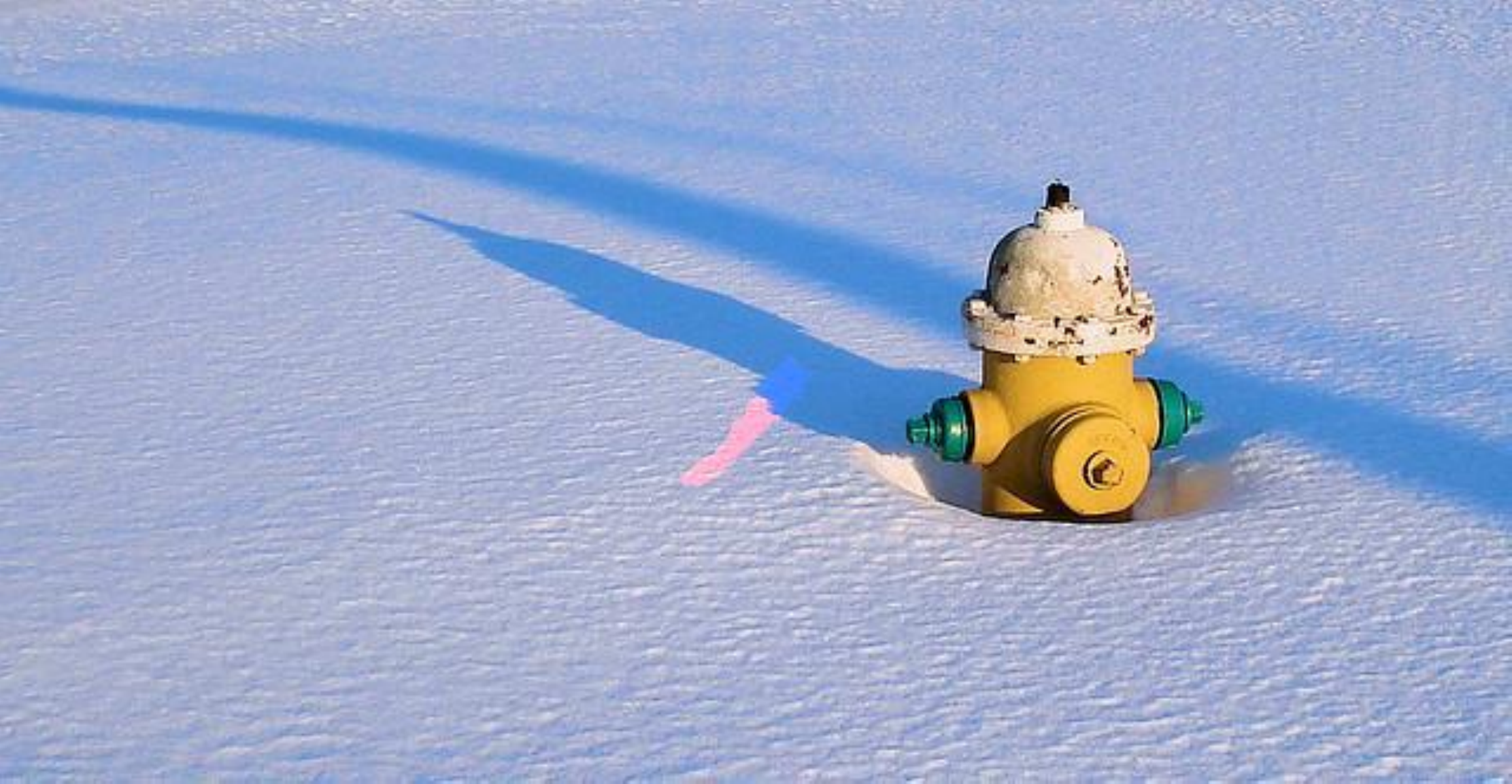}&\tbpic{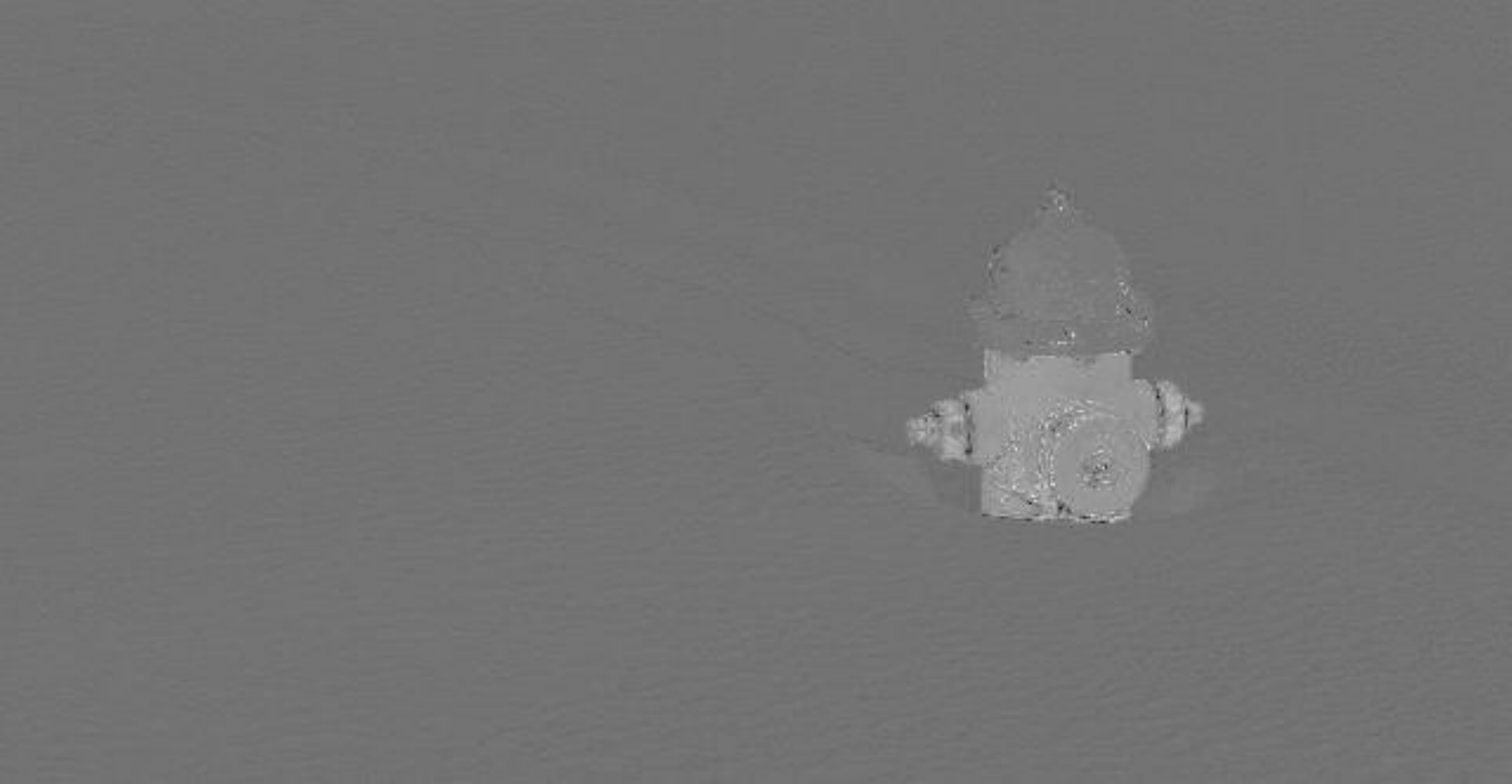}&\tbpic{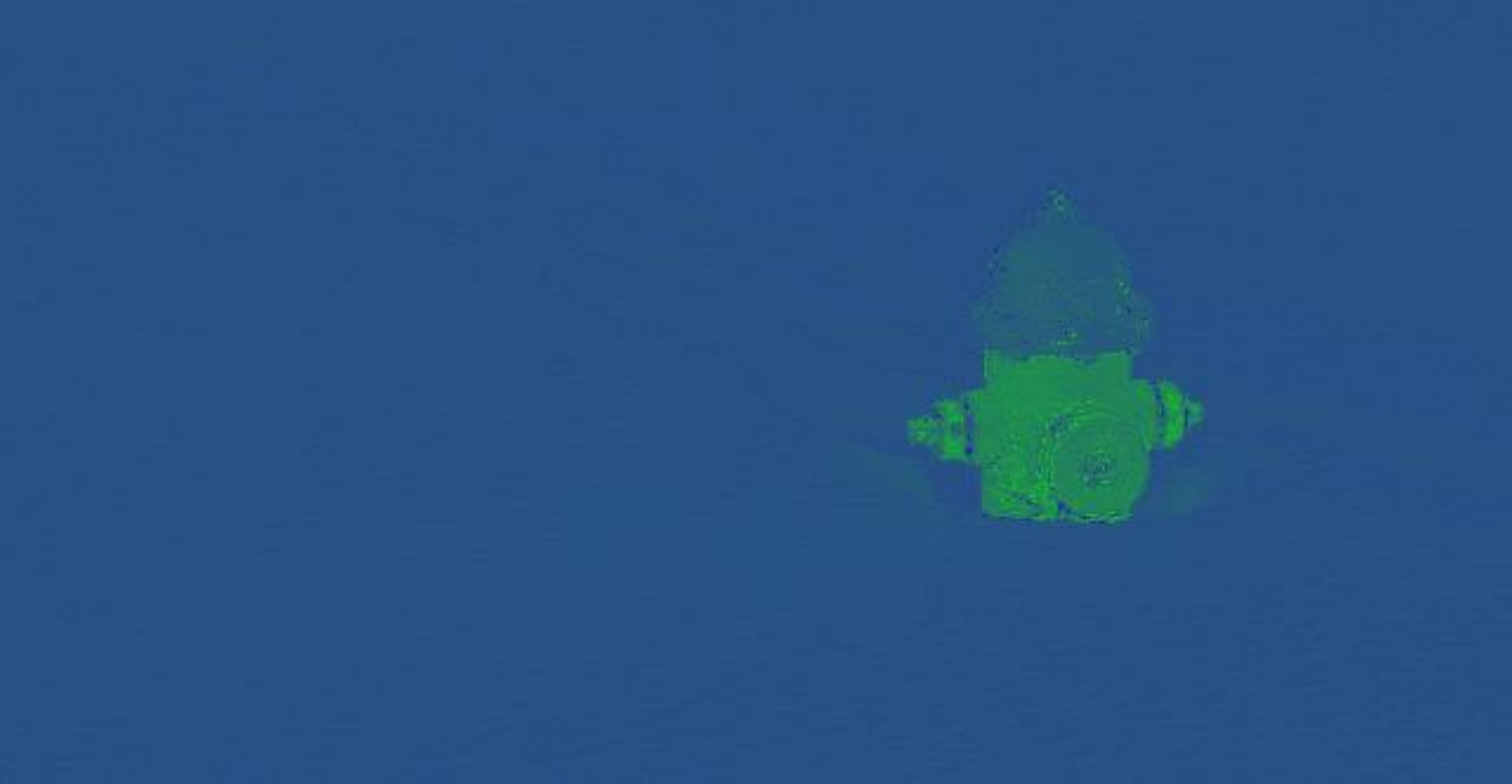}&\tbpic{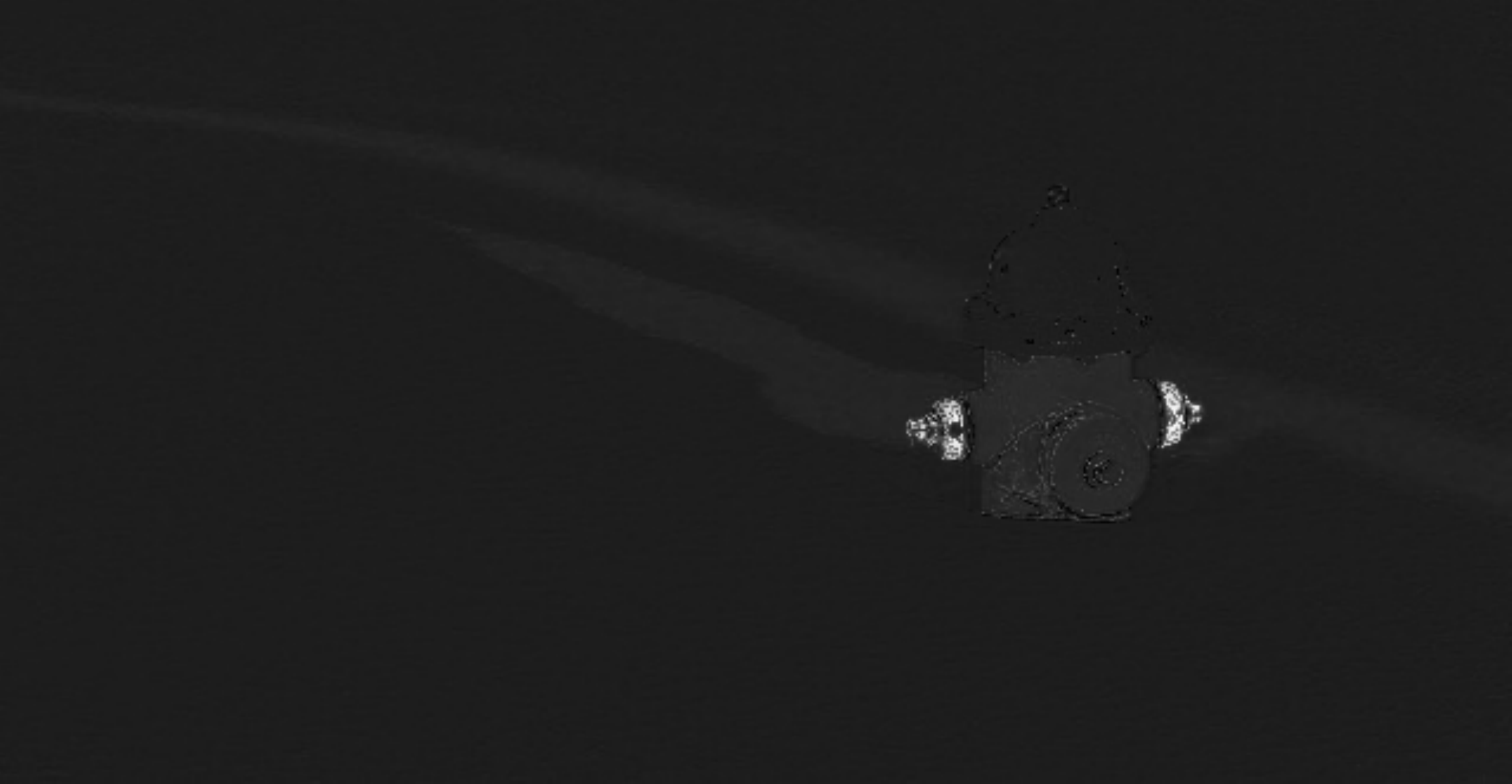}&\tbpic{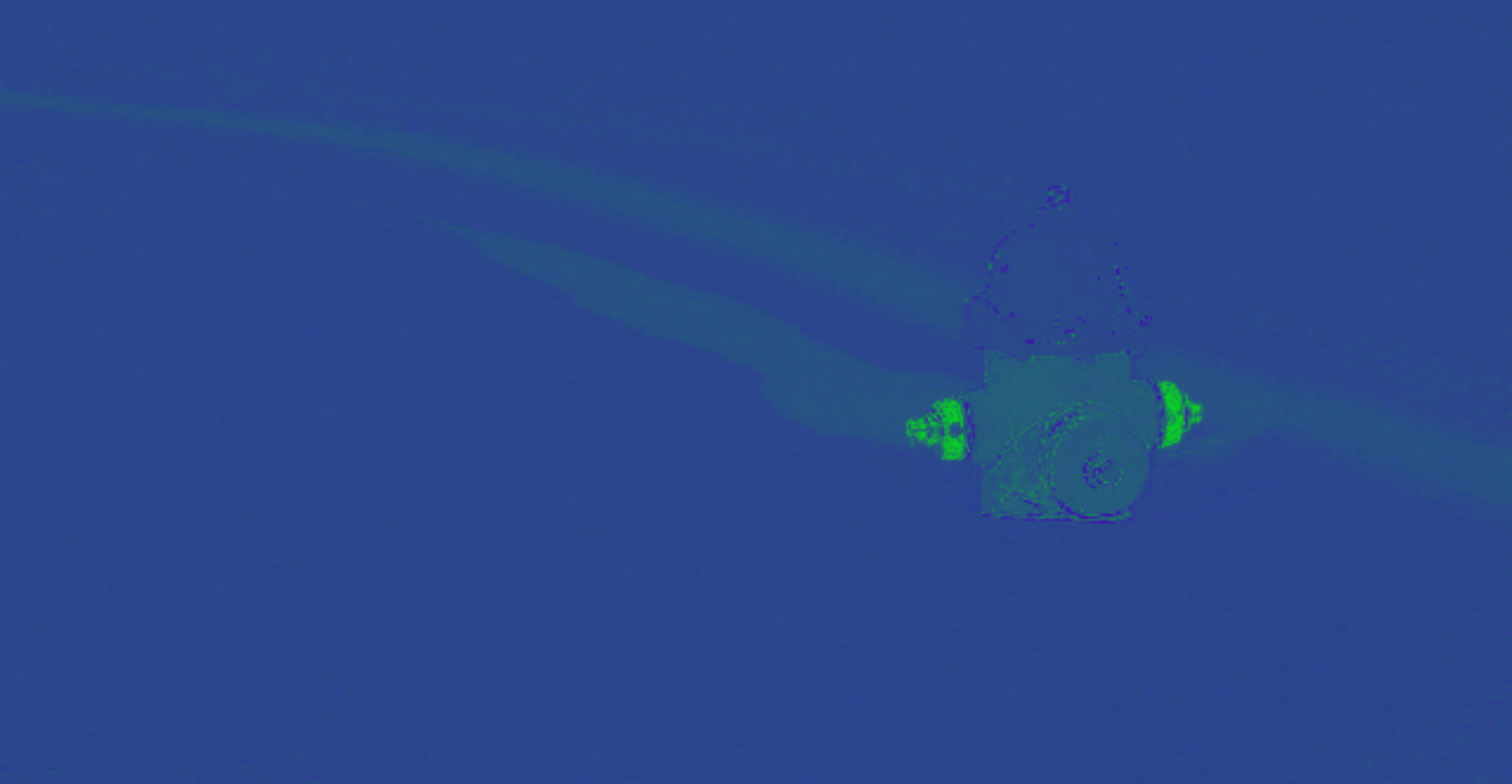}\\ \hline
2&\tbpic{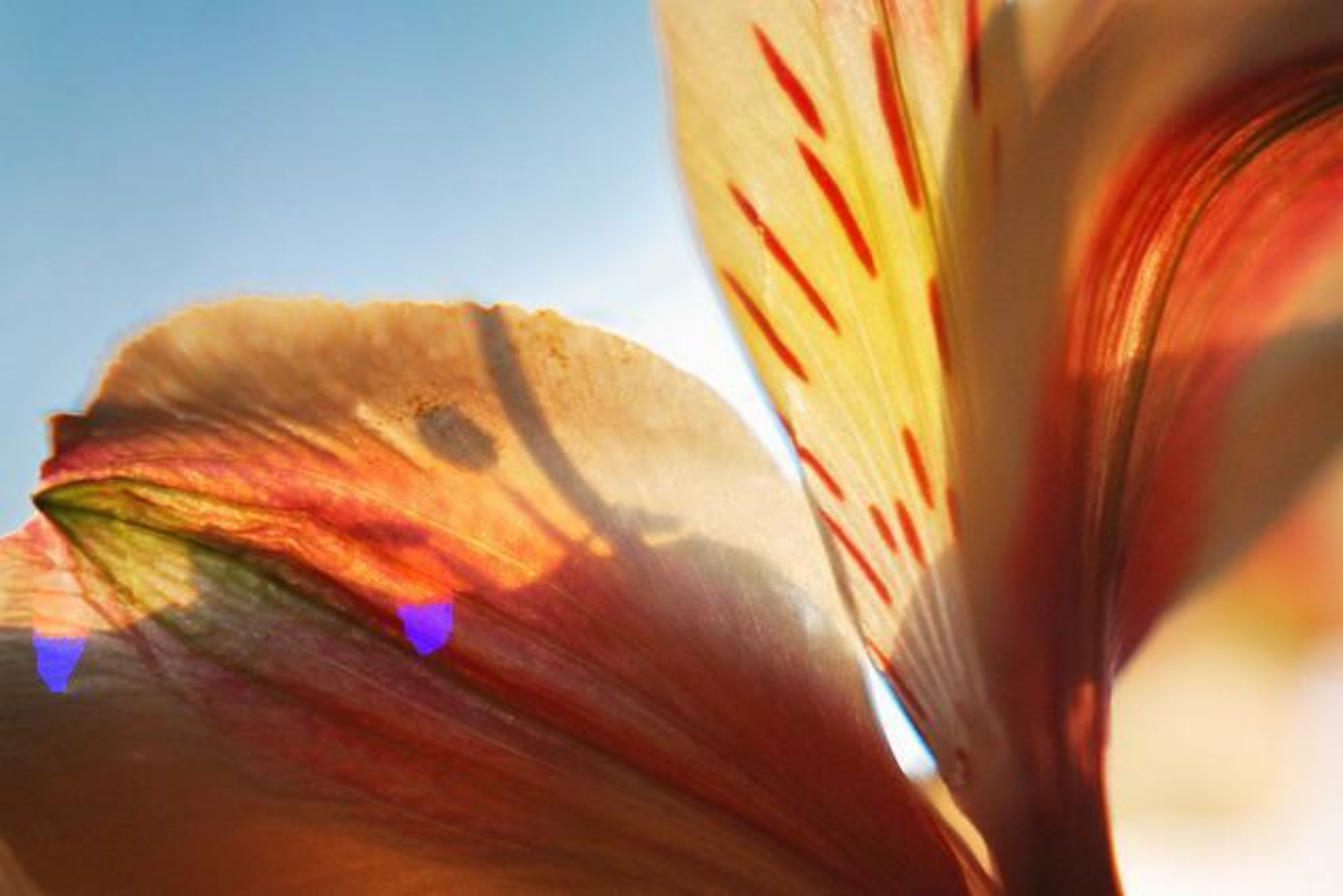}&\tbpic{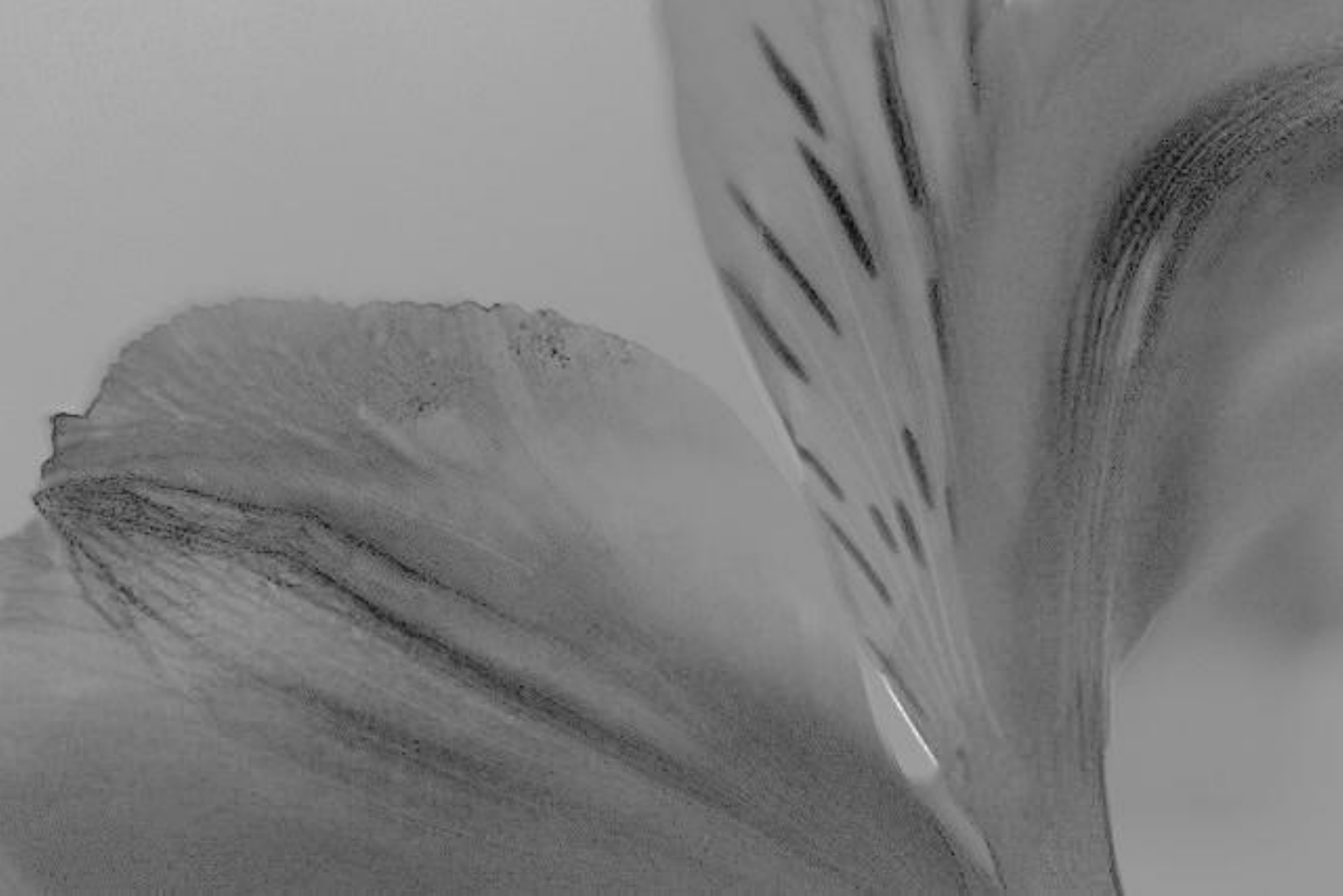}&\tbpic{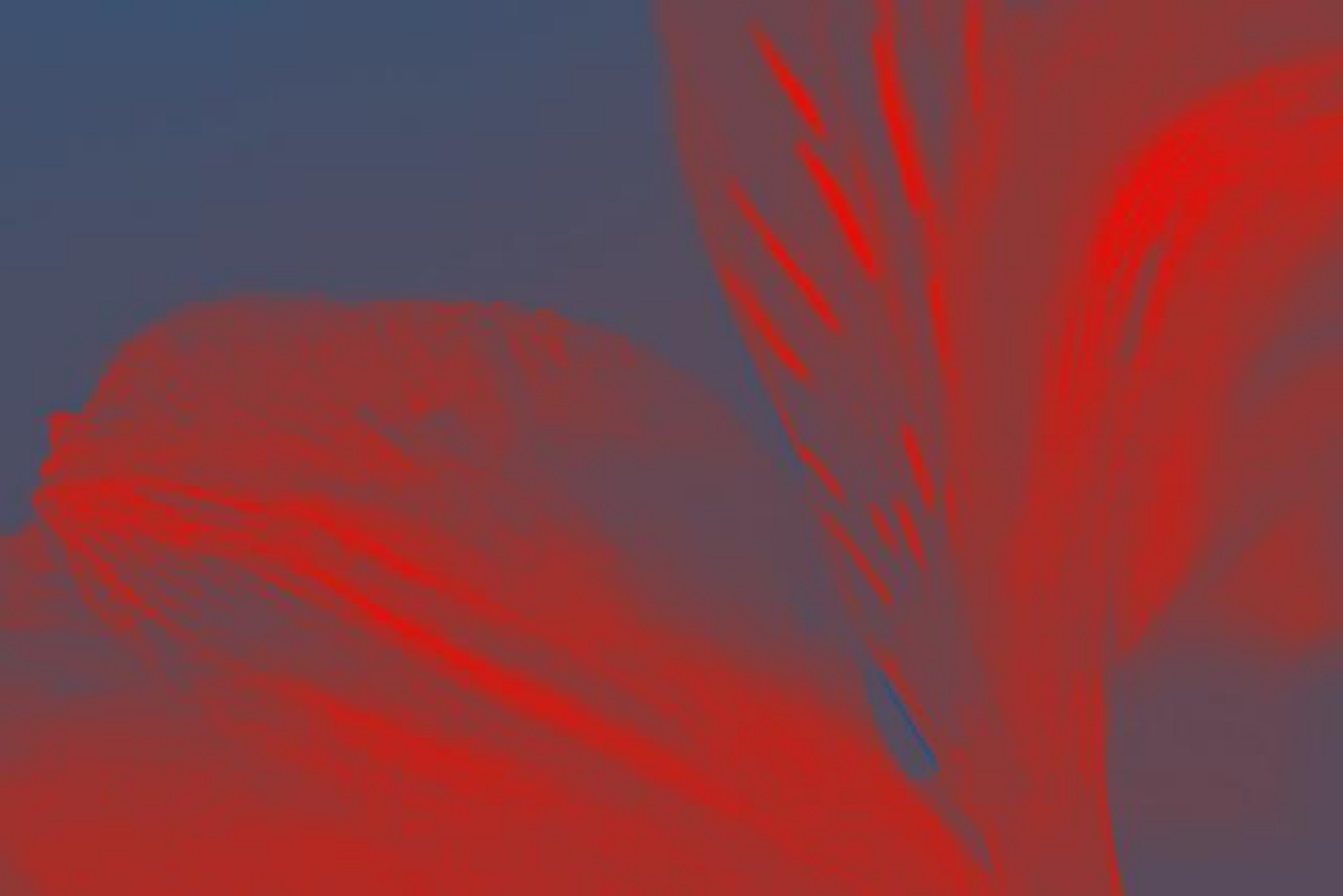}&\tbpic{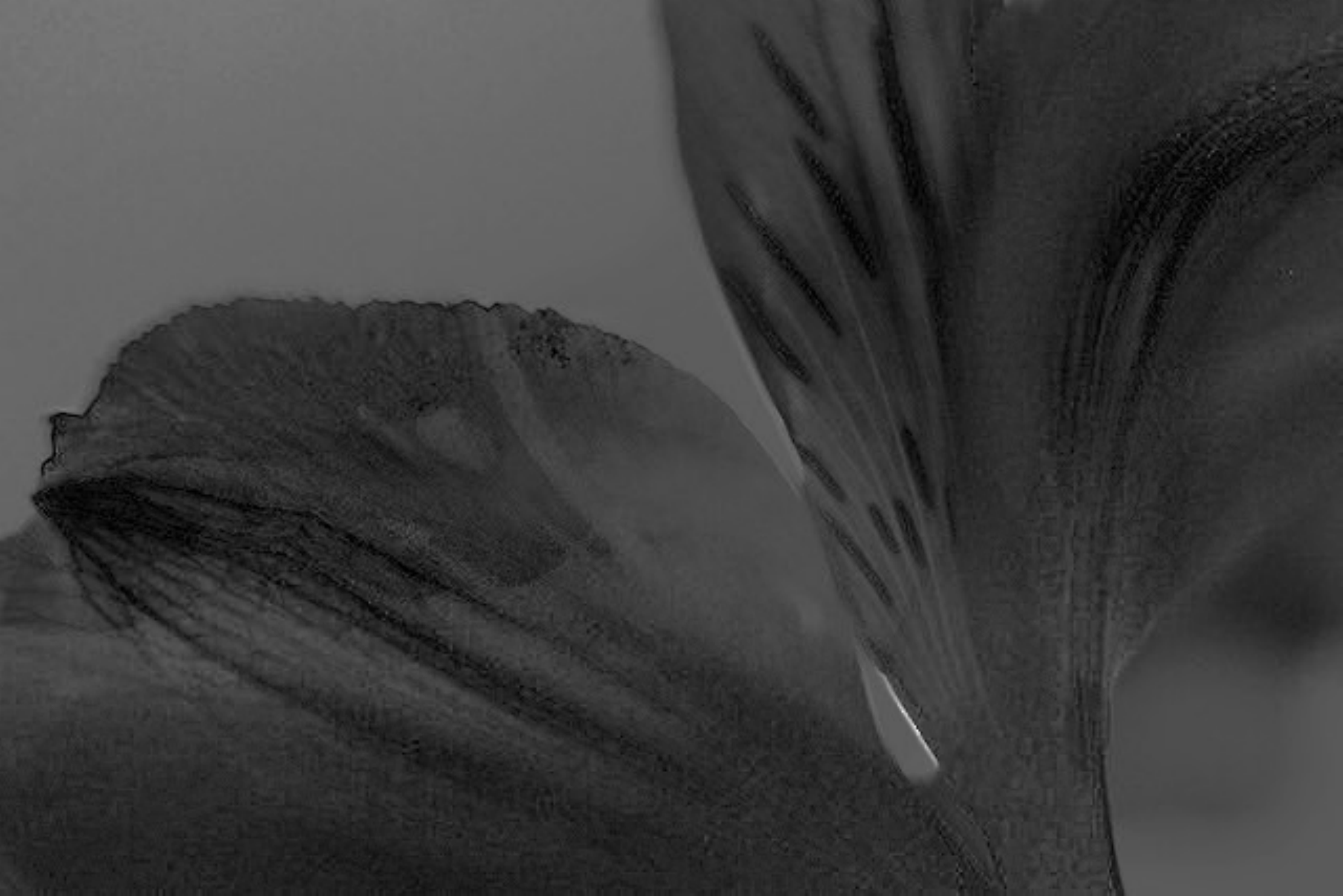}&\tbpic{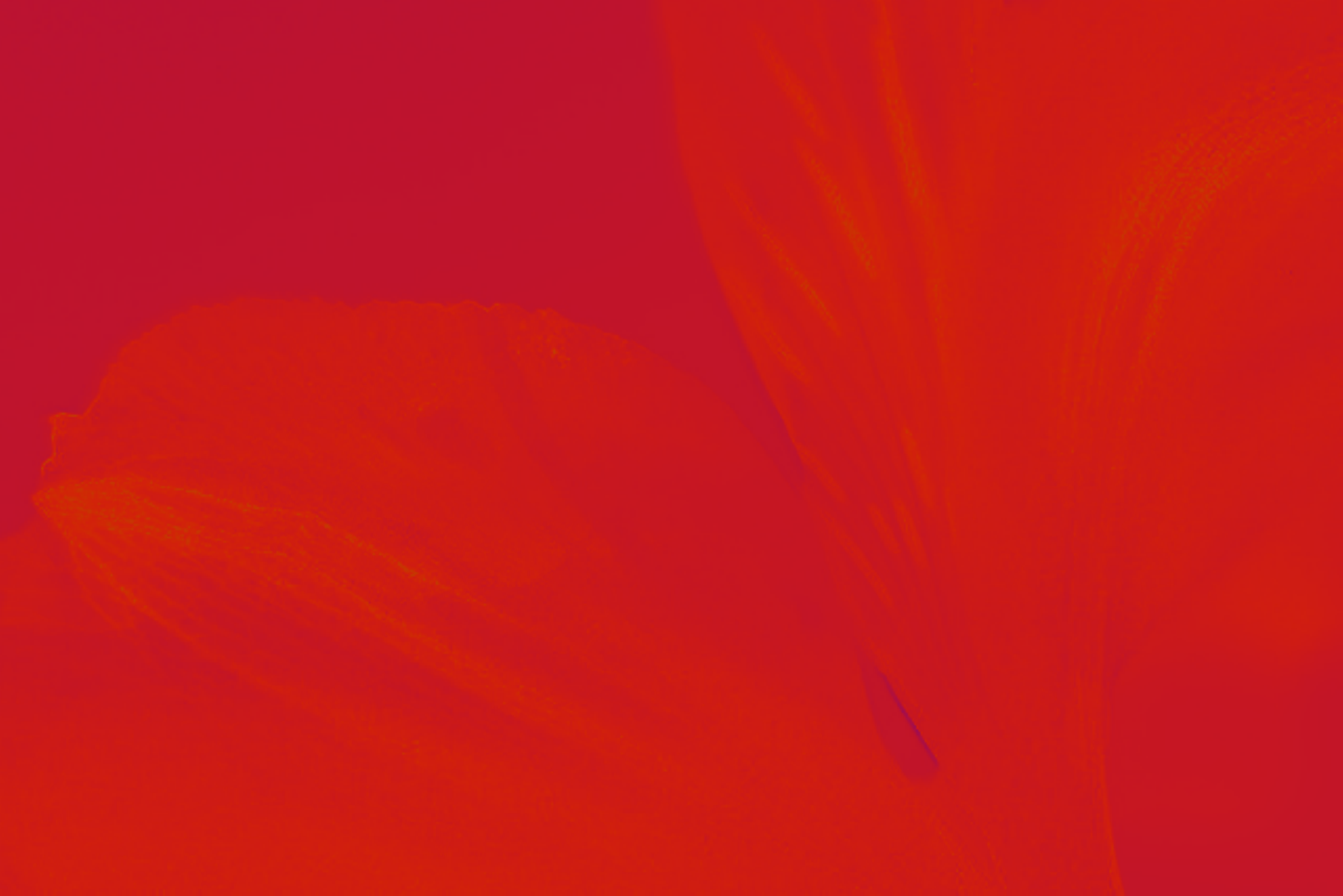}\\ \hline
3&\tbpic{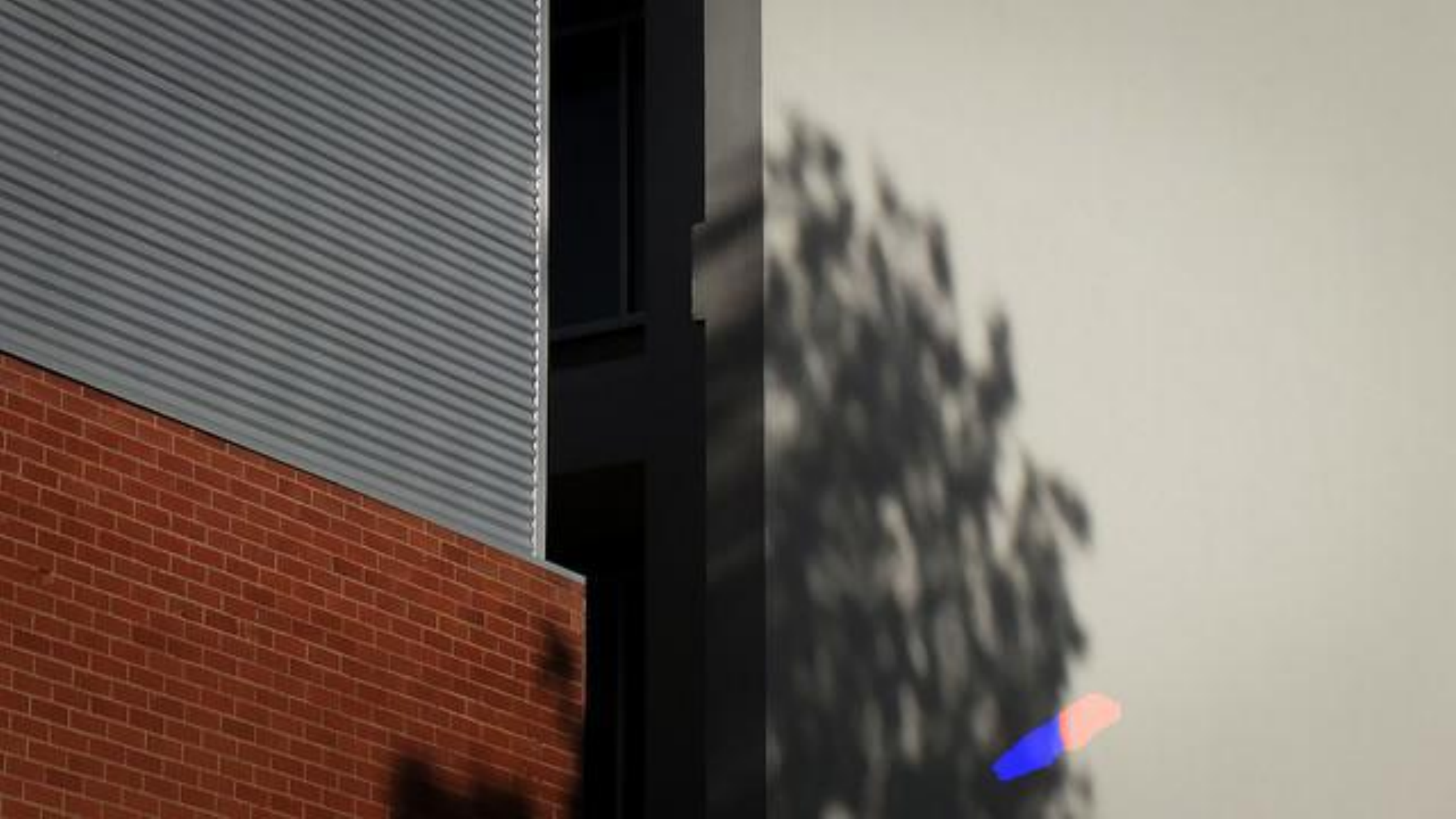}&\tbpic{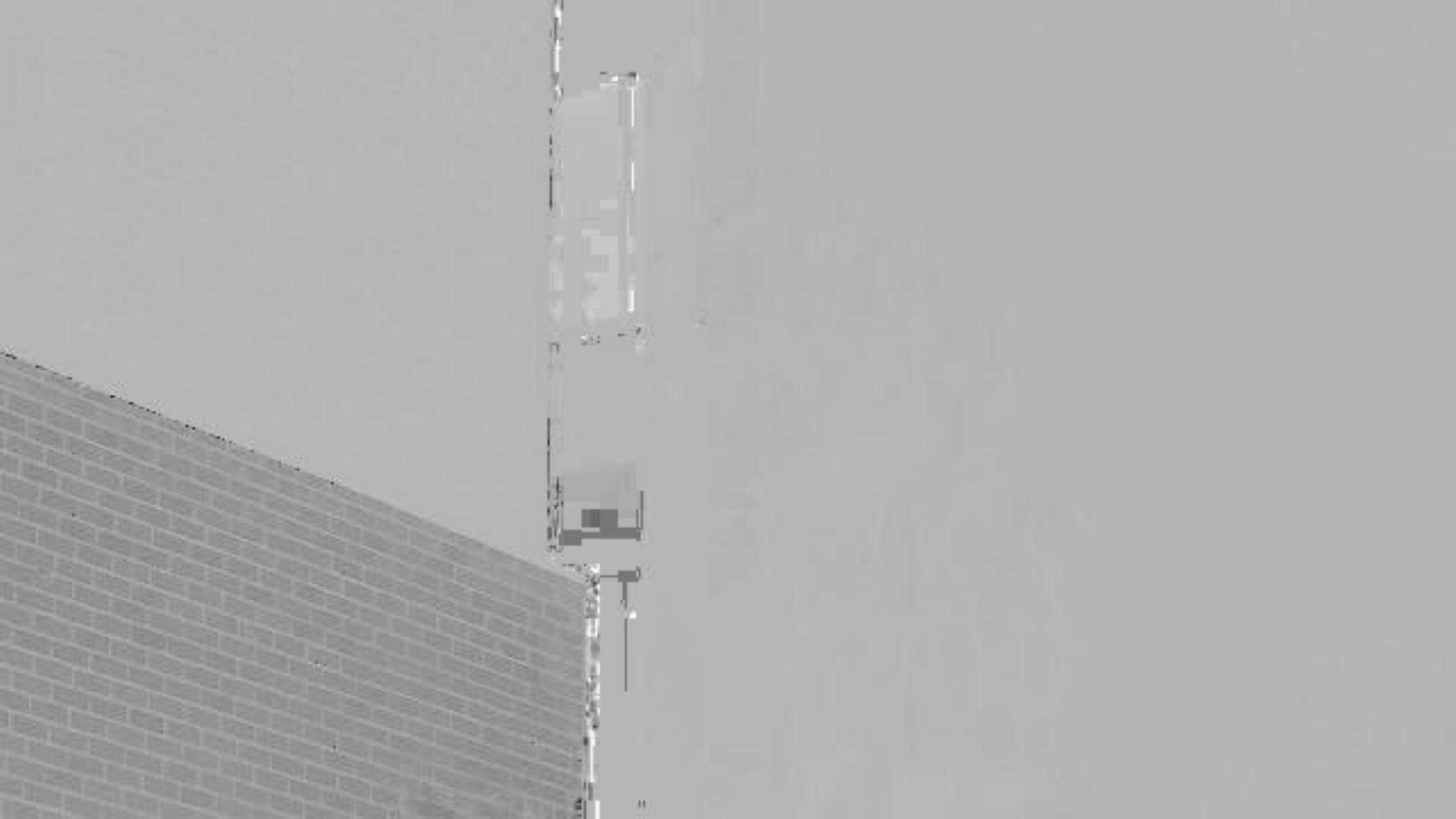}&\tbpic{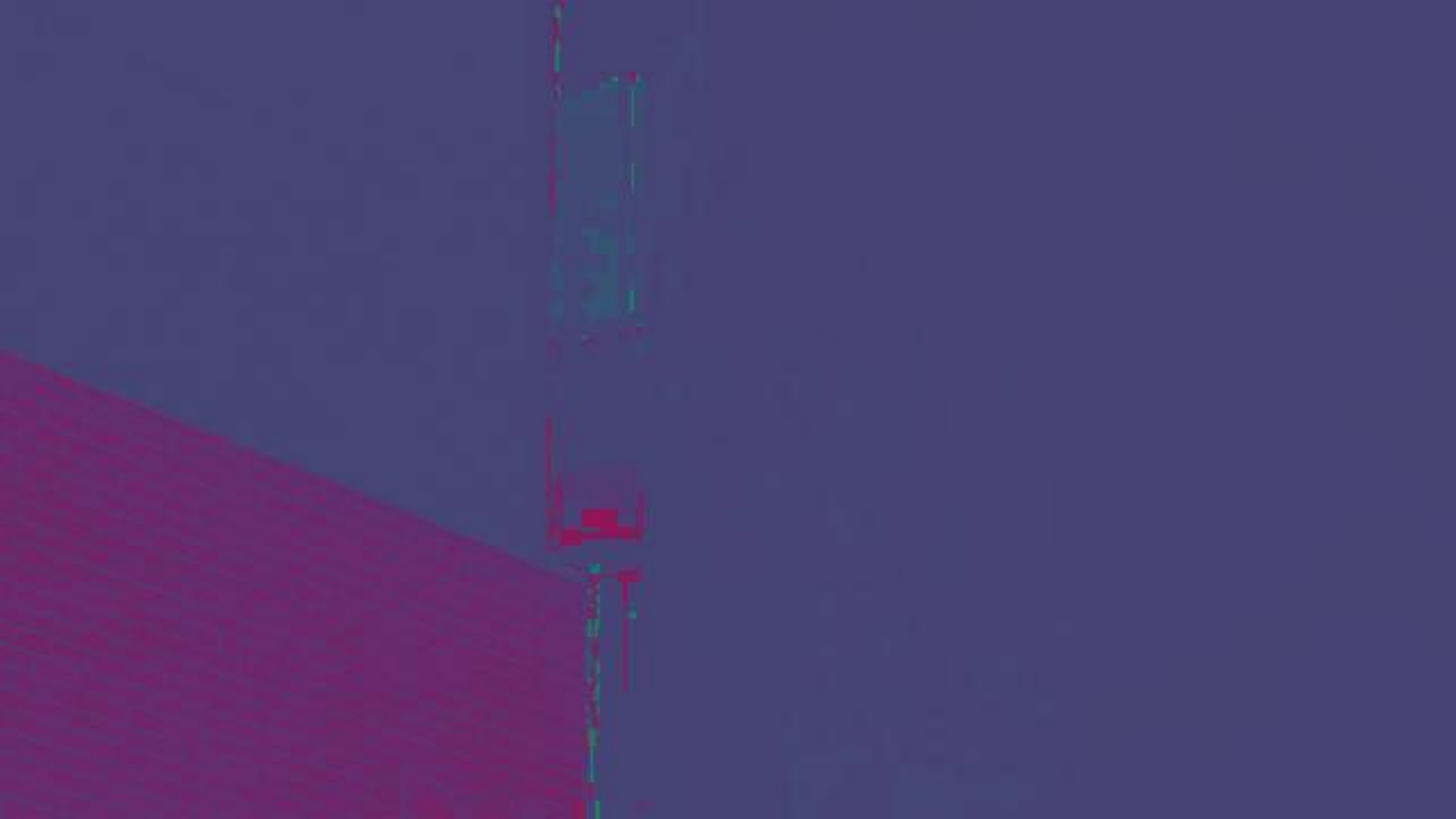}&\tbpic{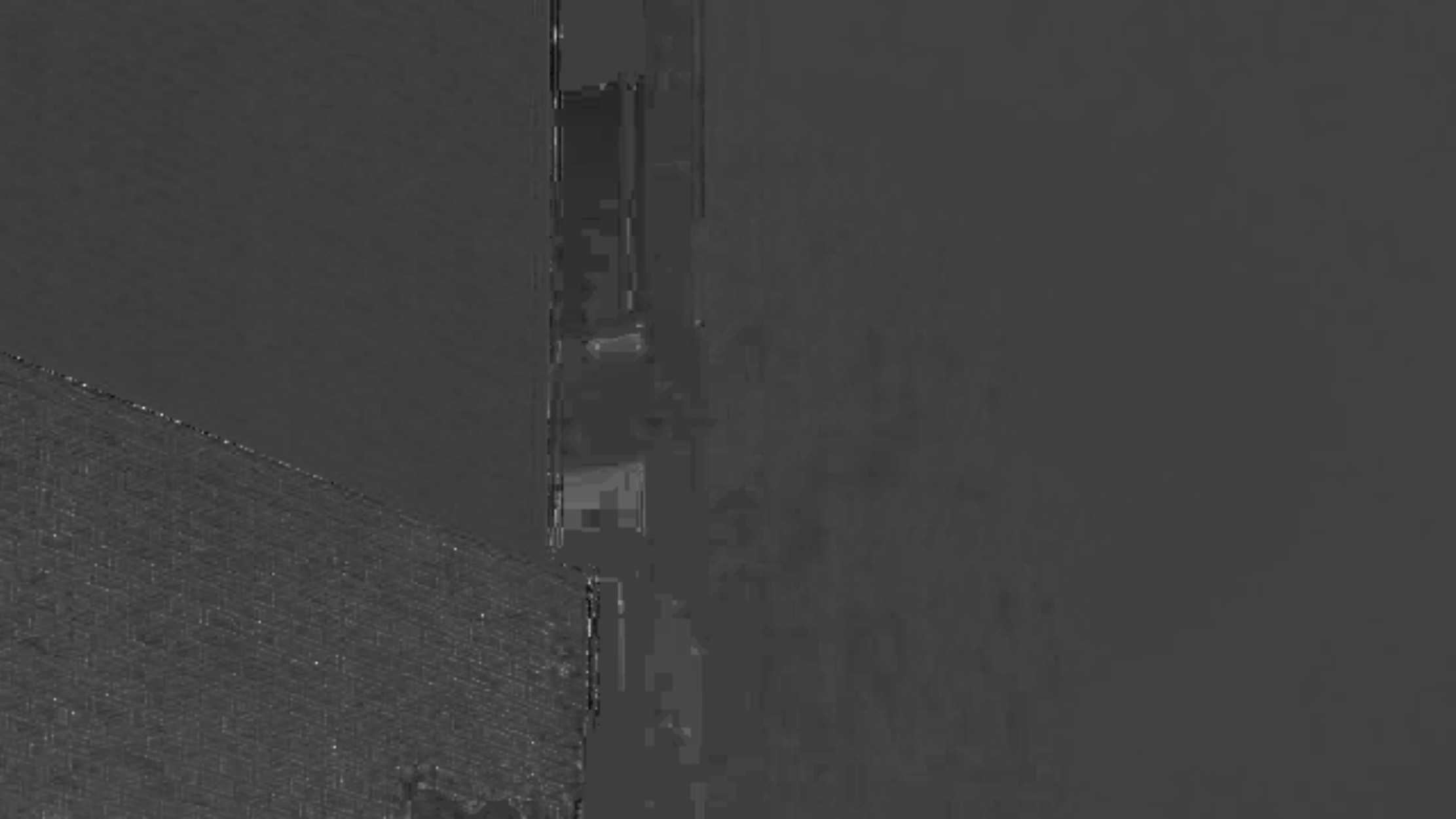}&\tbpic{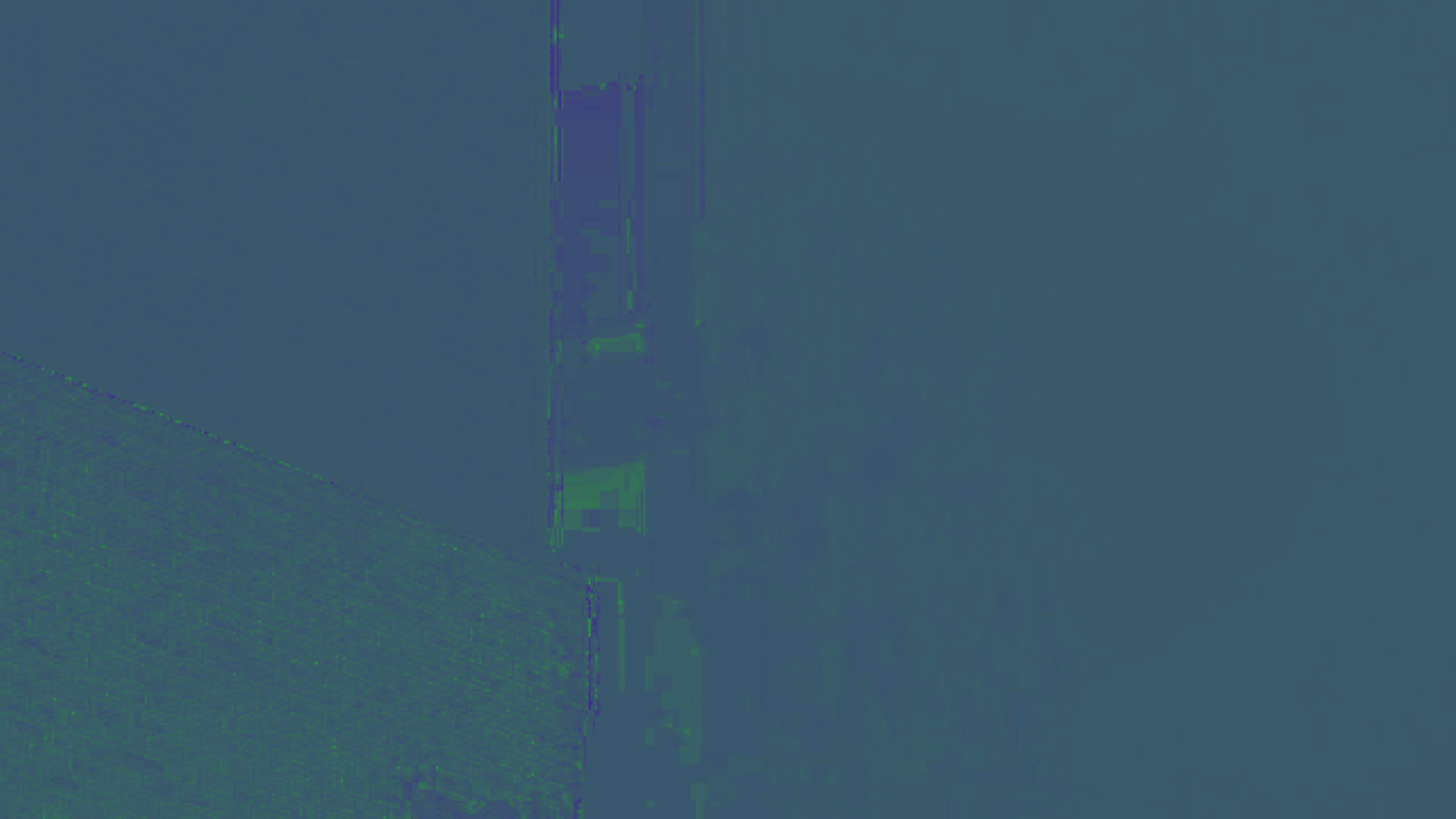}\\ \hline
4&\tbpic{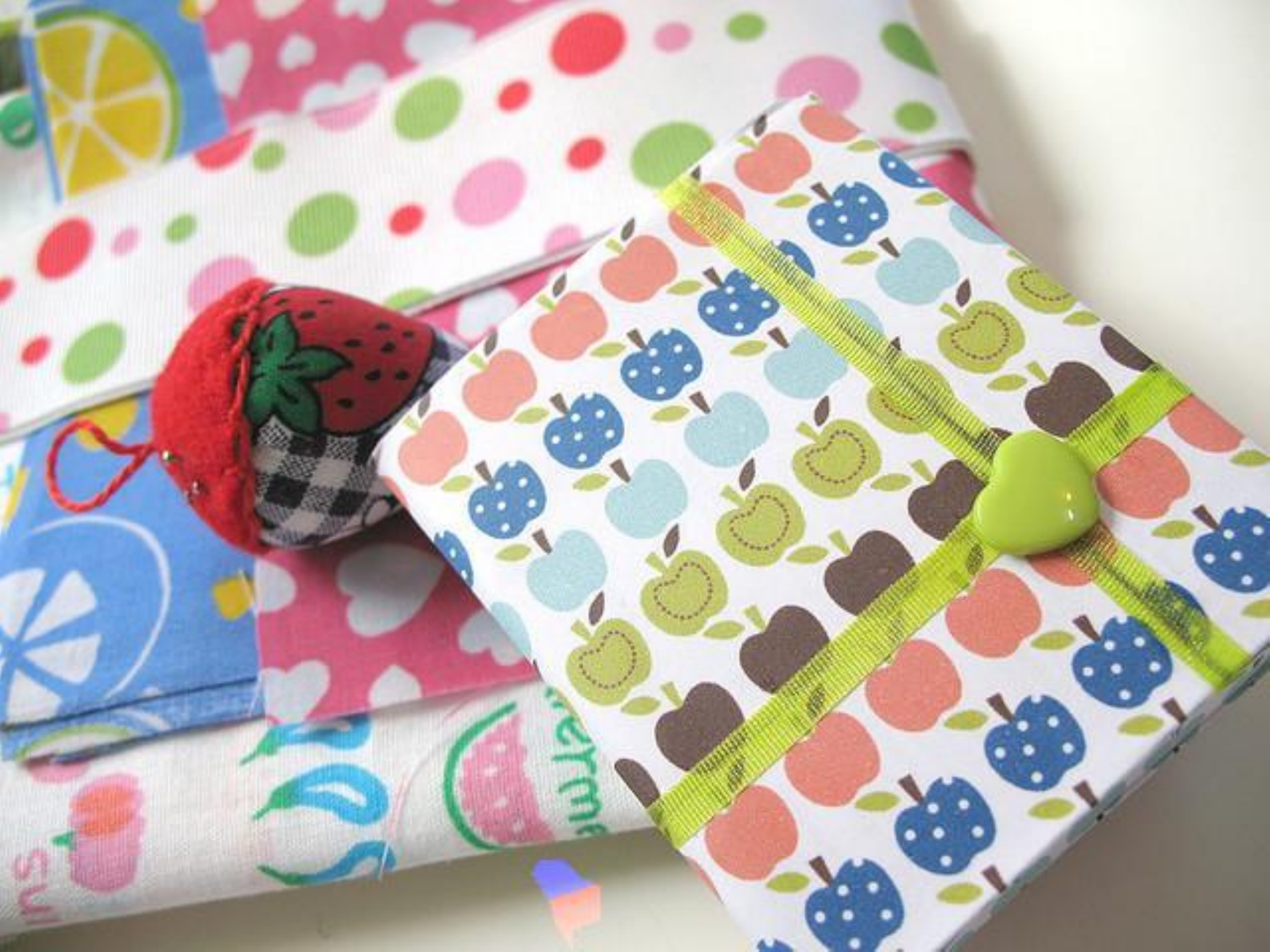}&\tbpic{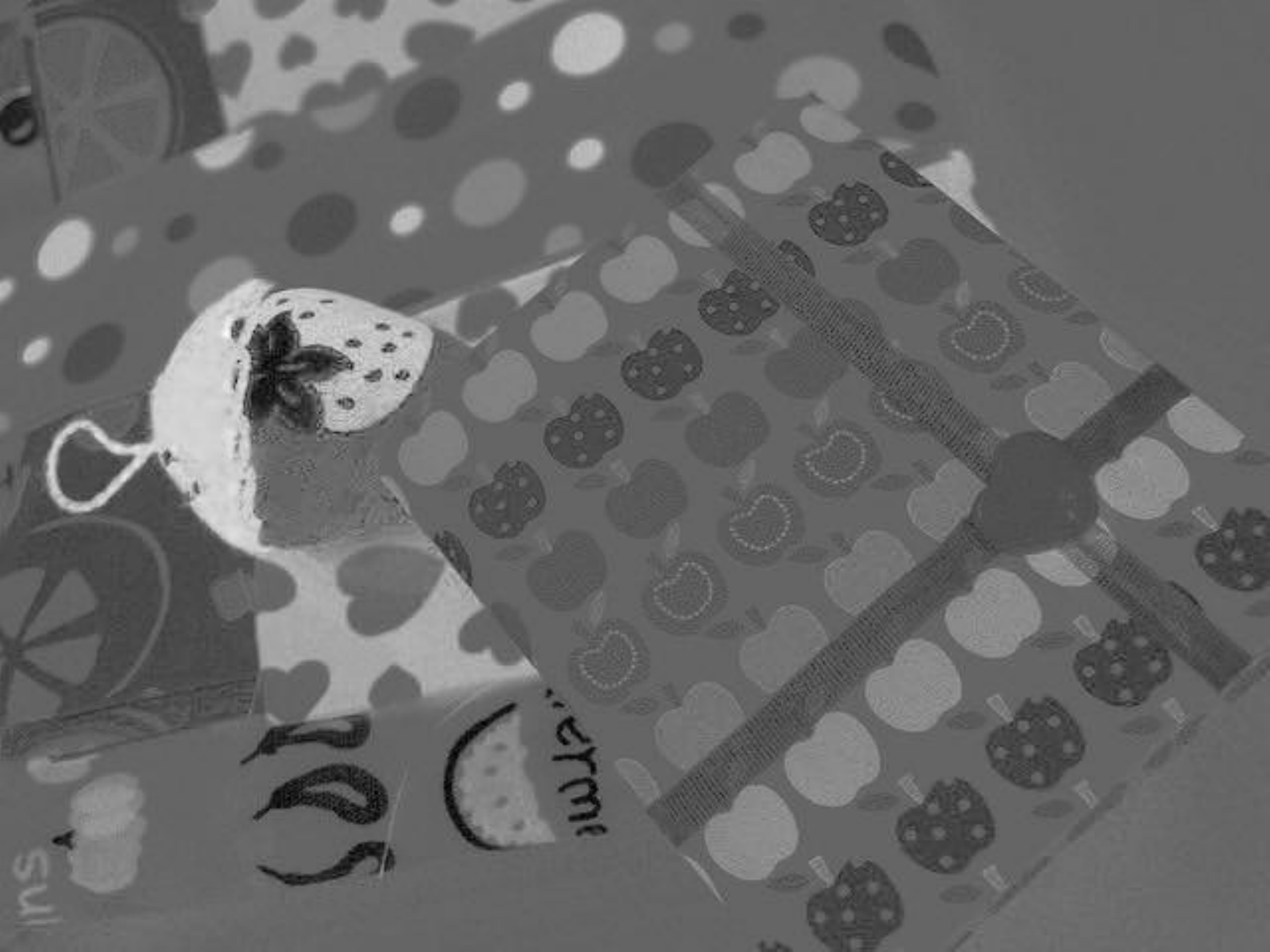}&\tbpic{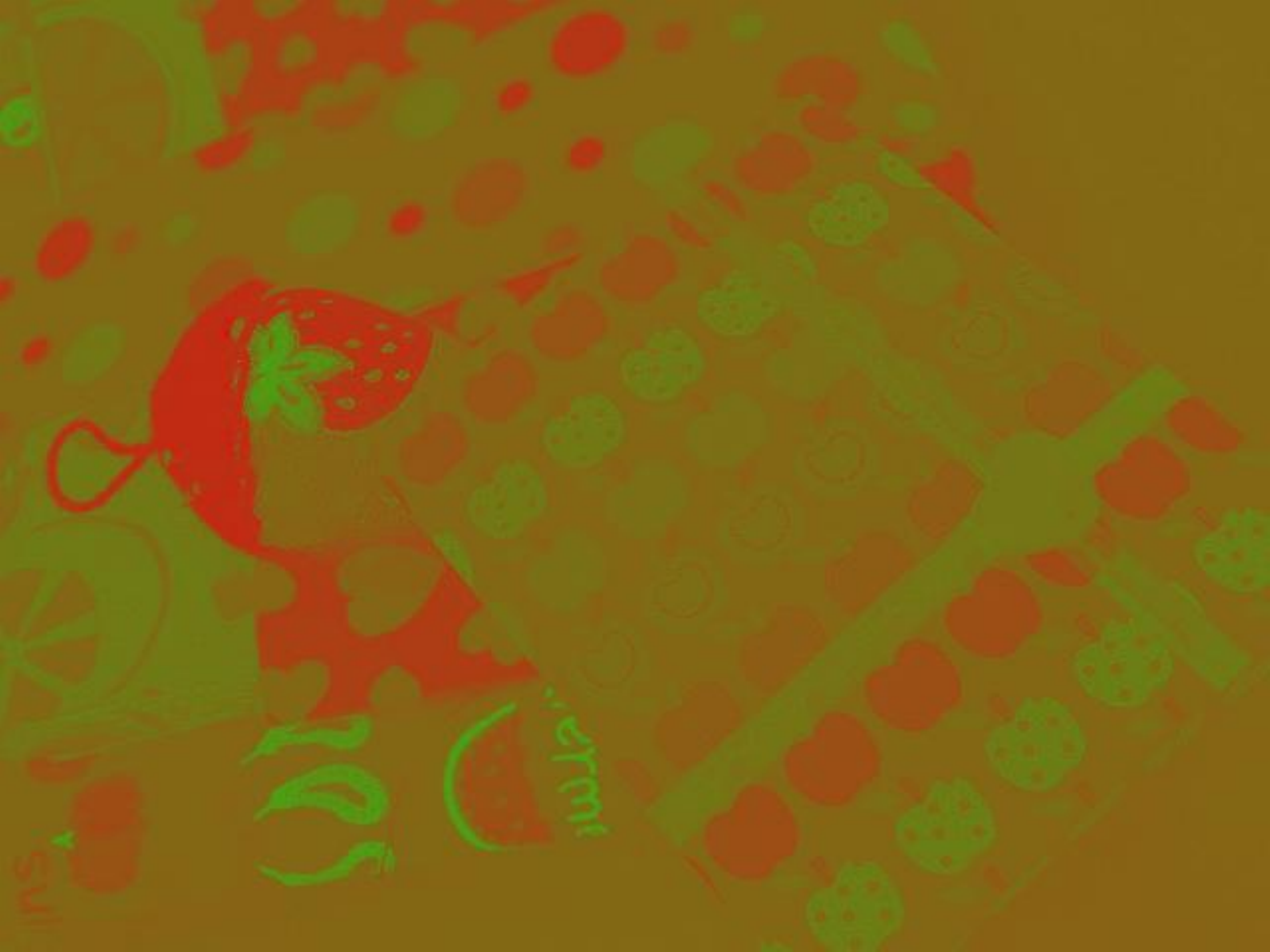}&\tbpic{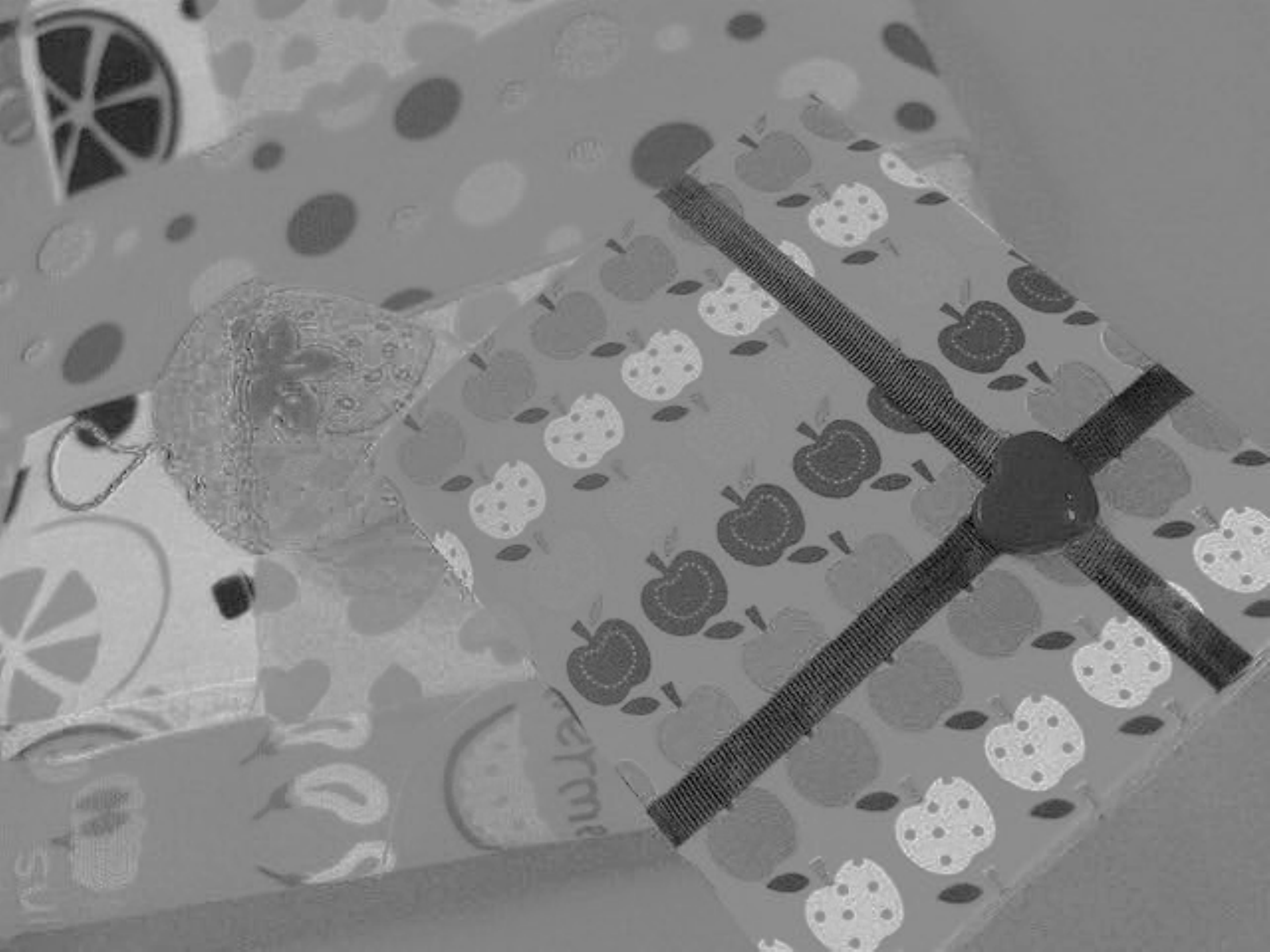}&\tbpic{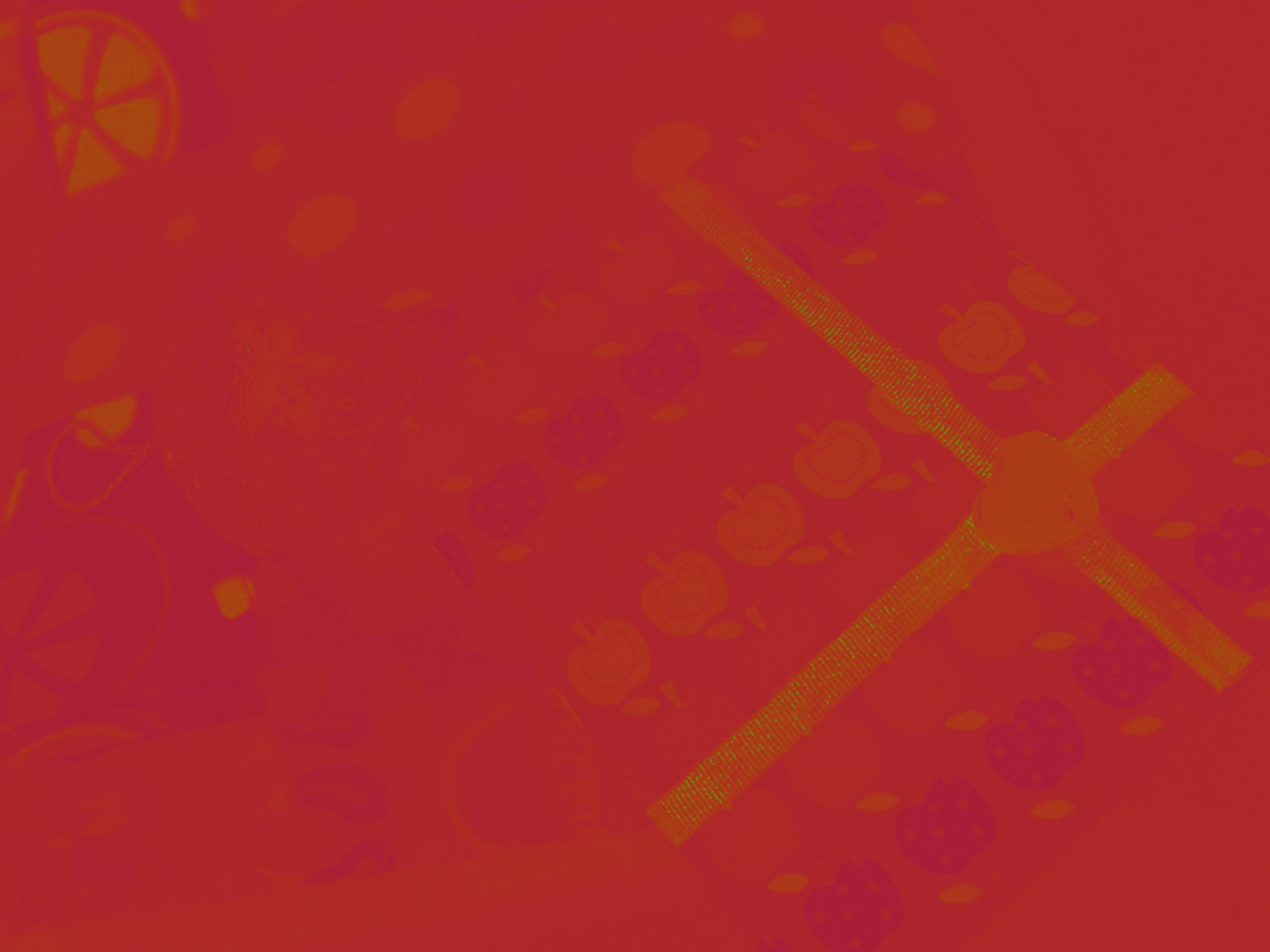}\\ \hline
5&\tbpic{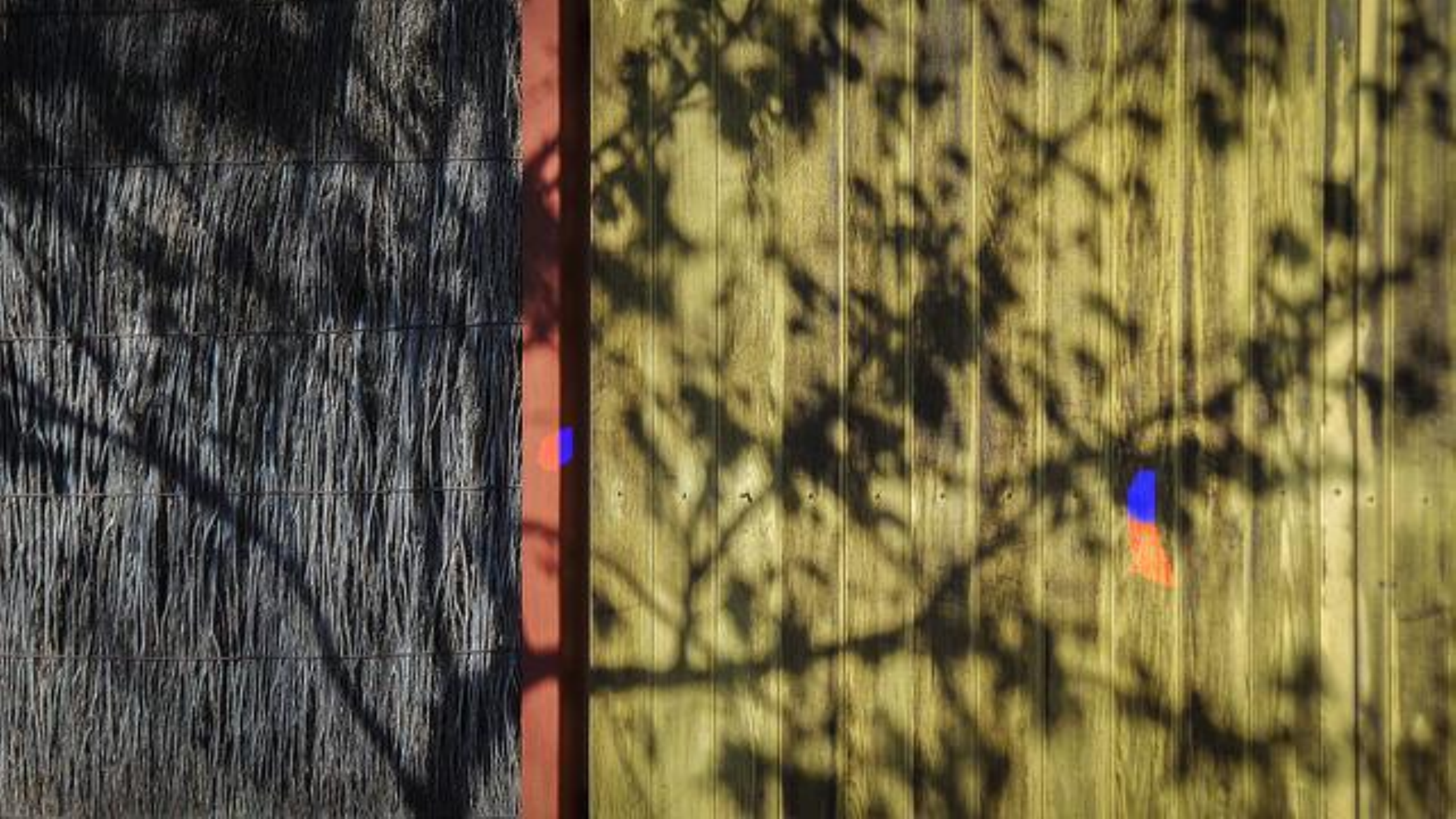}&\tbpic{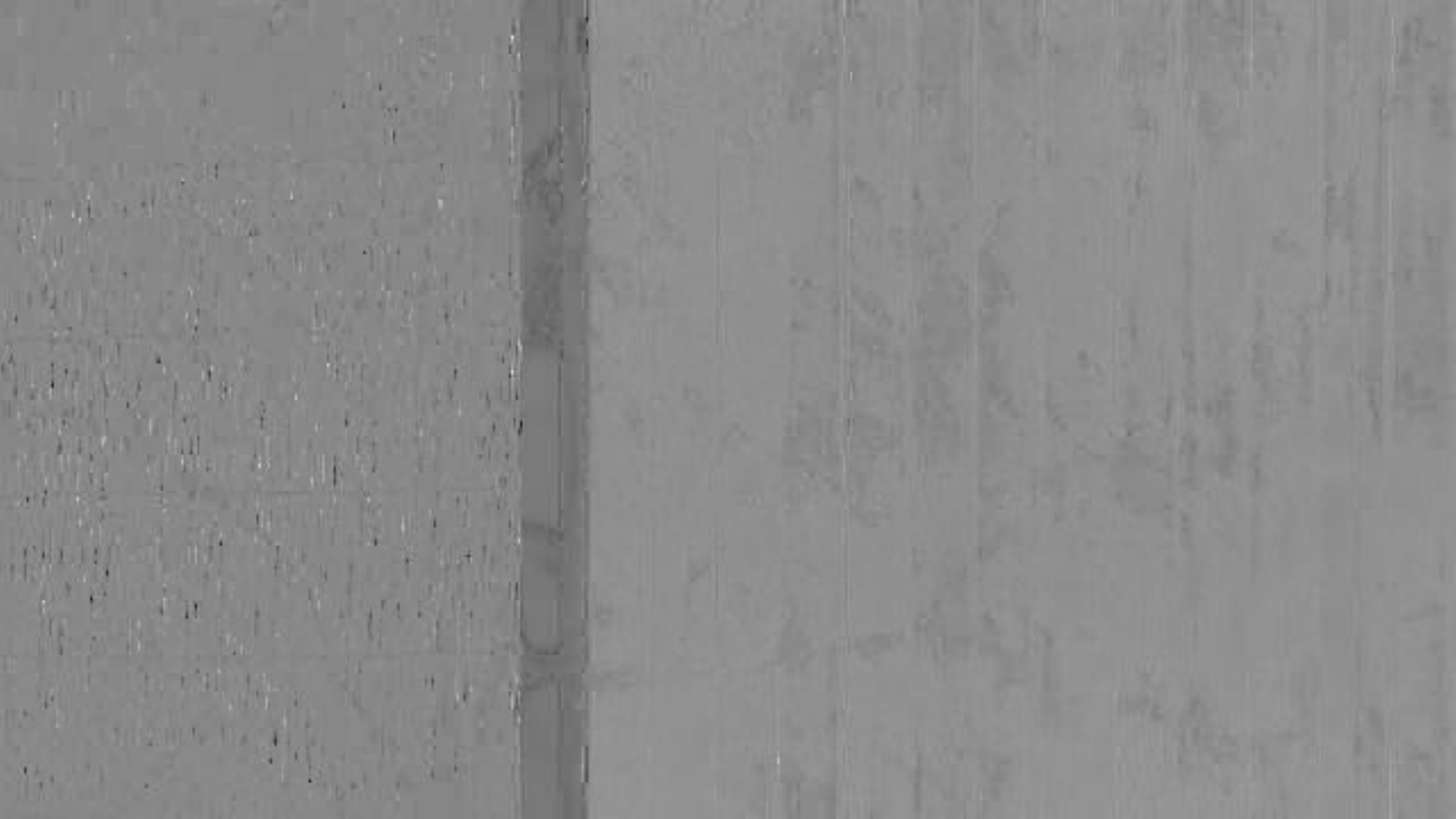}&\tbpic{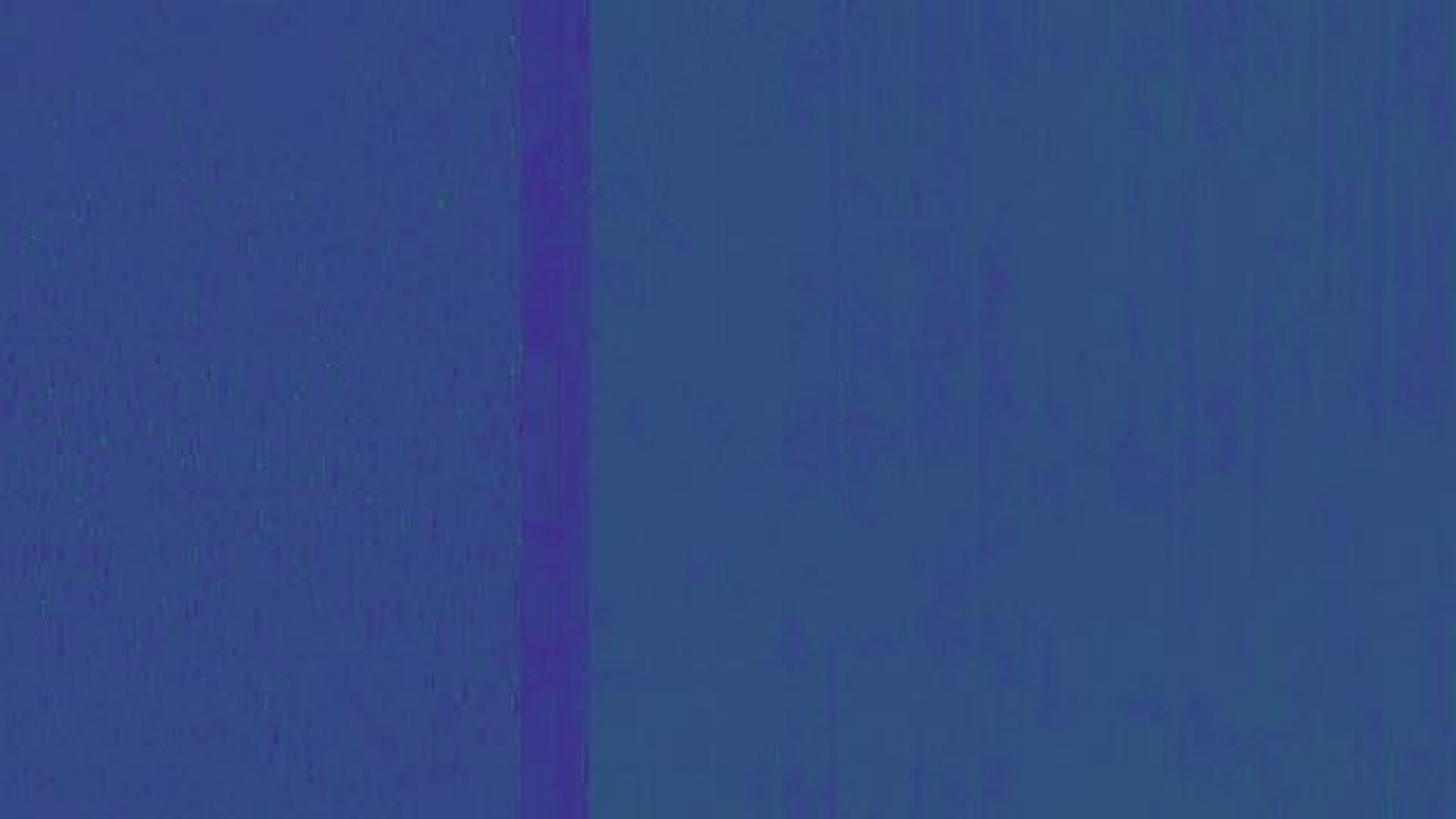}&\tbpic{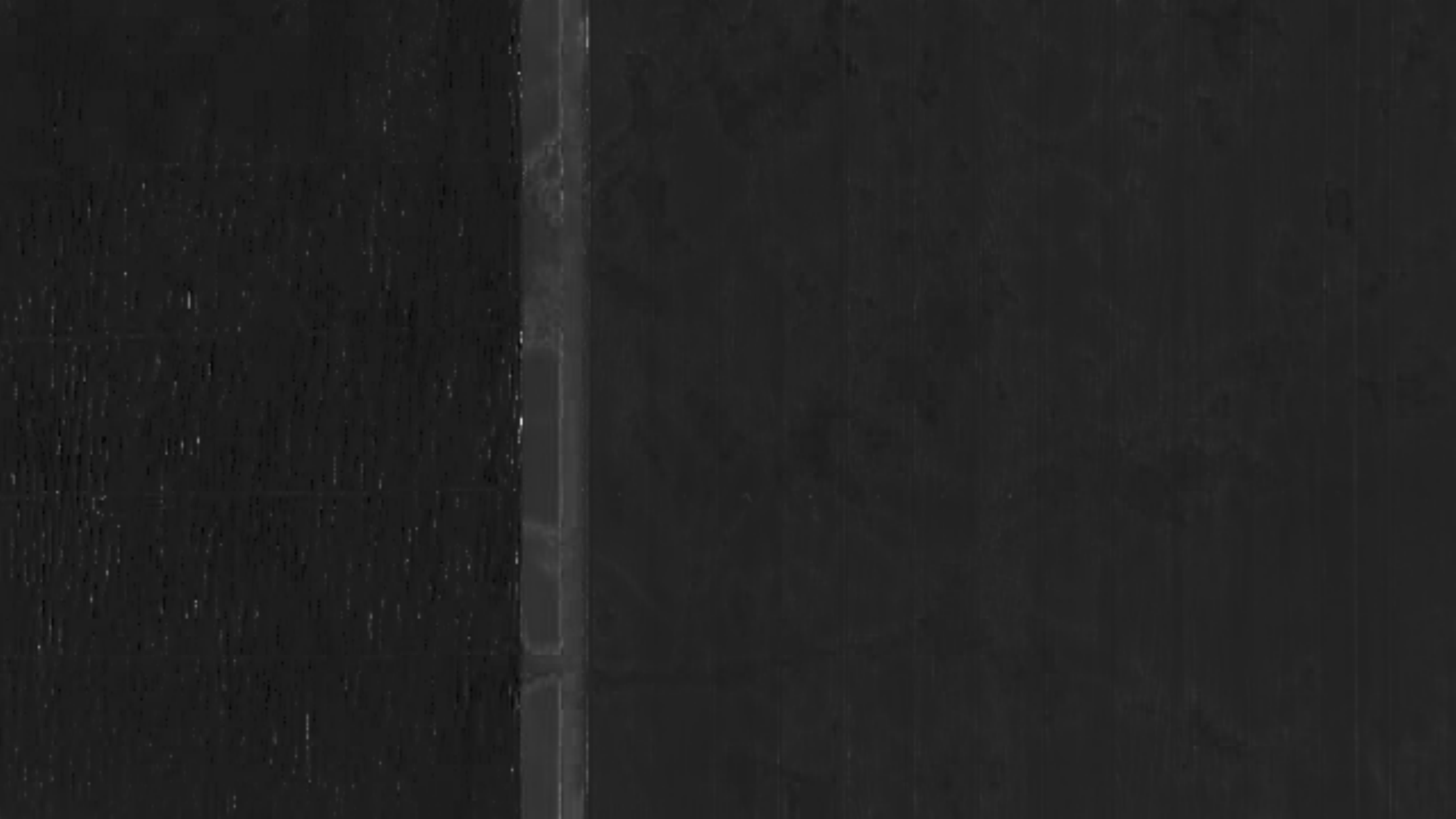}&\tbpic{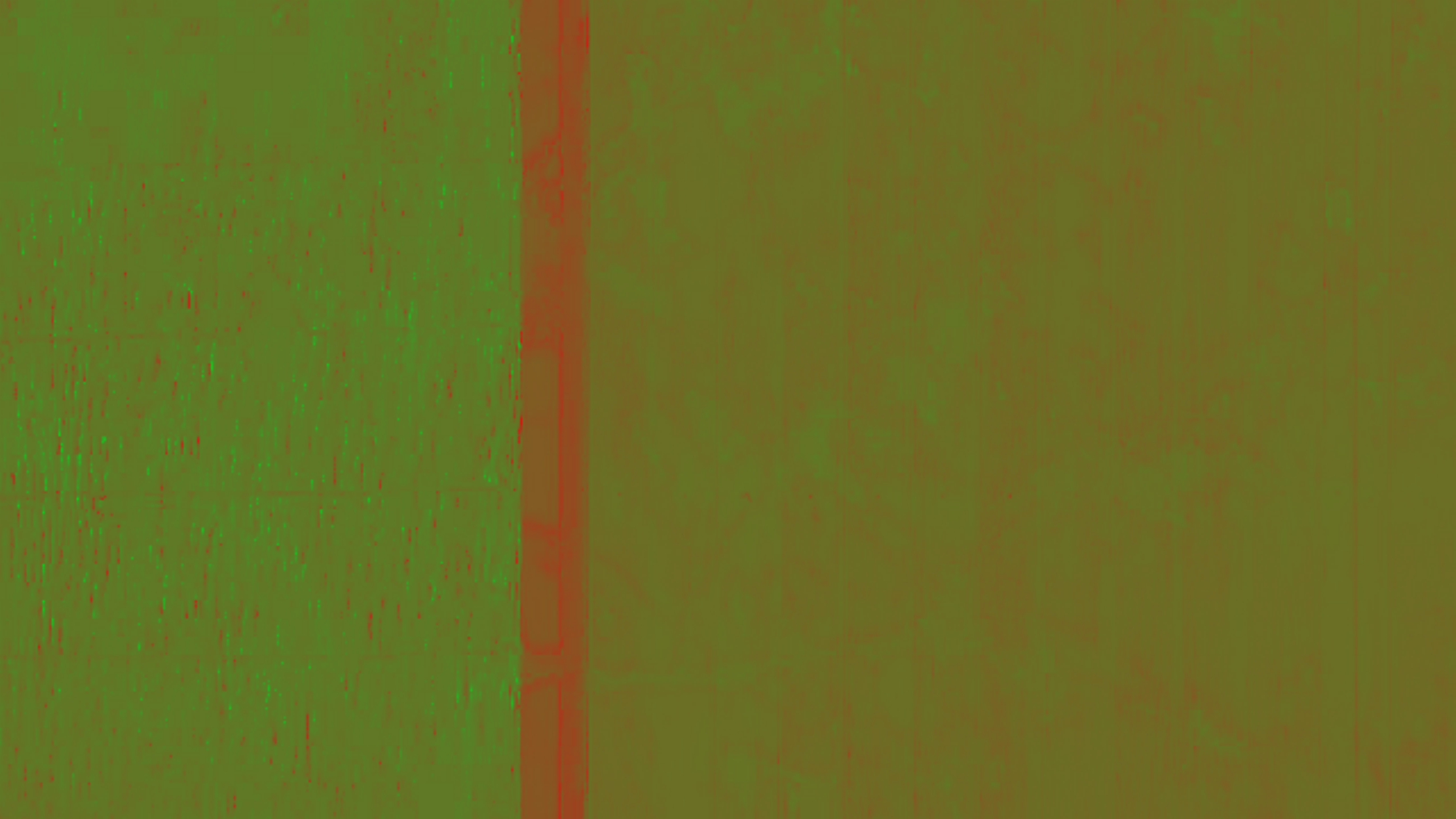}\\ \hline
6&\tbpic{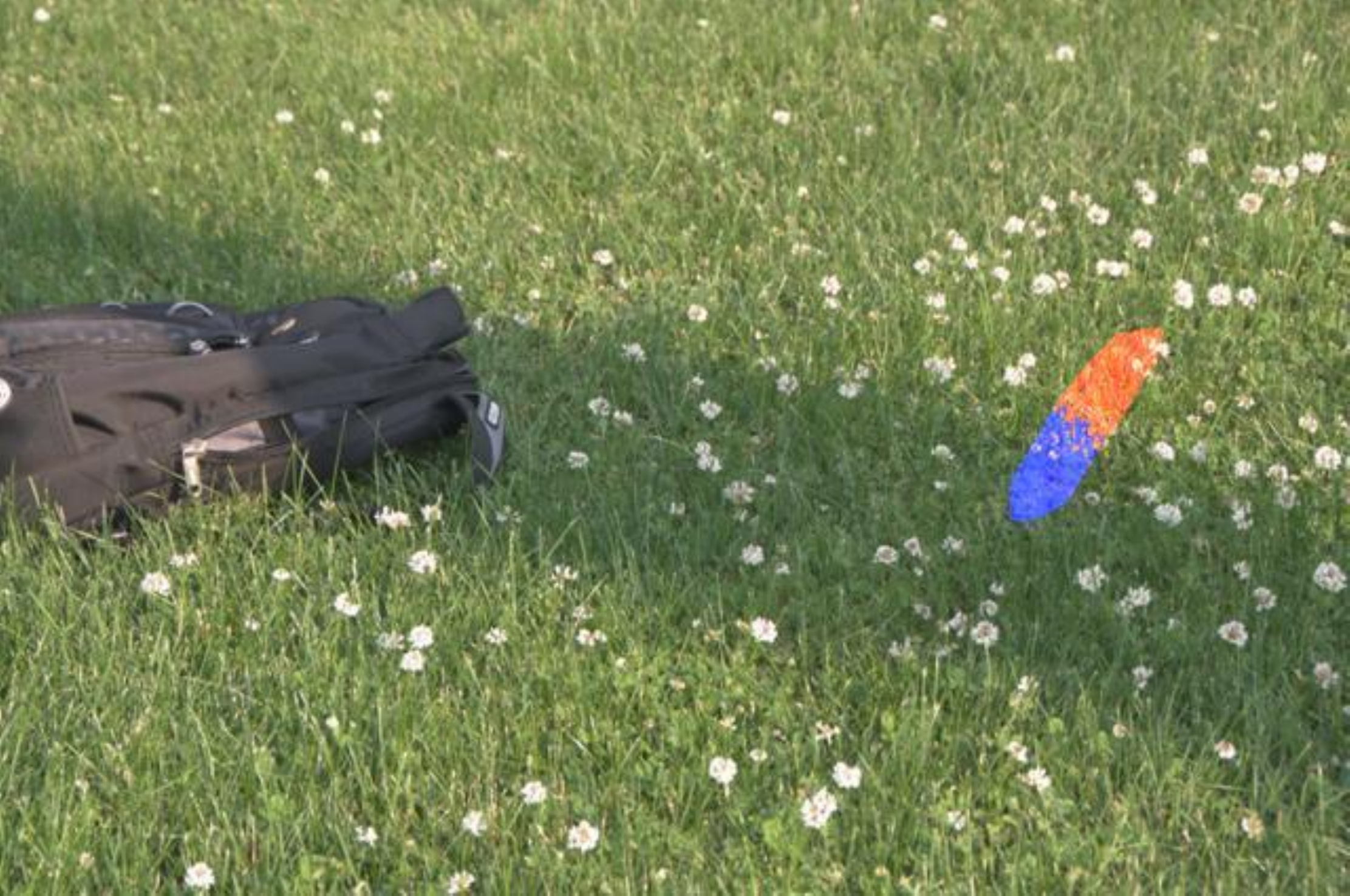}&\tbpic{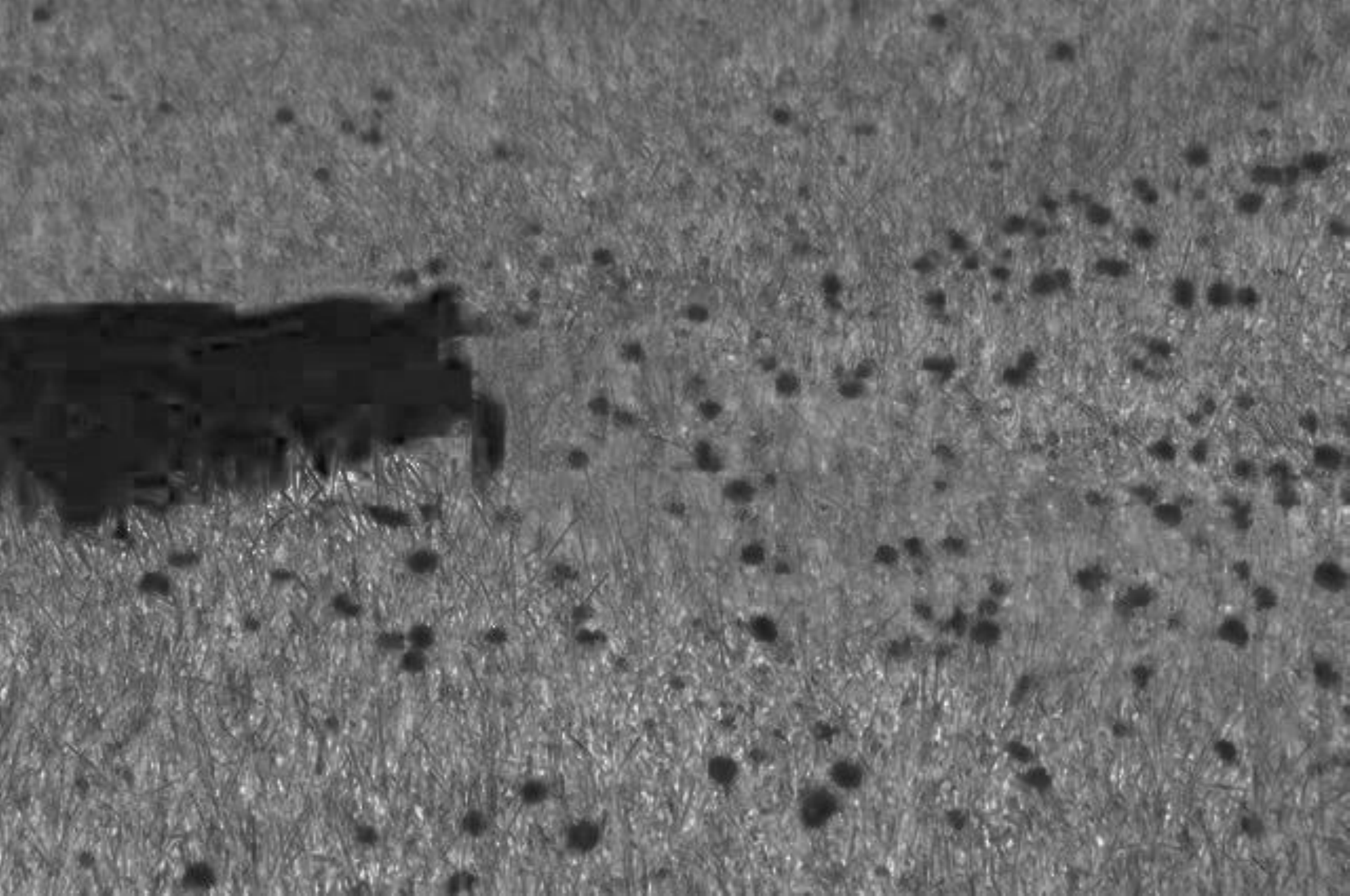}&\tbpic{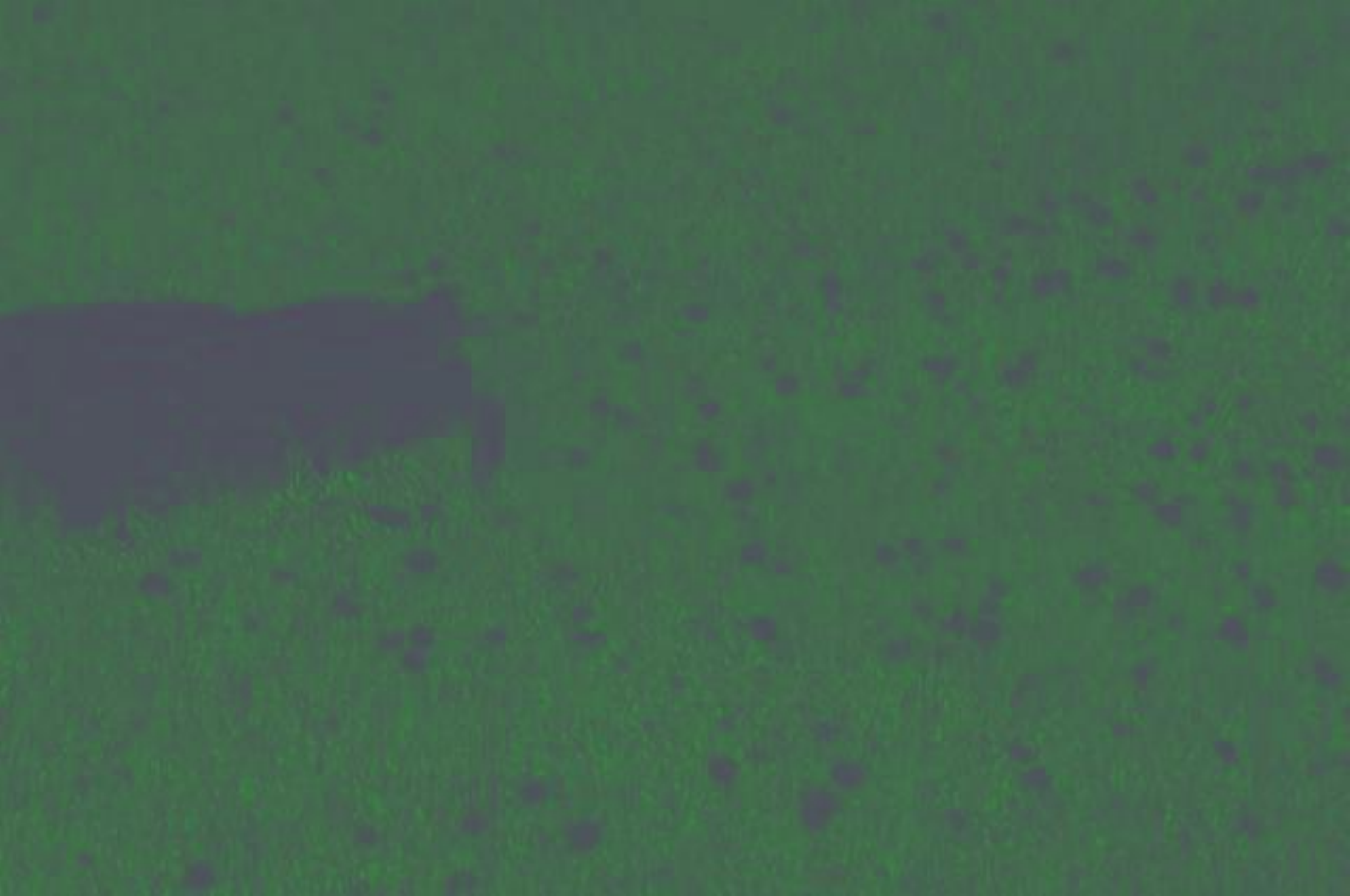}&\tbpic{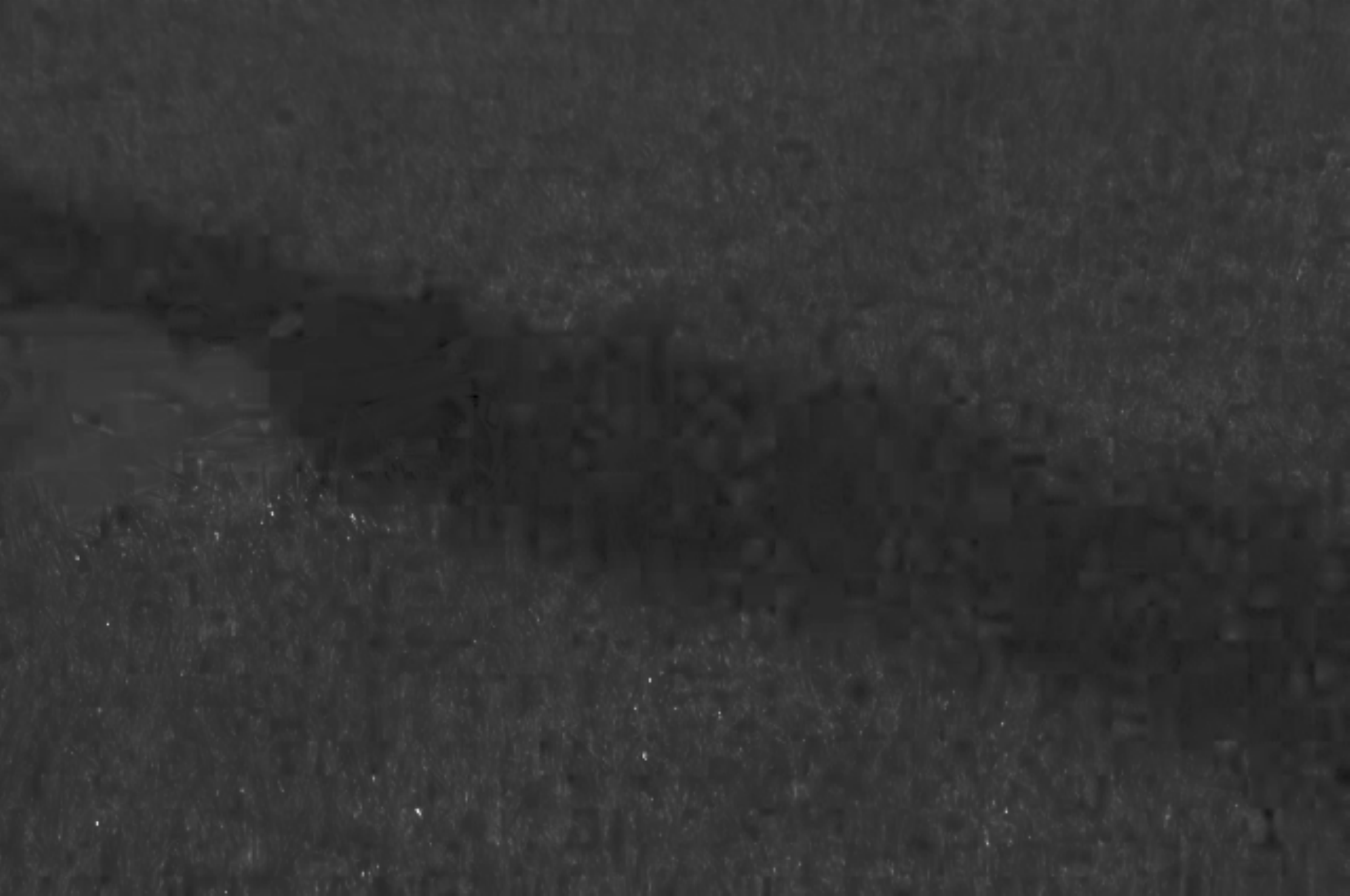}&\tbpic{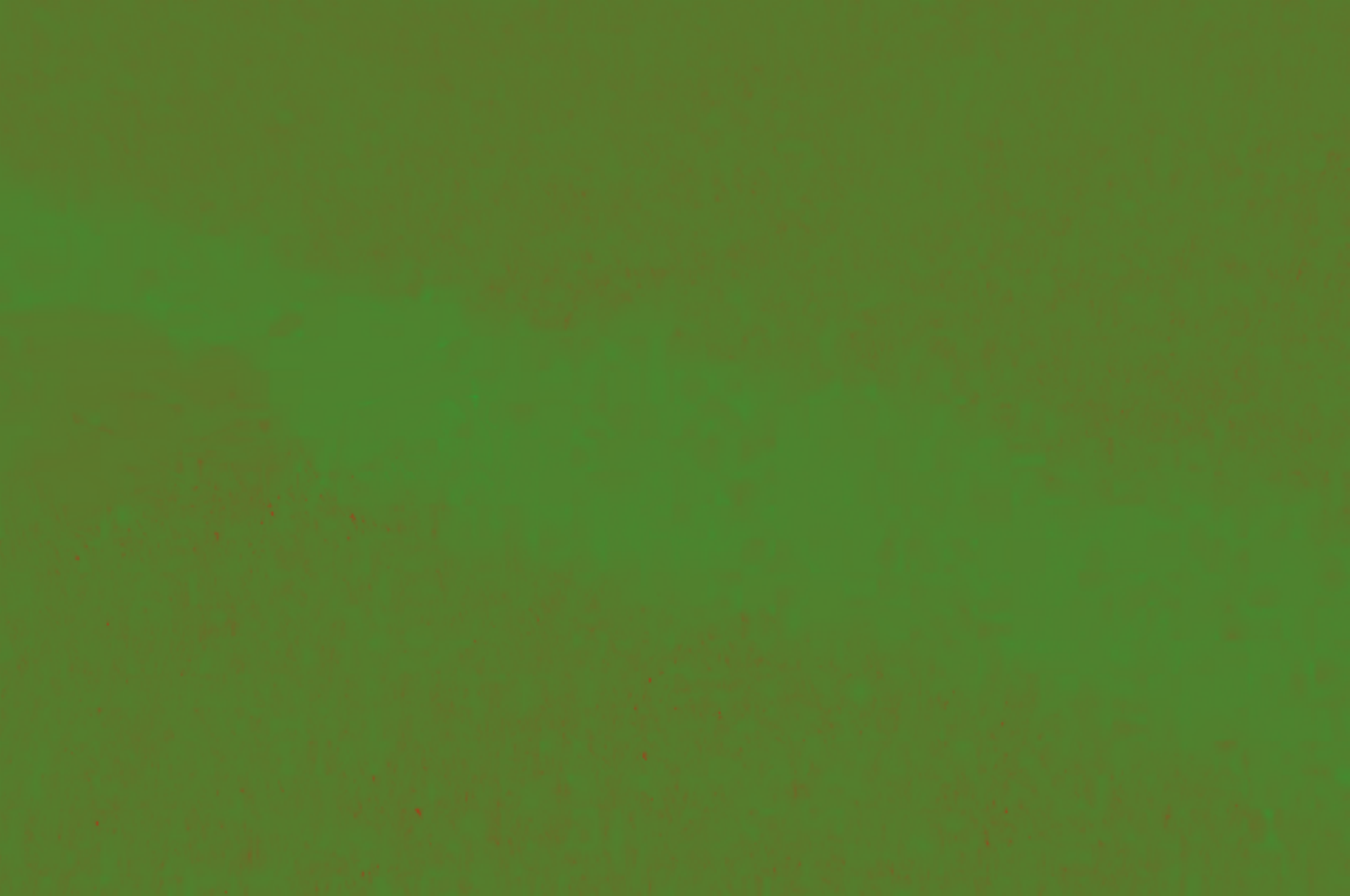}\\ \hline
7&\tbpic{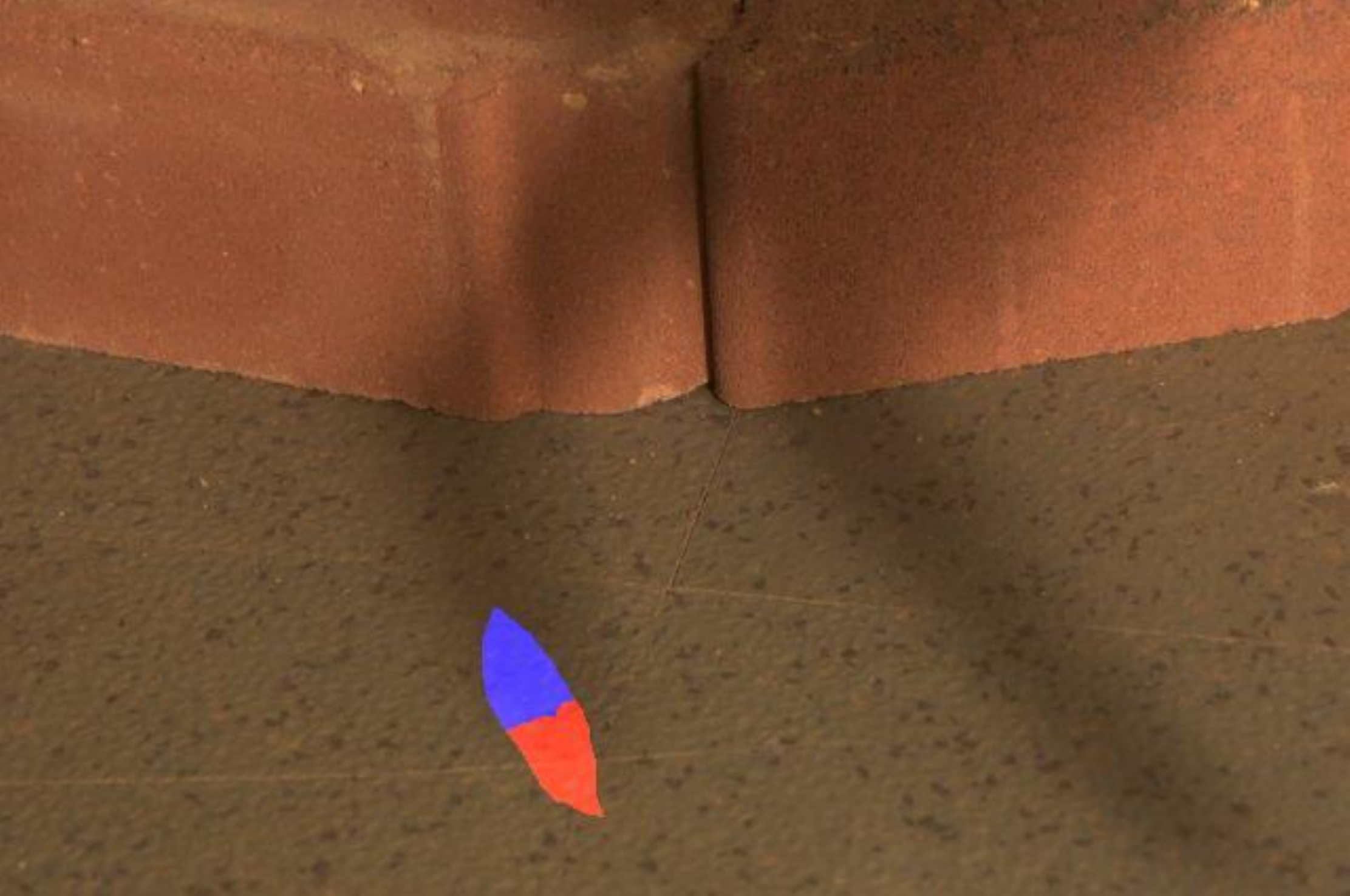}&\tbpic{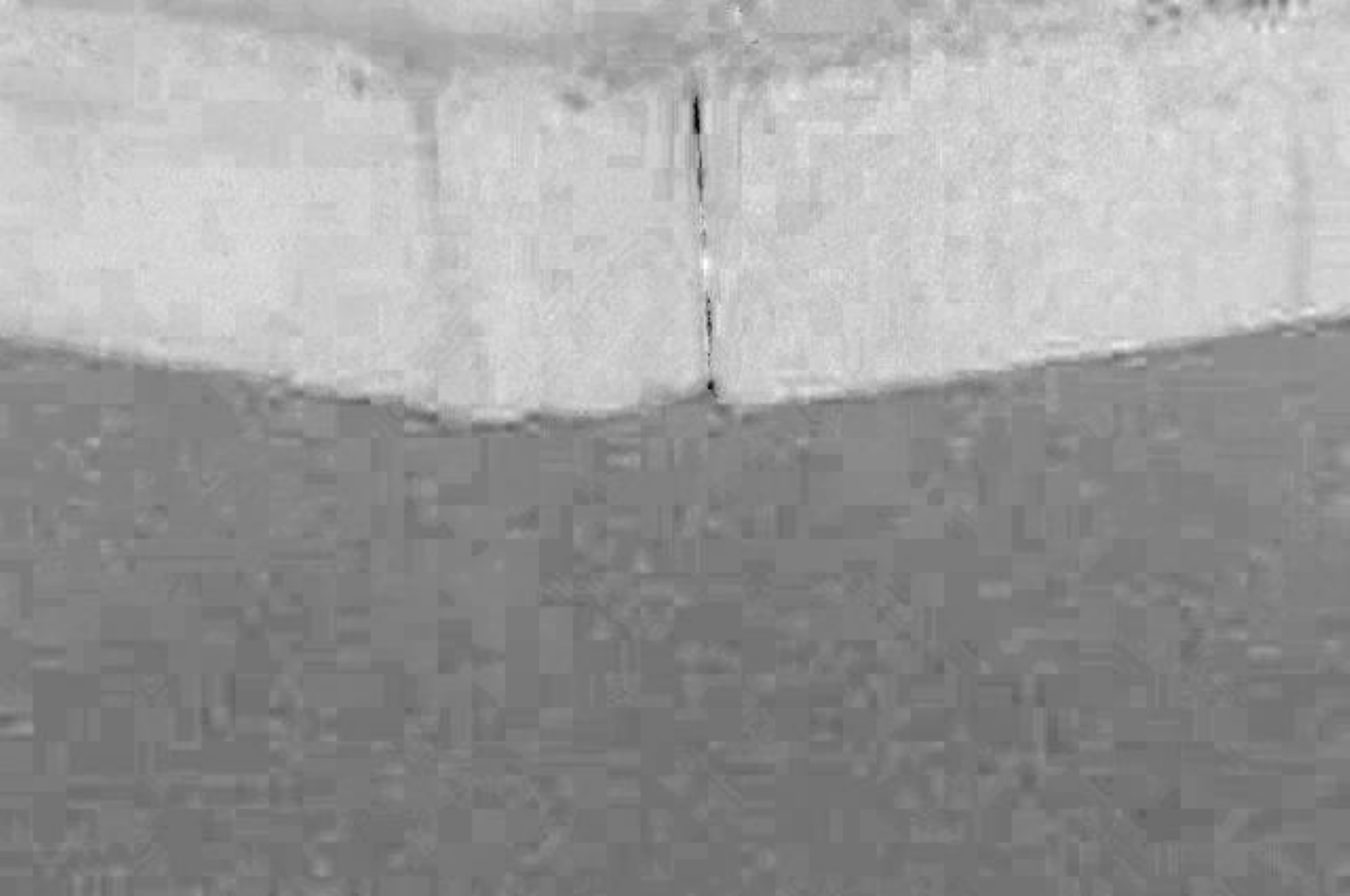}&\tbpic{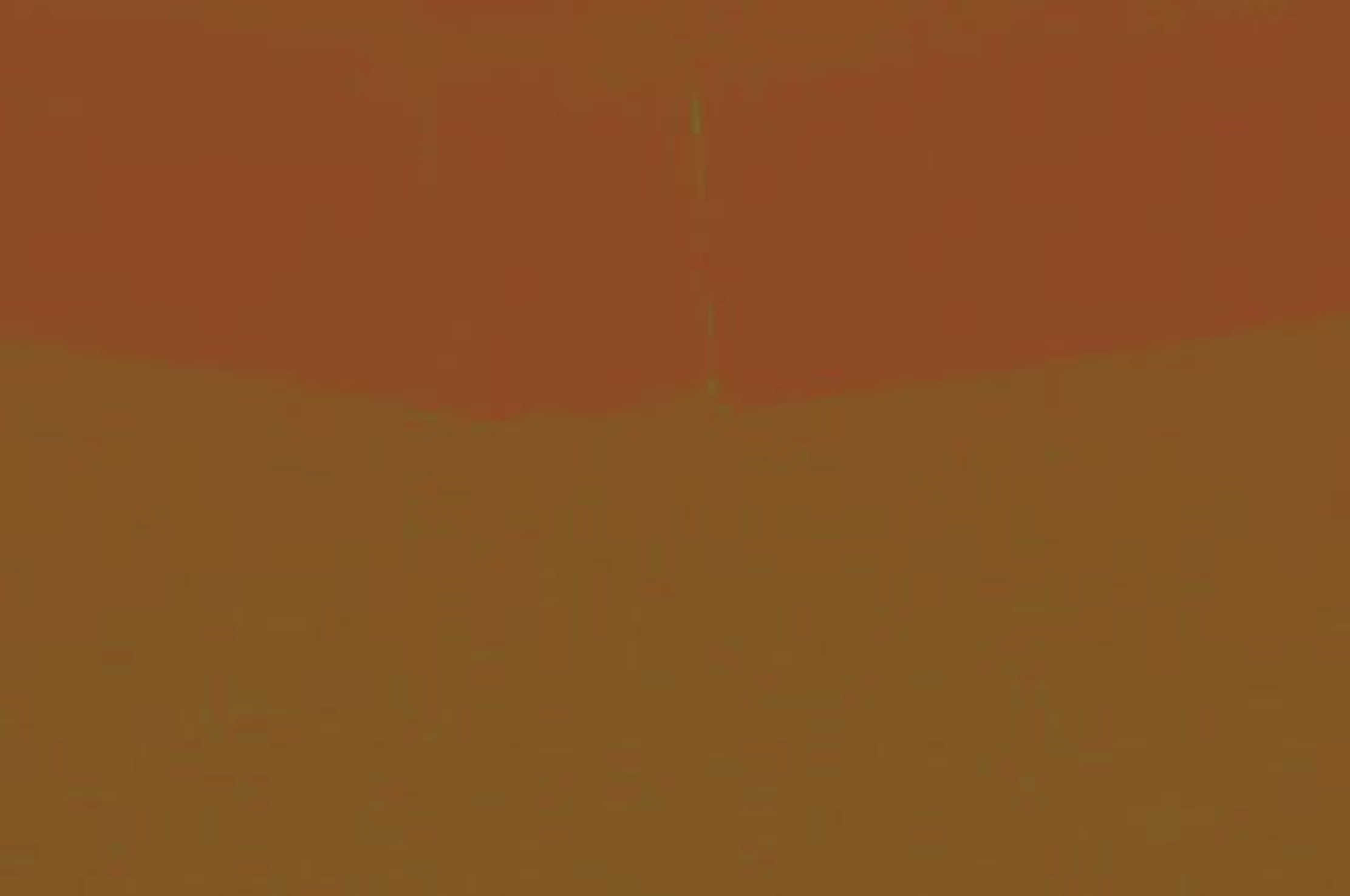}&\tbpic{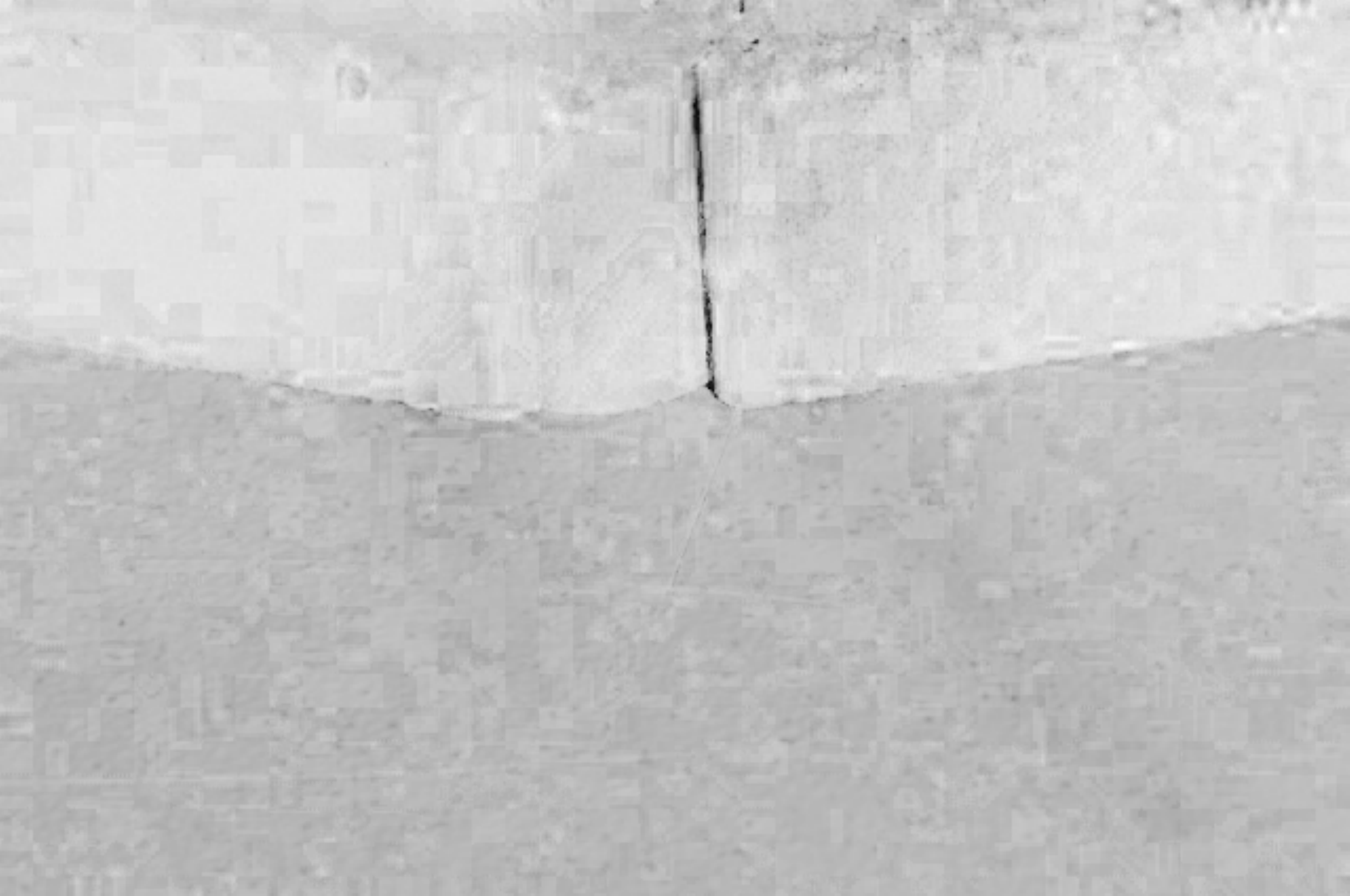}&\tbpic{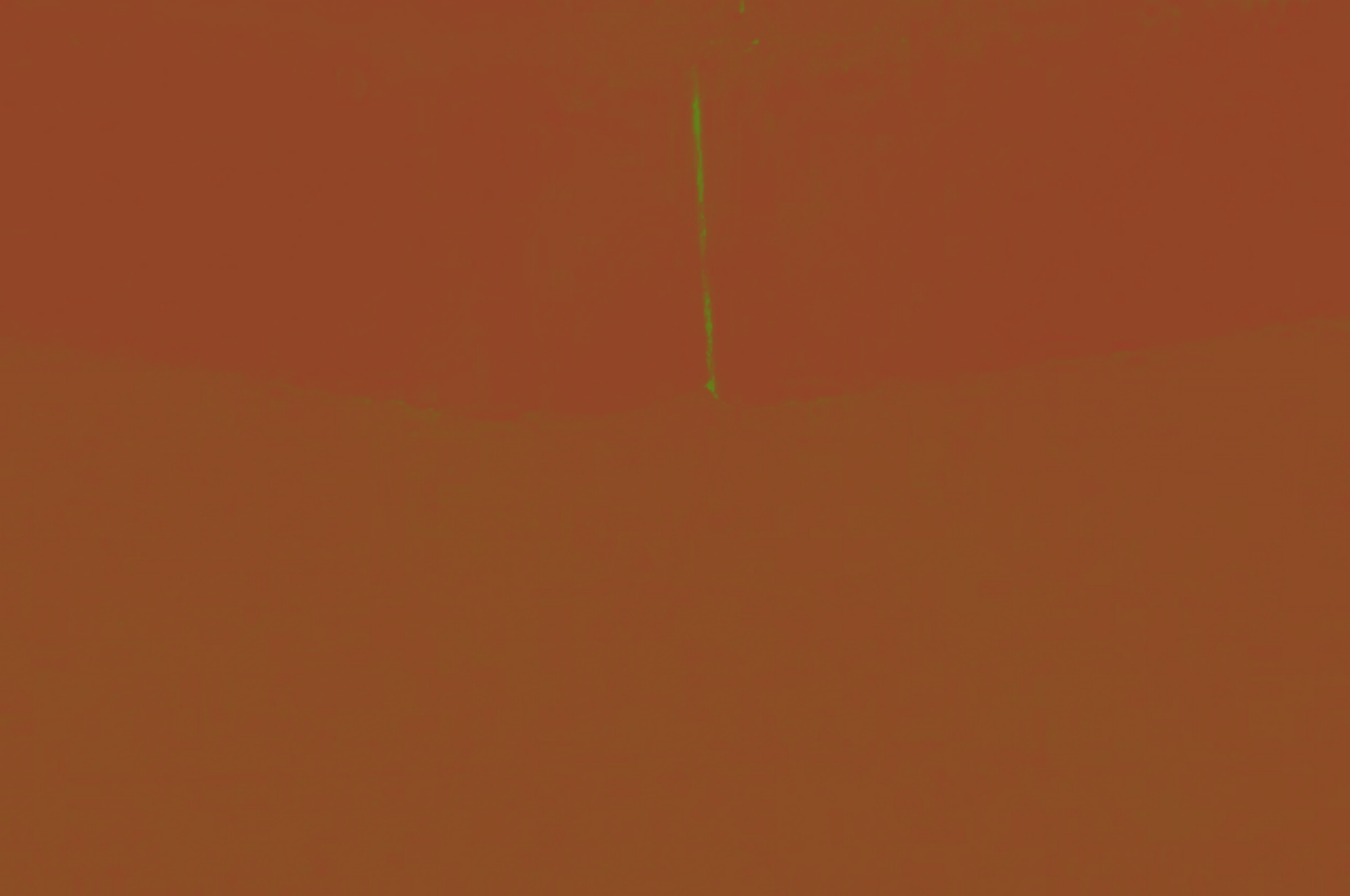}\\ \hline
8&\tbpic{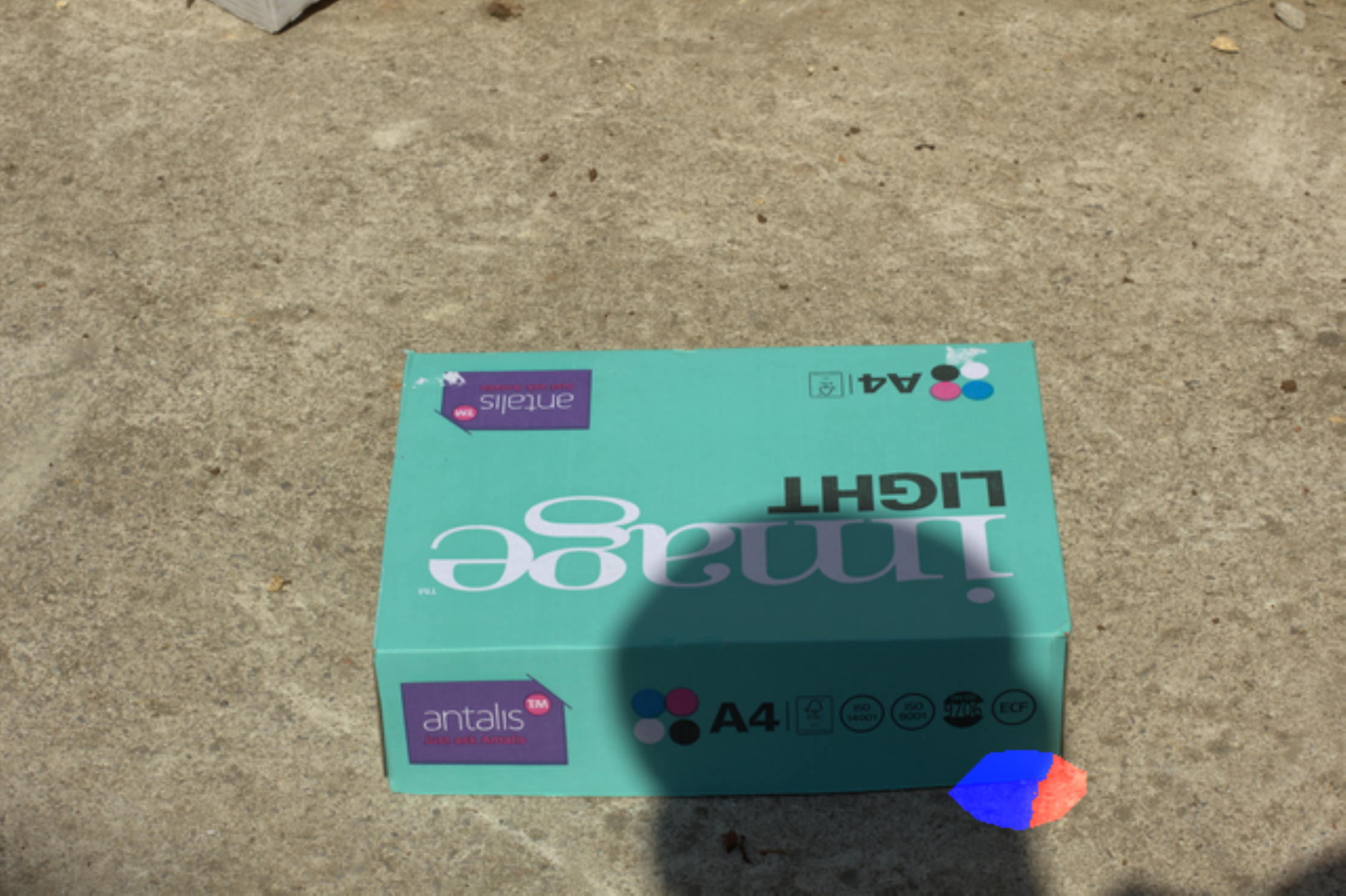}&\tbpic{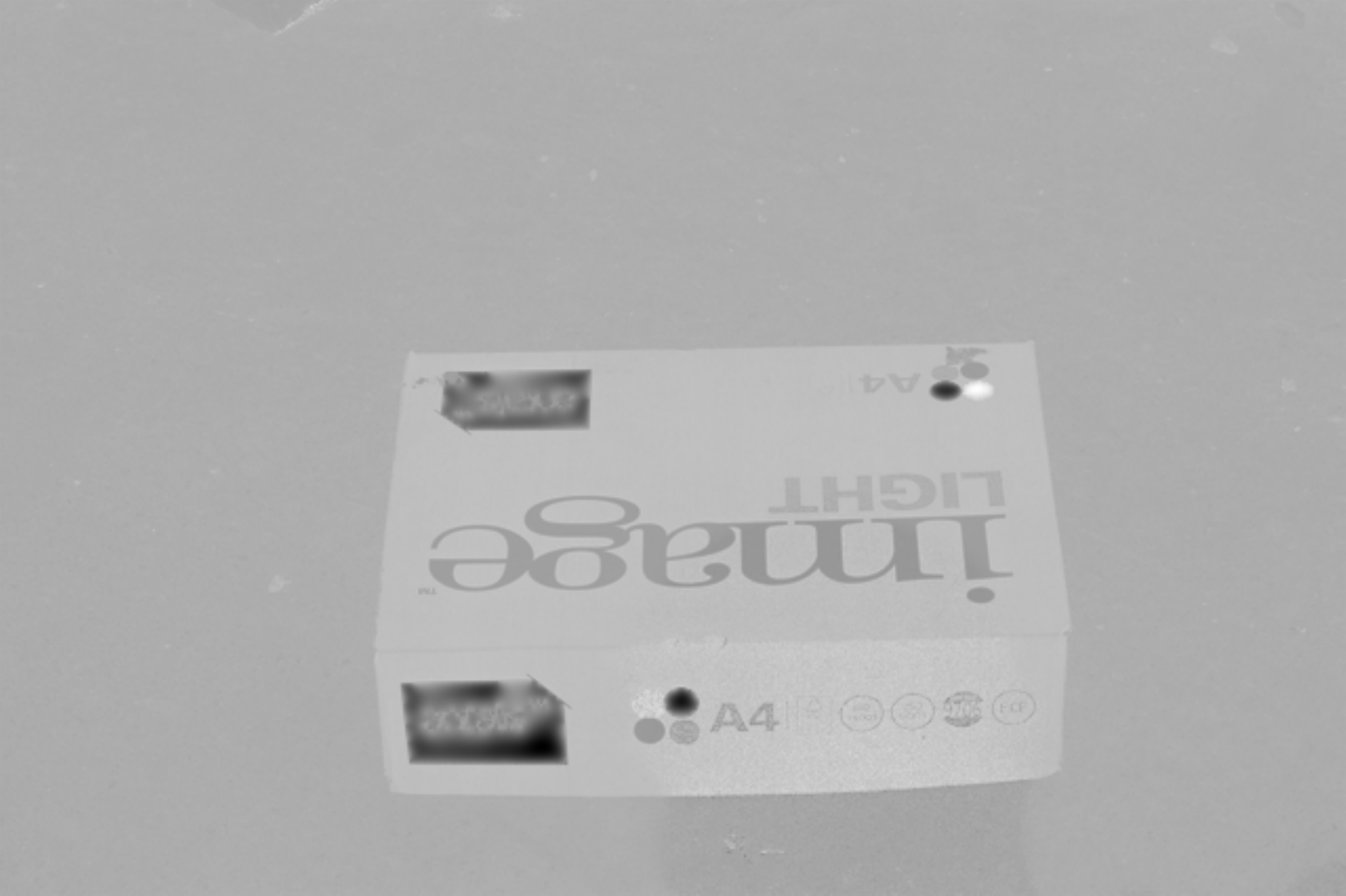}&\tbpic{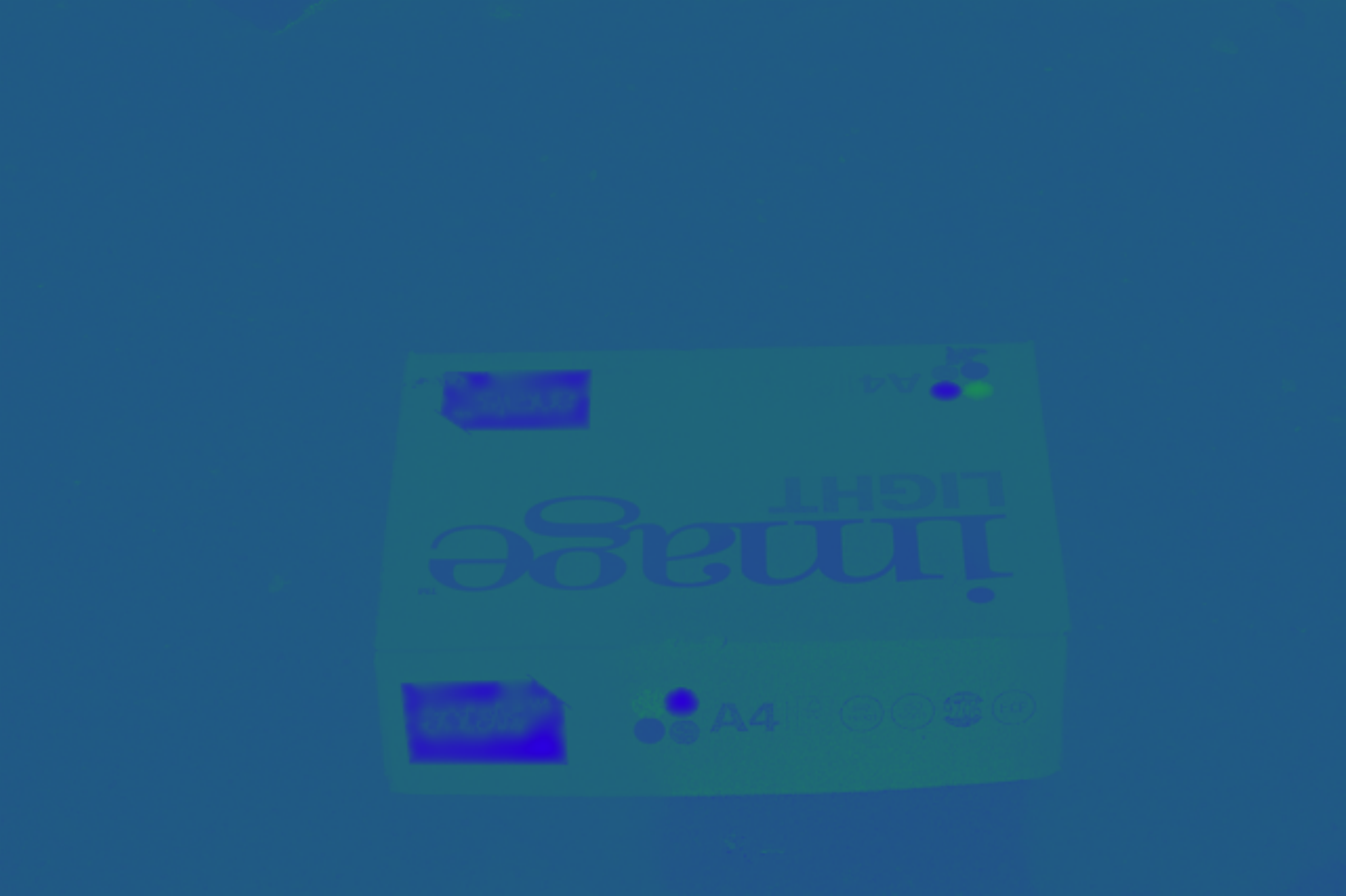}&\tbpic{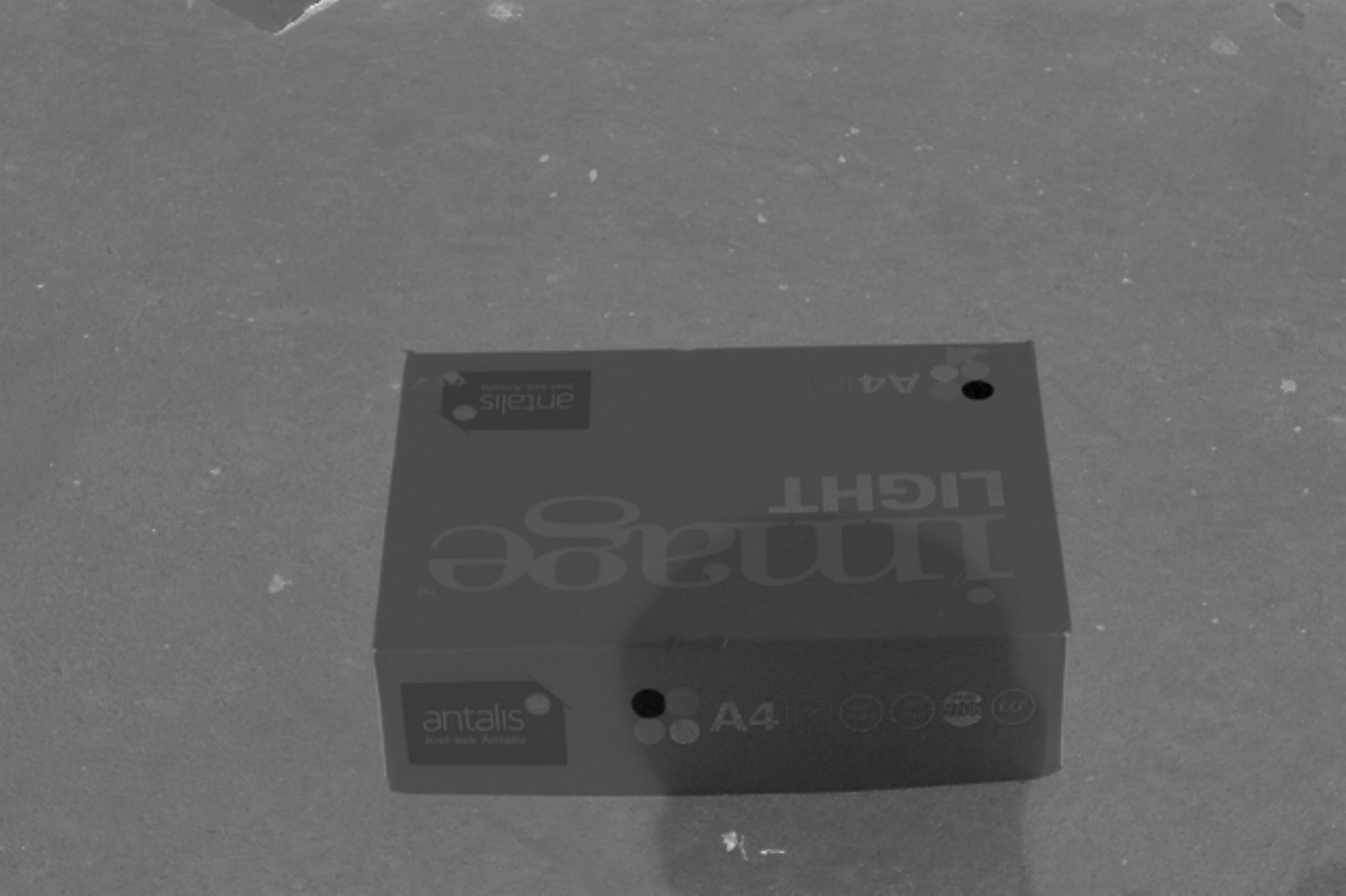}&\tbpic{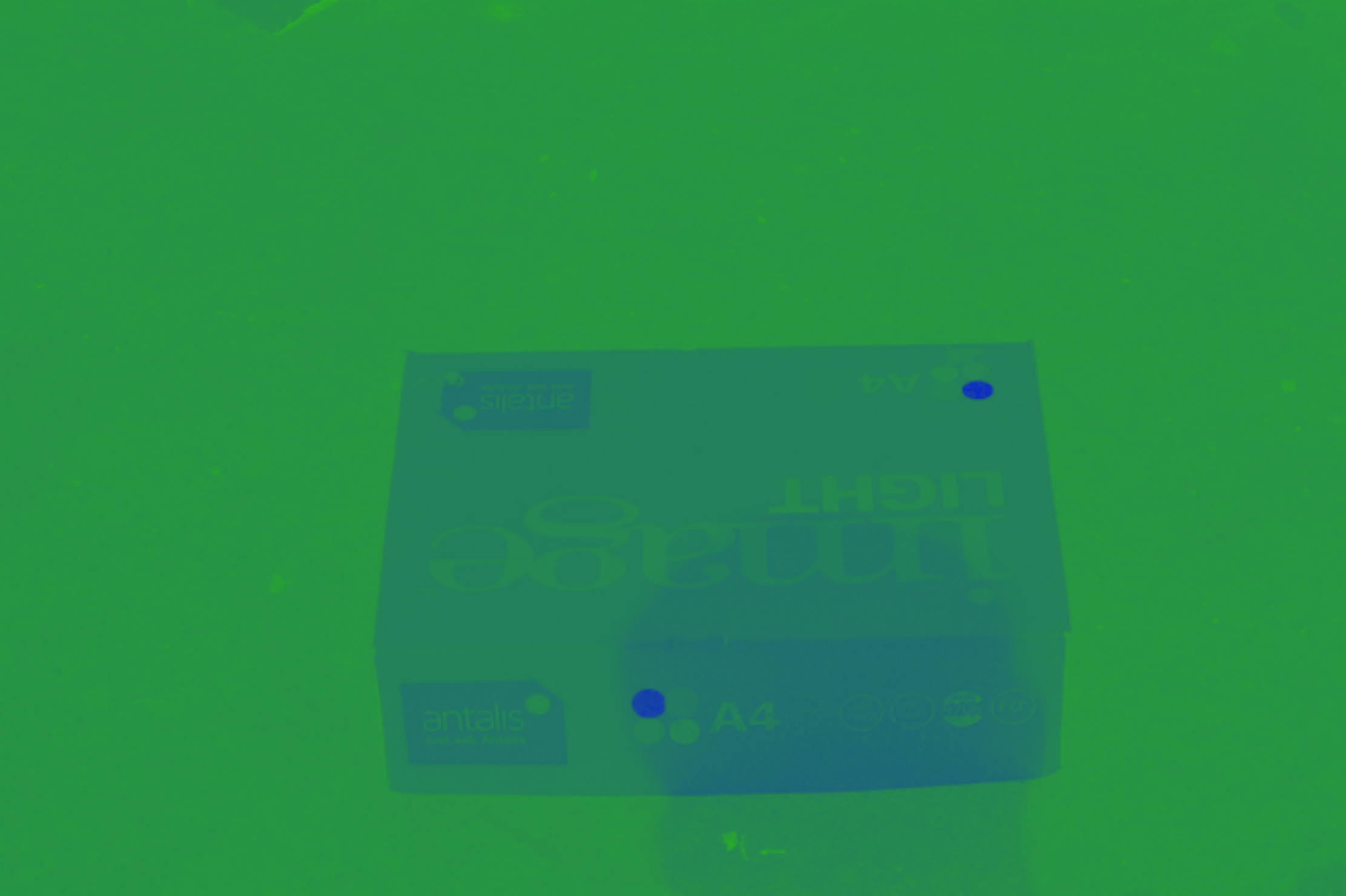}\\ \hline
9&\tbpic{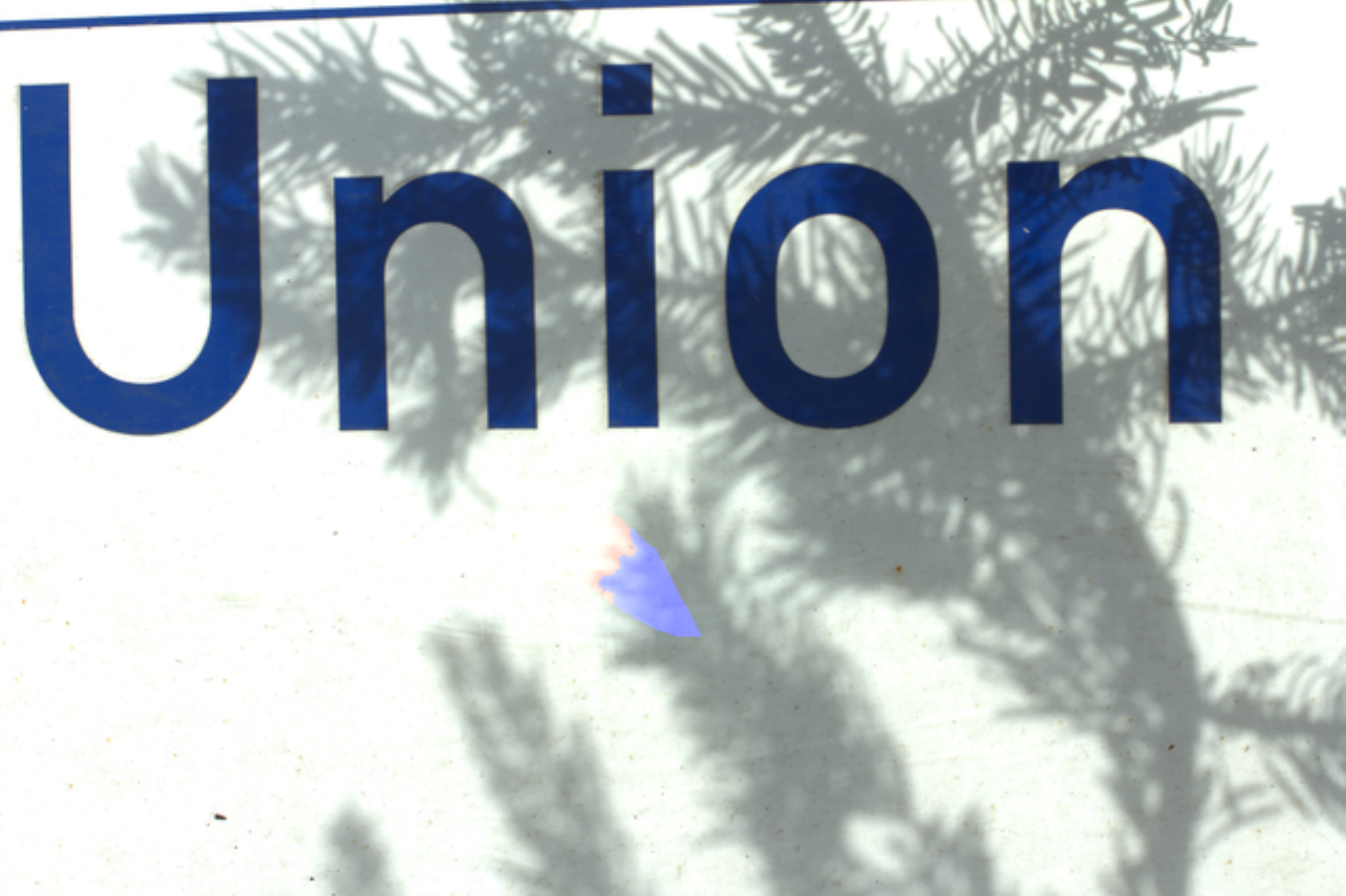}&\tbpic{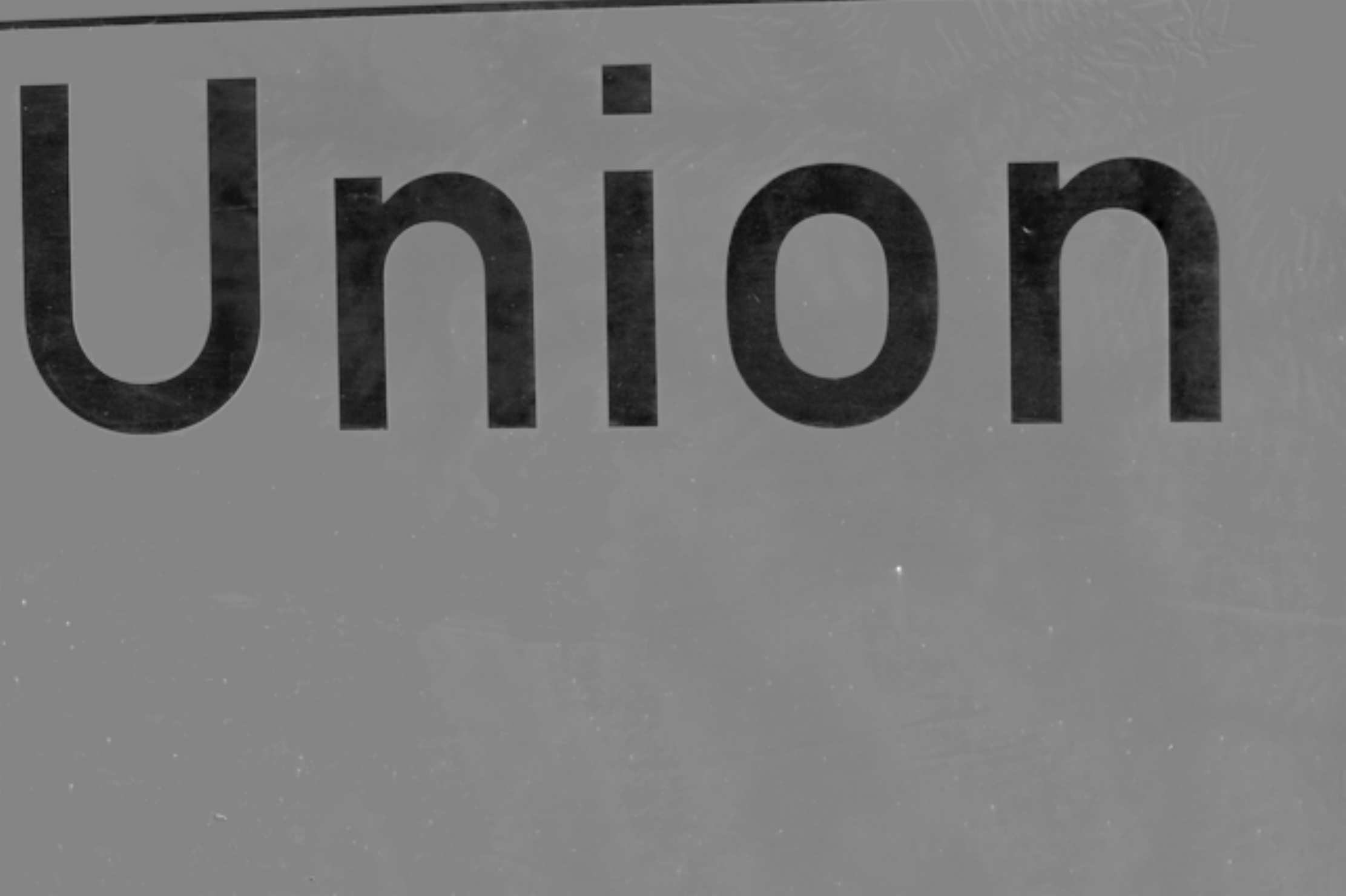}&\tbpic{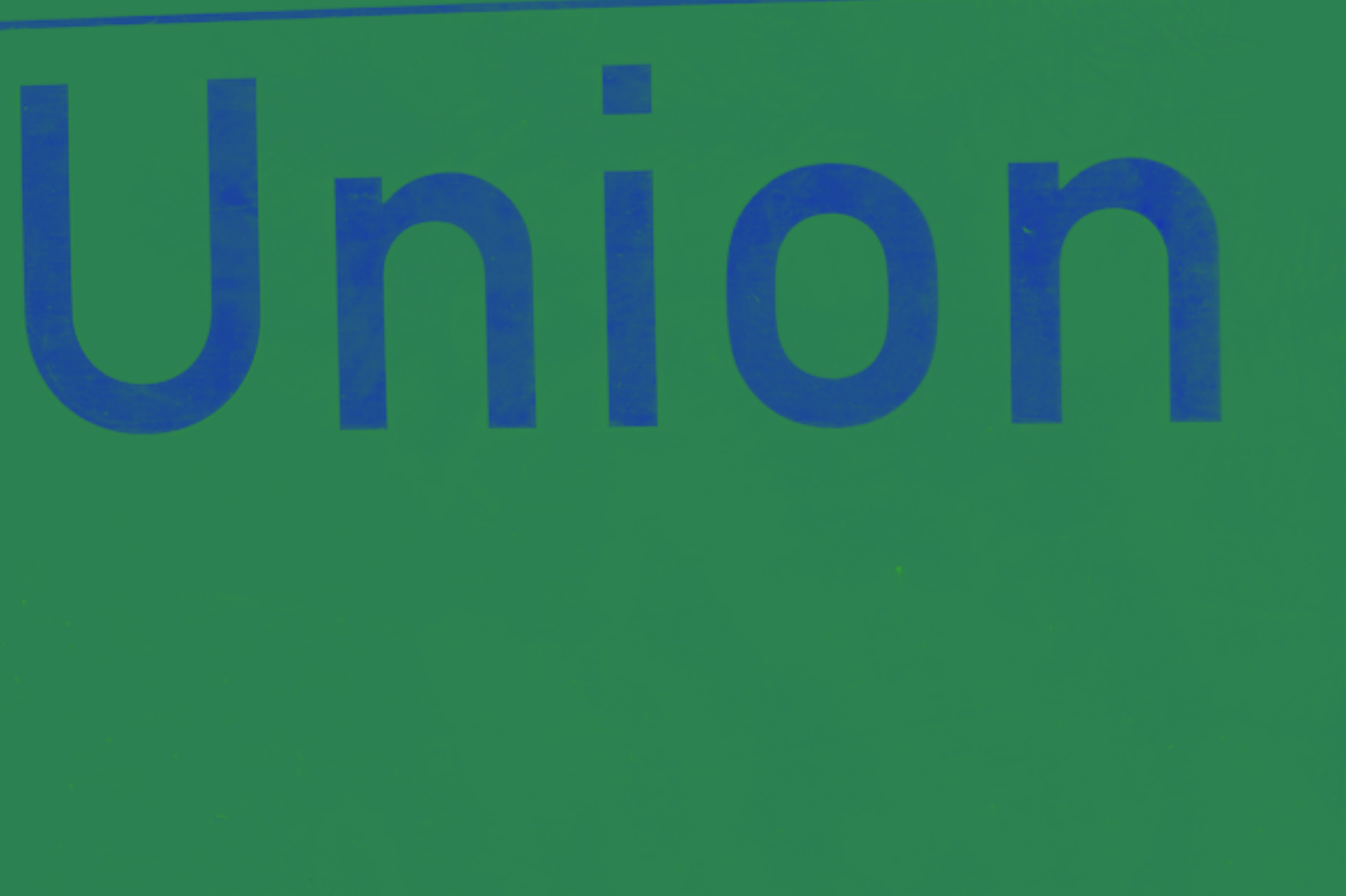}&\tbpic{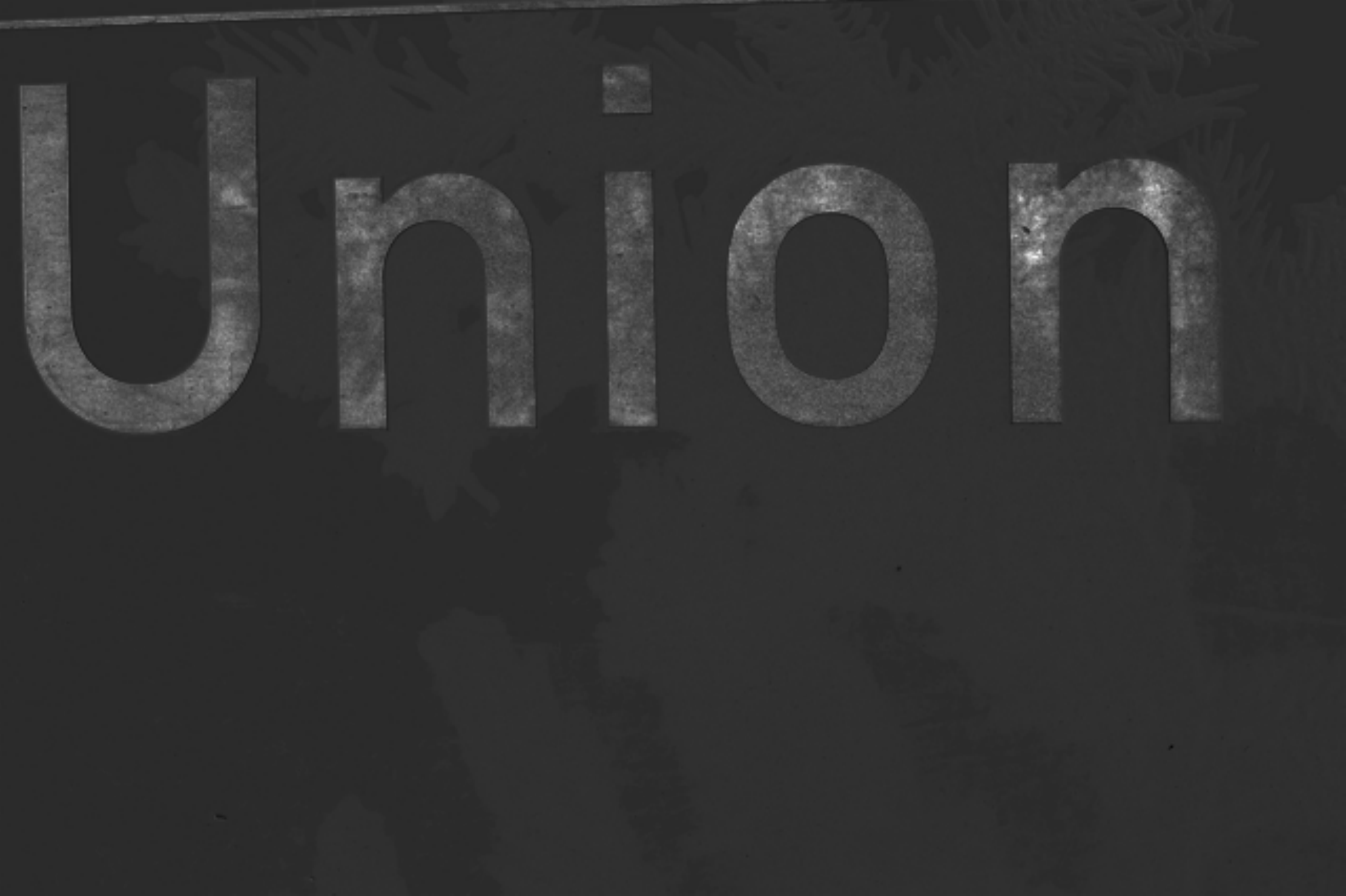}&\tbpic{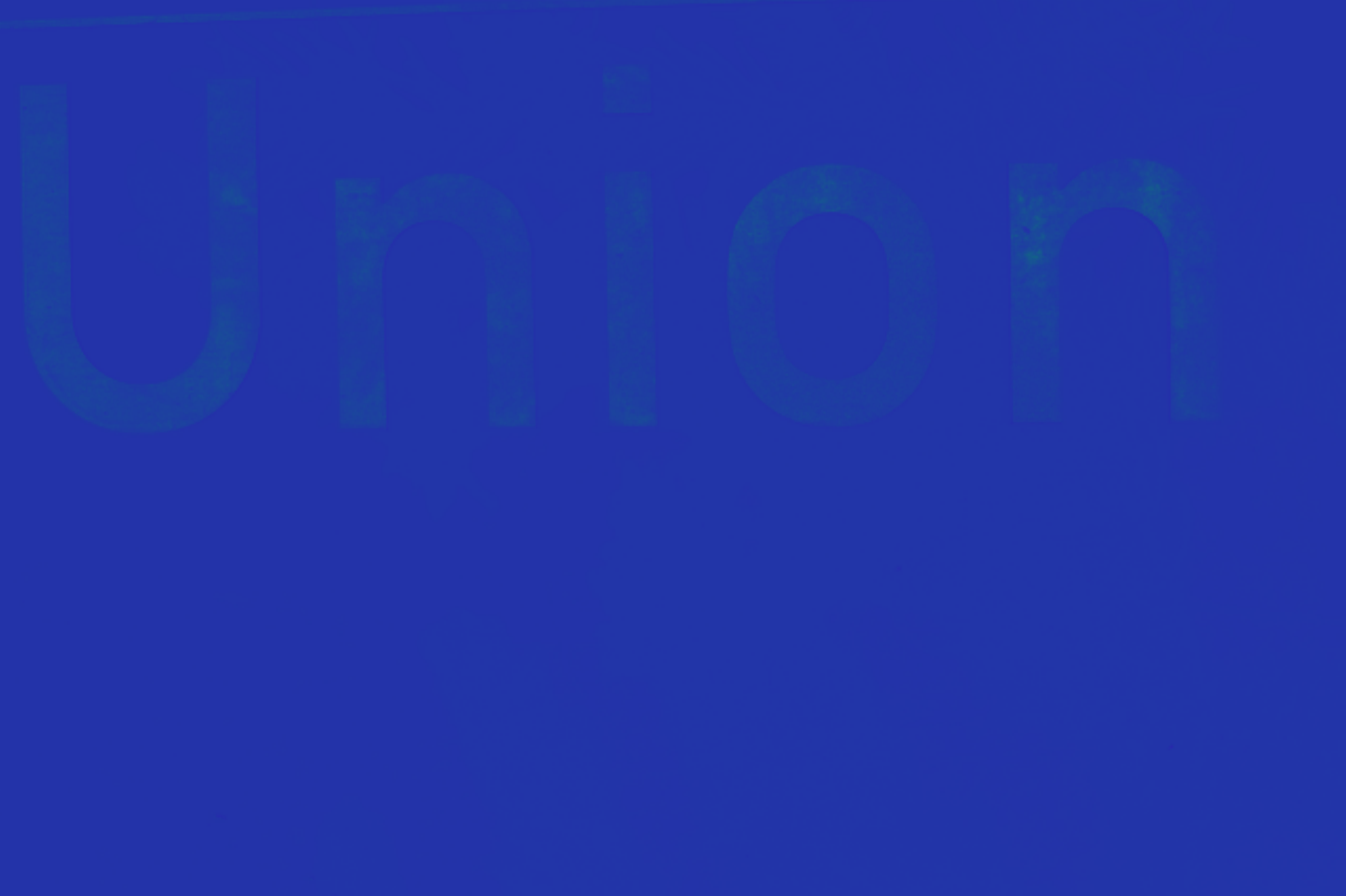}\\ \hline
10&\tbpic{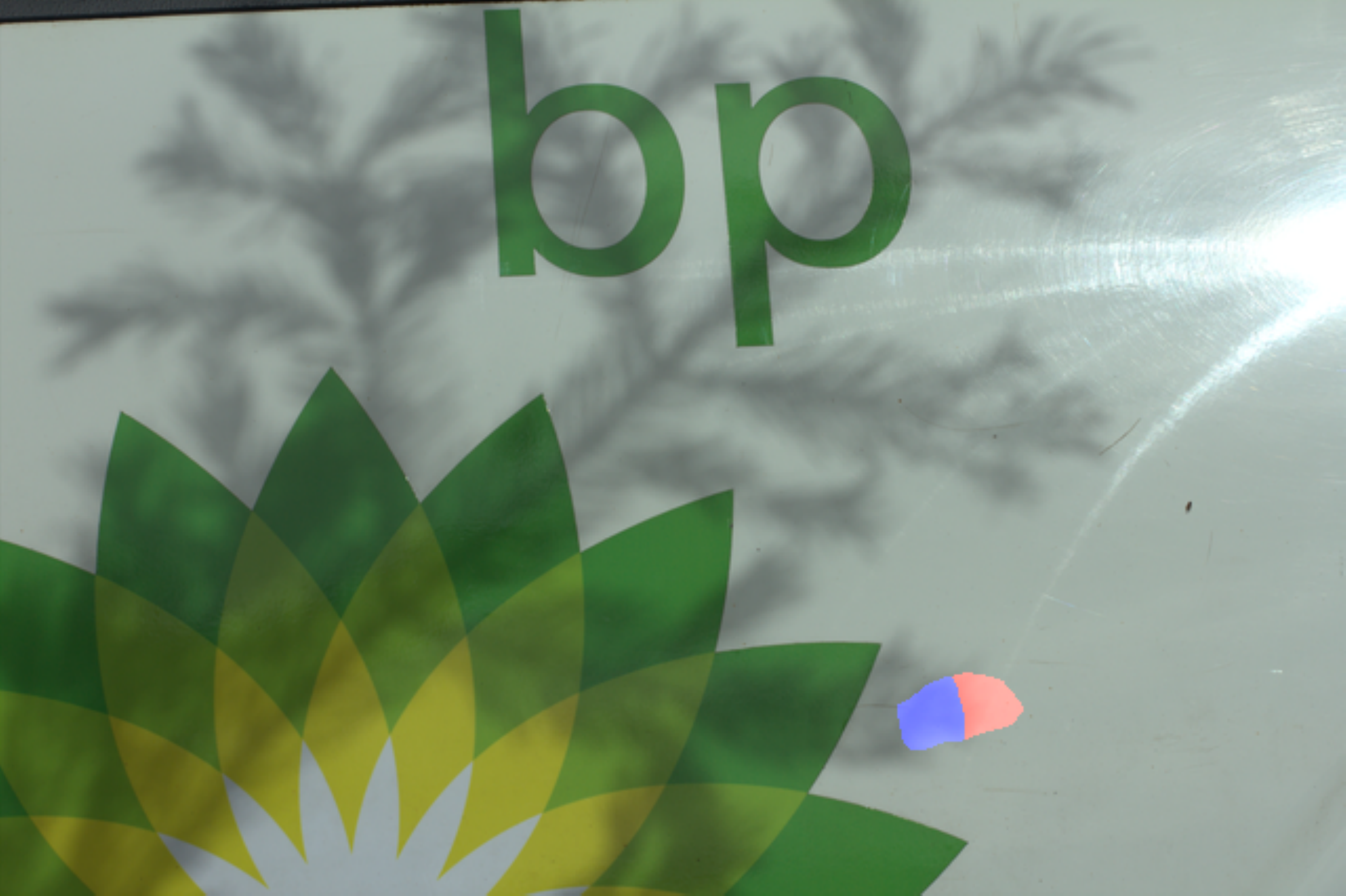}&\tbpic{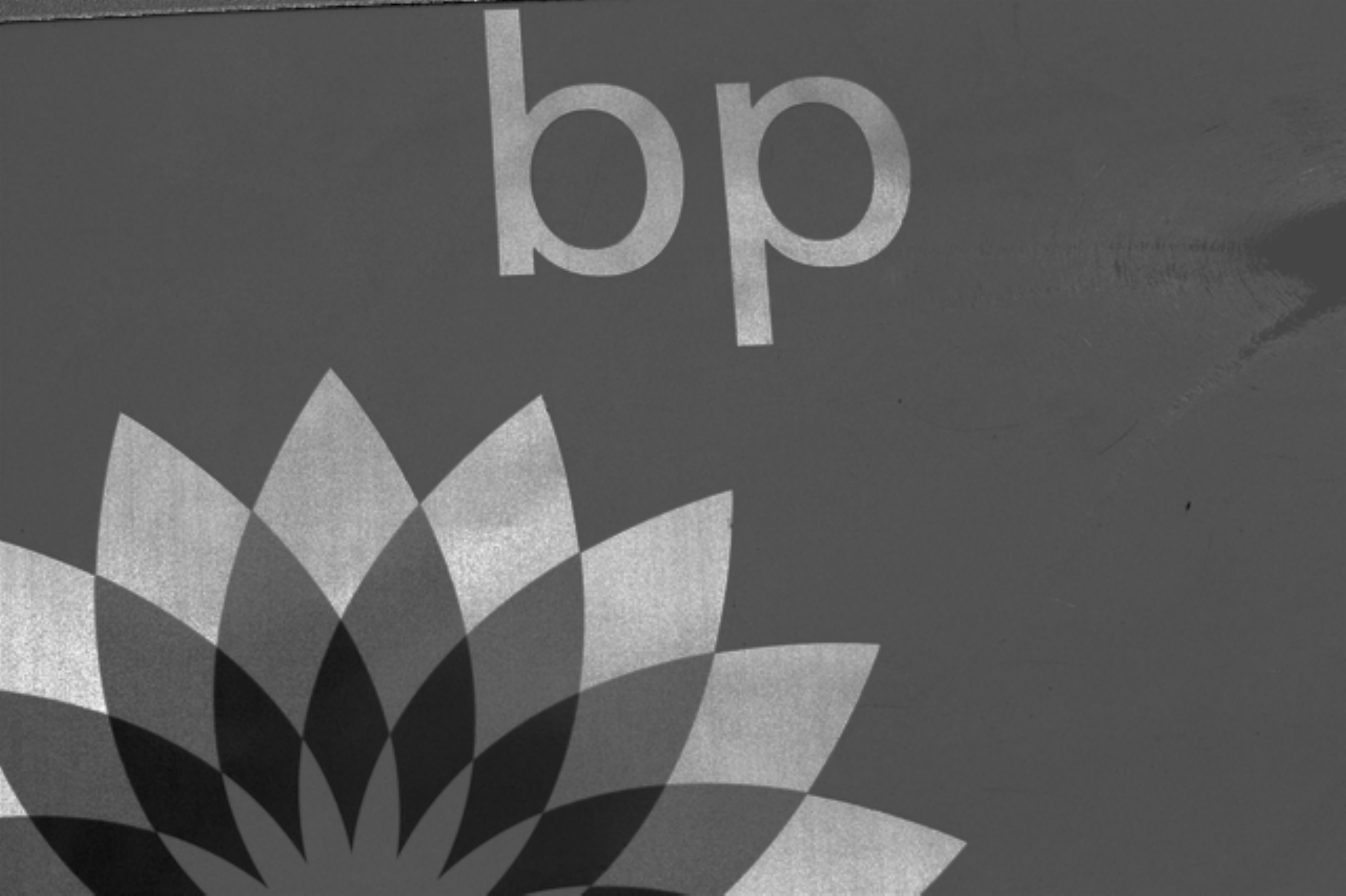}&\tbpic{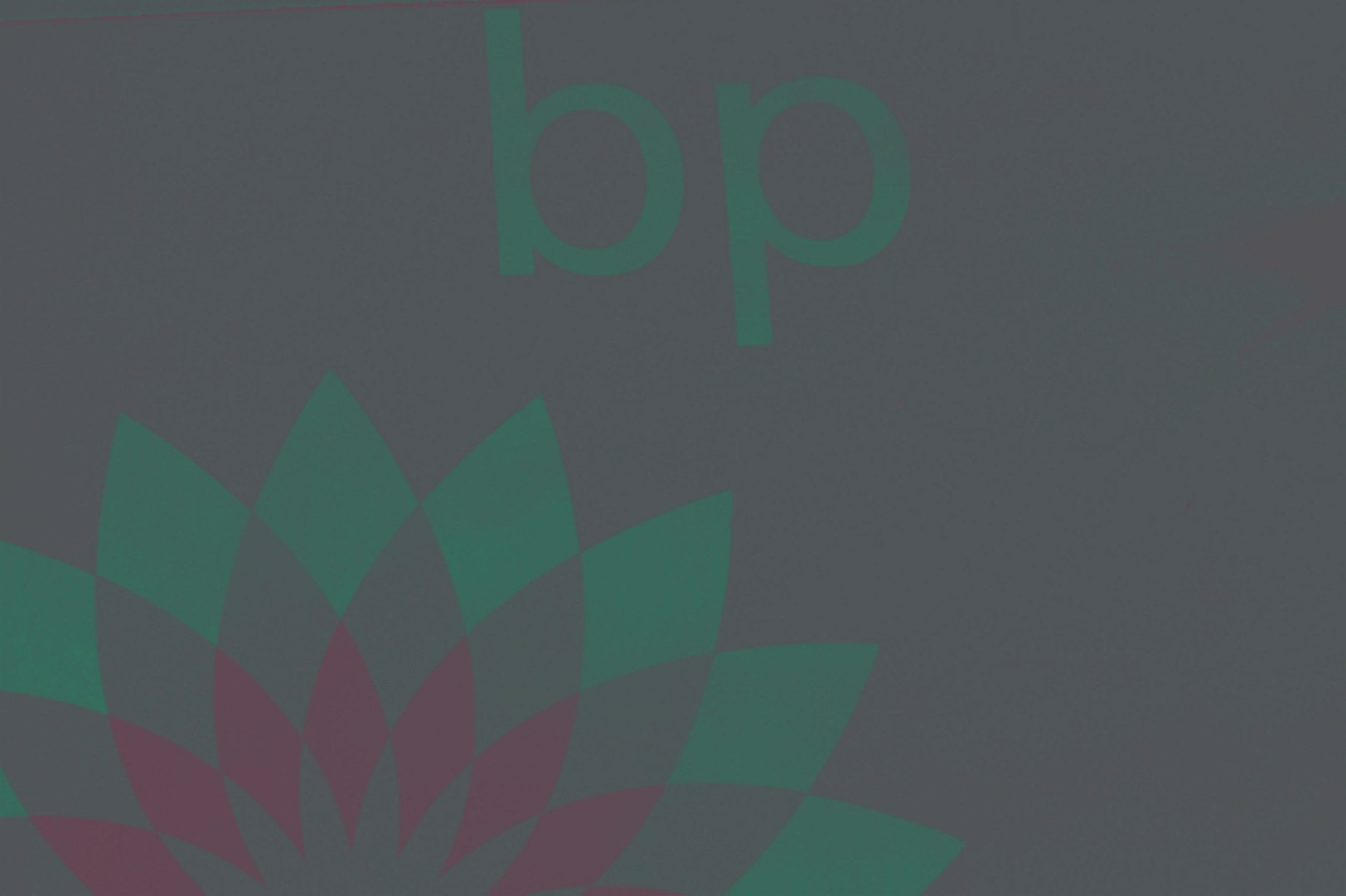}&\tbpic{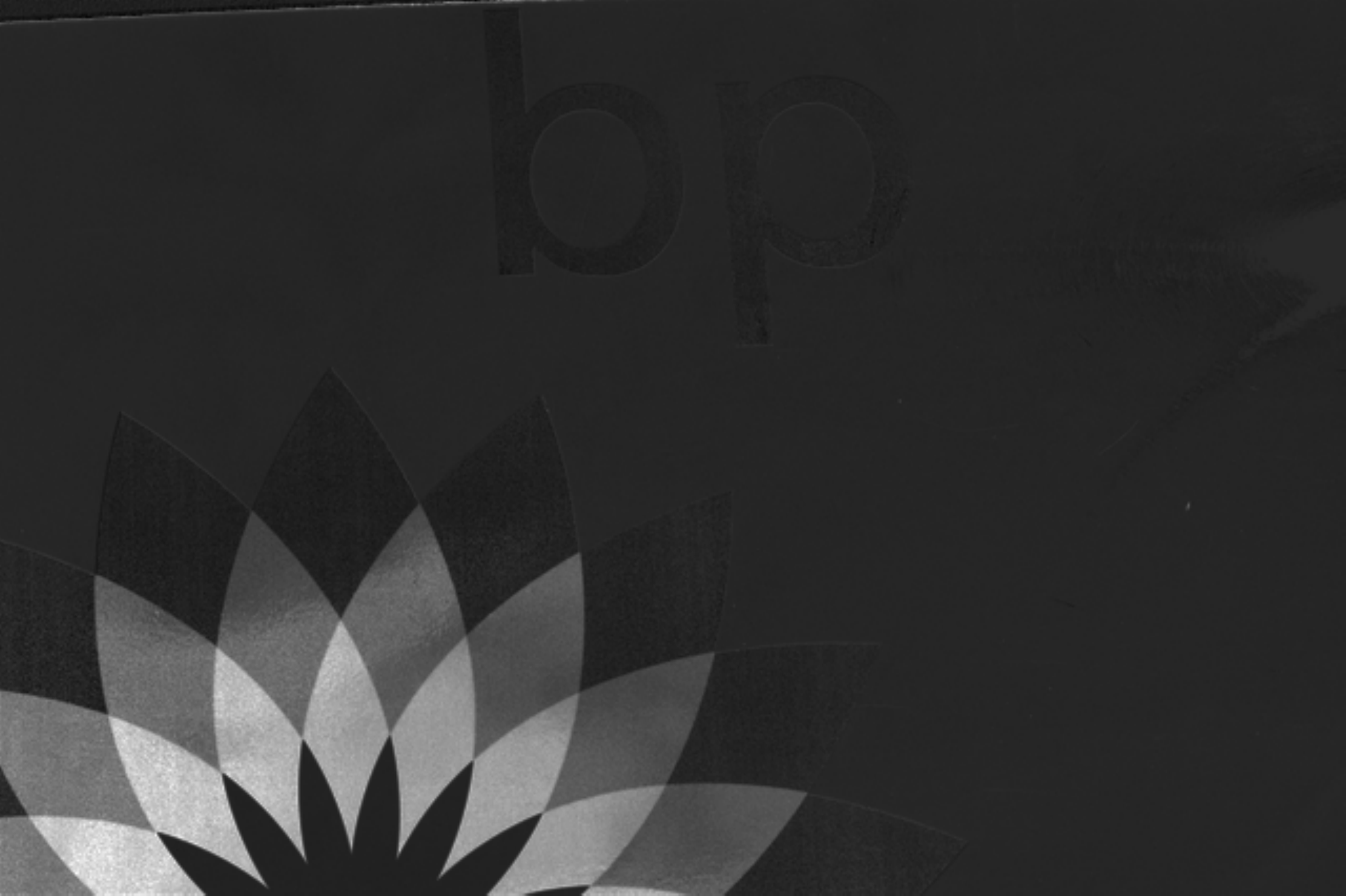}&\tbpic{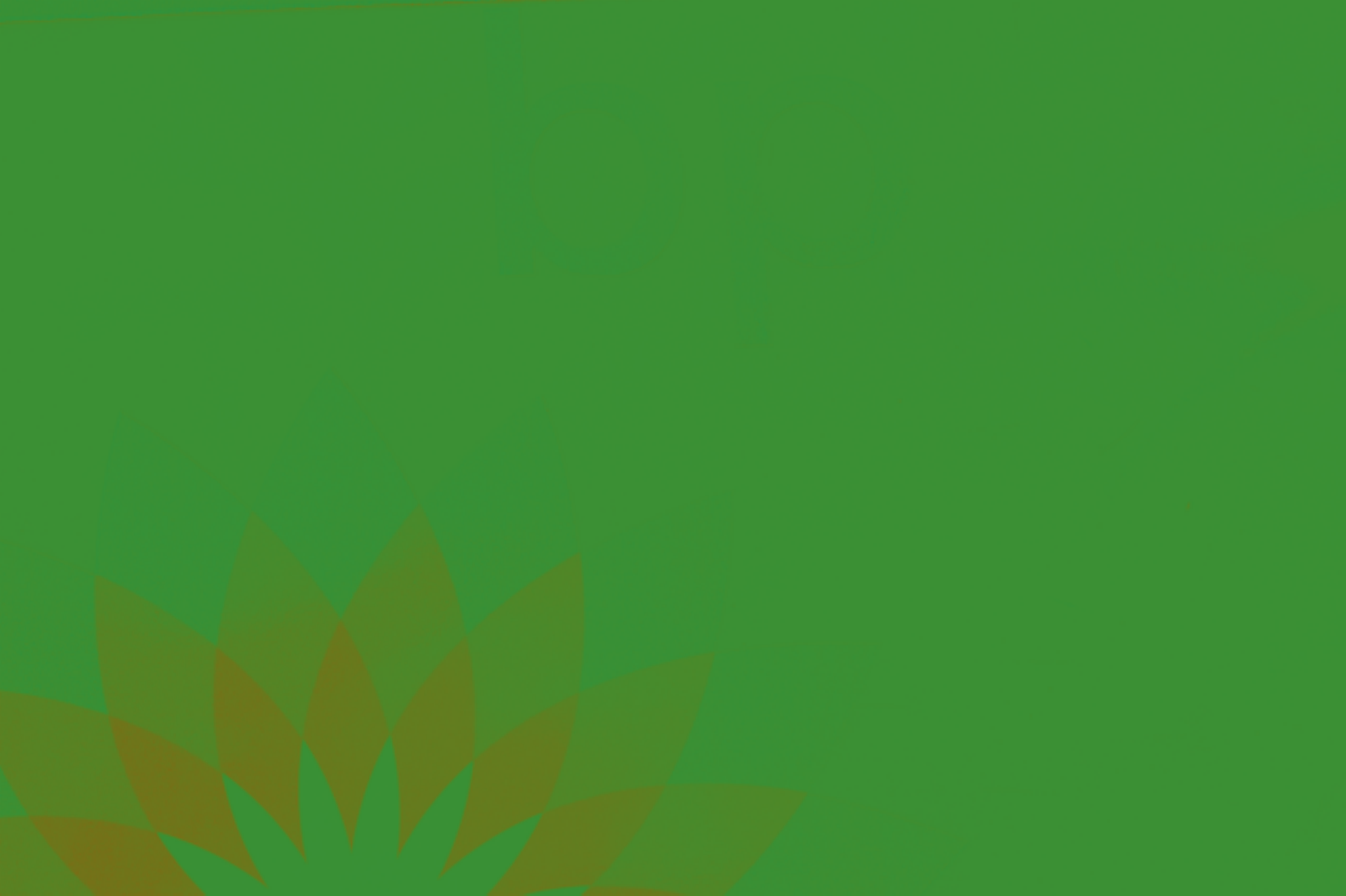}\\ \hline
11&\tbpic{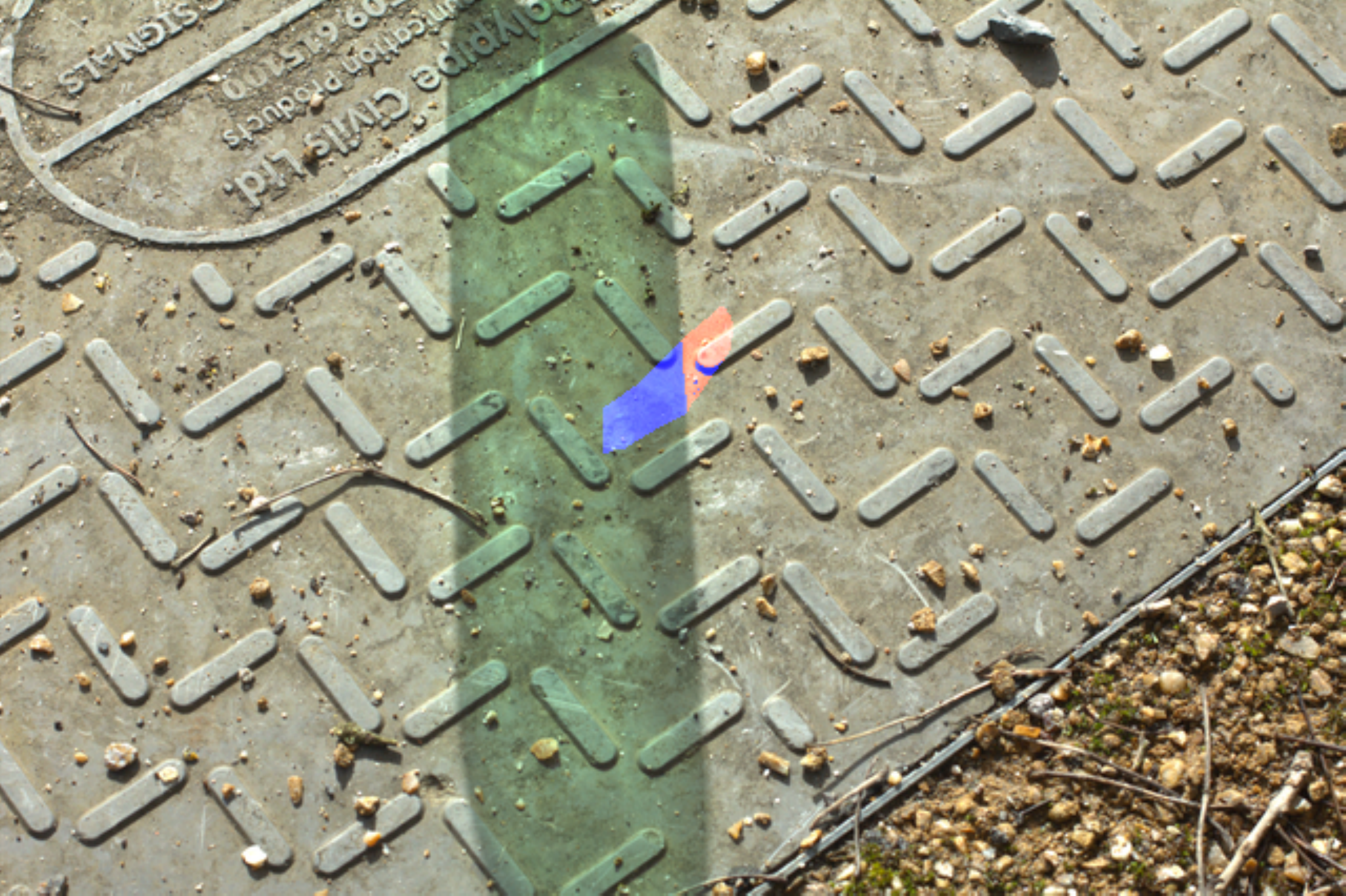}&\tbpic{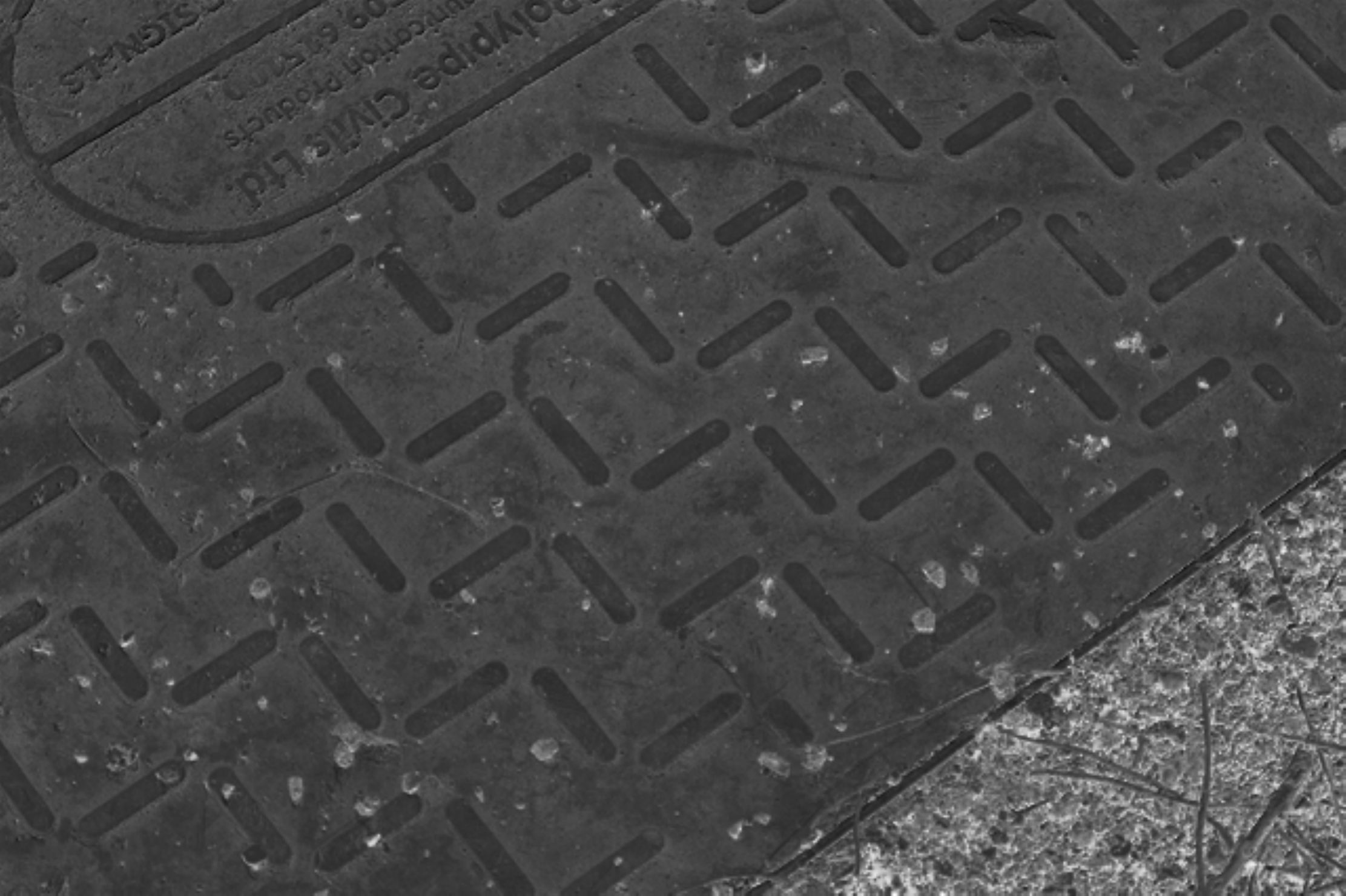}&\tbpic{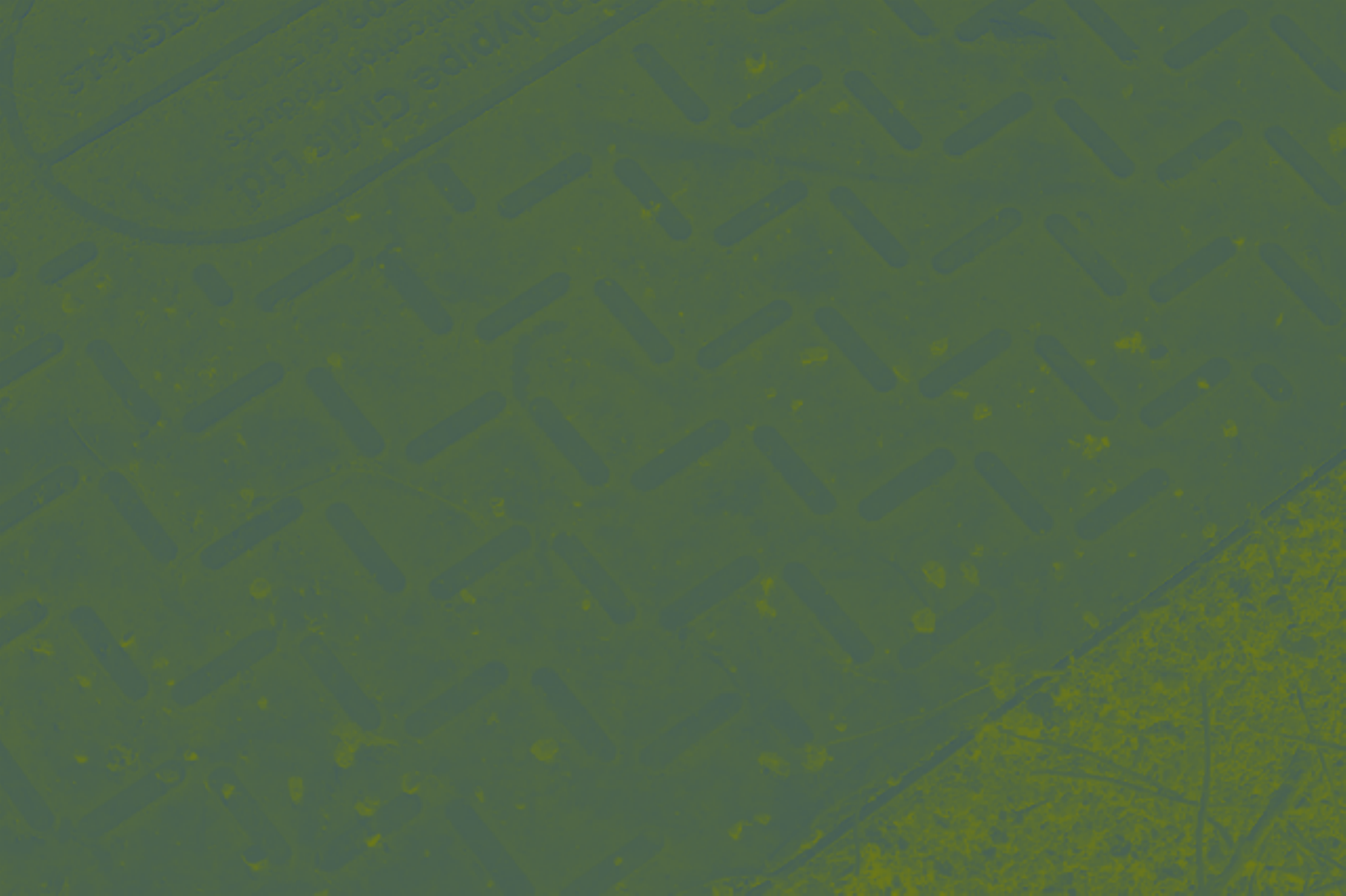}&\tbpic{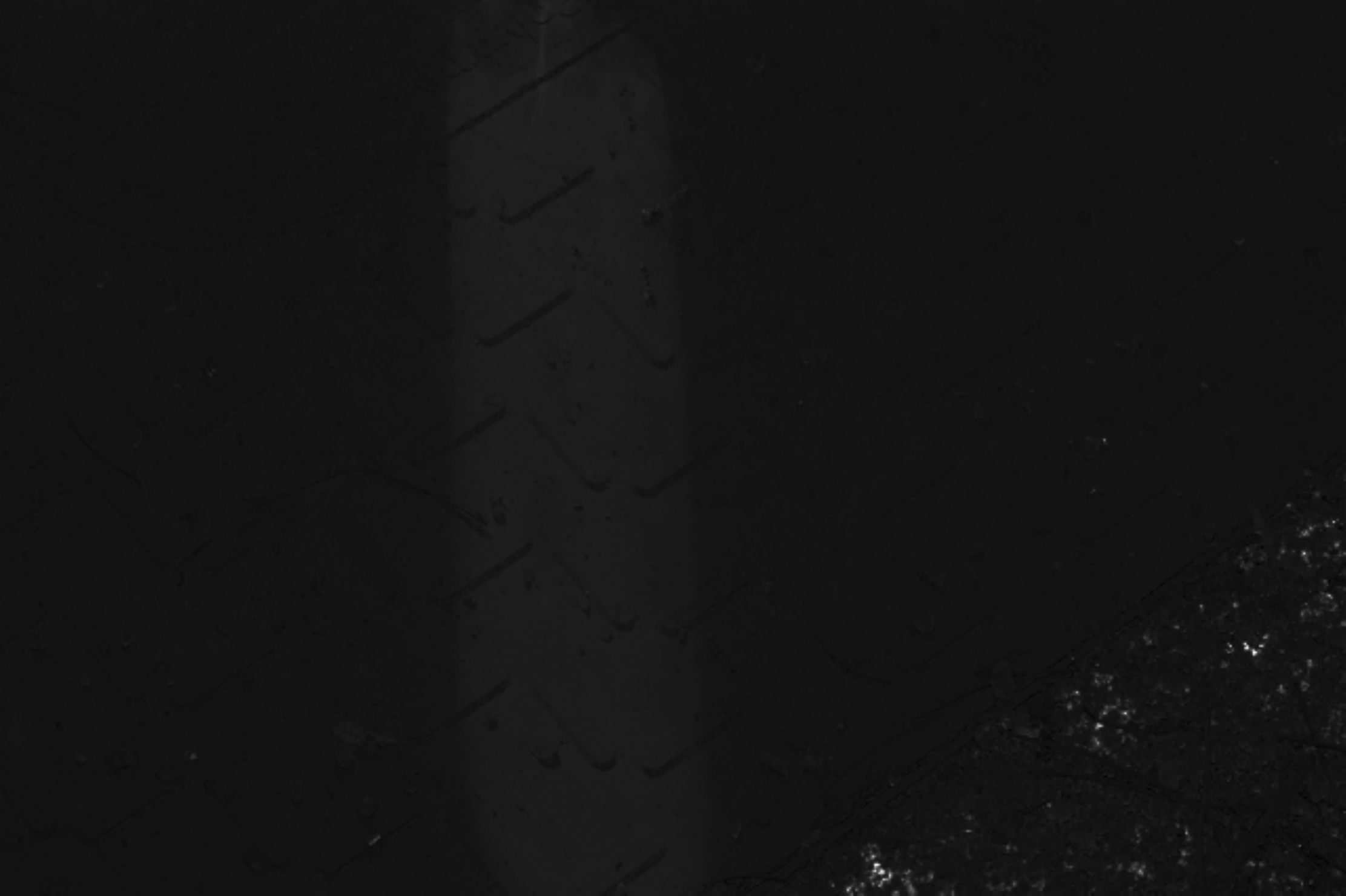}&\tbpic{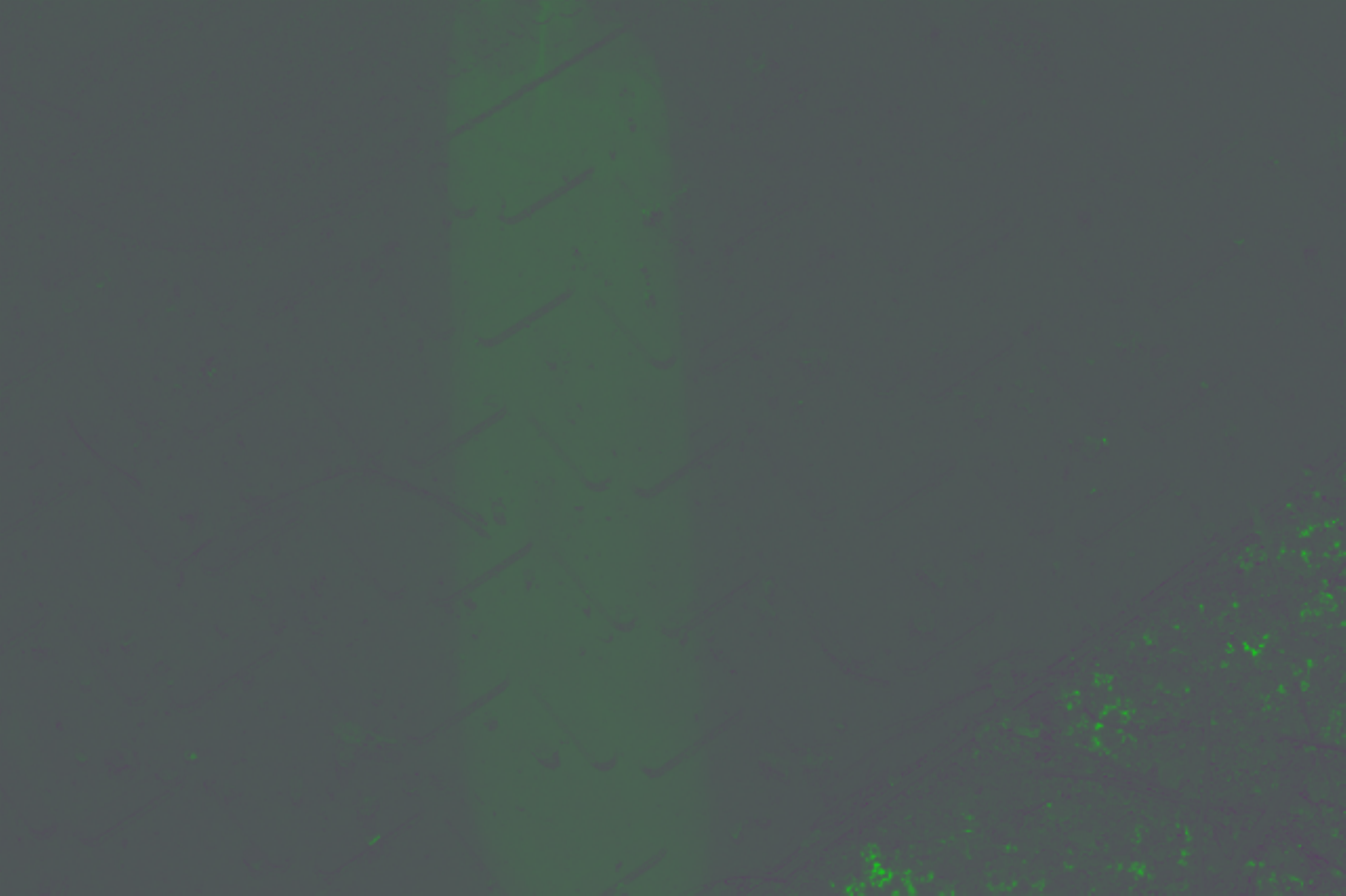}\\ \hline
\end{tabular}
}
    \caption{Results: All the images in this figure are gamma corrected. All the images, except for the bottom four images, are JPEG compressed. The guidance marks are displayed on each original image (please zoom in the electronic version to examine some minor marks). These non-linear operations have severally confound the entropy minimisation approach~\cite{finlayson2004intrinsic}. Images 1-5, 6-7, and 8-11 are sourced from \textbf{Flickr}, \cite{guo2013paired}, and \cite{gong2014interactive} respectively.}
    \label{fig:res}
\end{figure*}

In Fig.~\ref{fig:res}, some typical examples are shown. This set of examples covers a wide range of different illumination conditions, such as broken shadows (row 3, 4, 9, and 10), soft shadows (row 2 and 7), coloured shadows (row 1 and 11) as well as the other trivial shadows. Compared with the automated entropy minimisation approach~\cite{finlayson2004intrinsic}, our method produces more clean (shadow-free) illumination-invariant images for row 1, 2, 6, 8, 9, and 11. Our results in row 3, 4, 7, and 10 also retain more surface texture and chromaticity information. For the images in row 5, the energy minimisation approach produces slightly better results than ours. For a 0.3 mega-pixel image, our single thread MATLAB implementation usually takes 0.3s to process on a 2.4GHz machine. In summary, with the assistance of user input, our approach produces better results for most difficult images at a very fast processing speed.
\subsection{Failure Cases}
\begin{figure}
    \centering
    \includegraphics[width=0.32\linewidth]{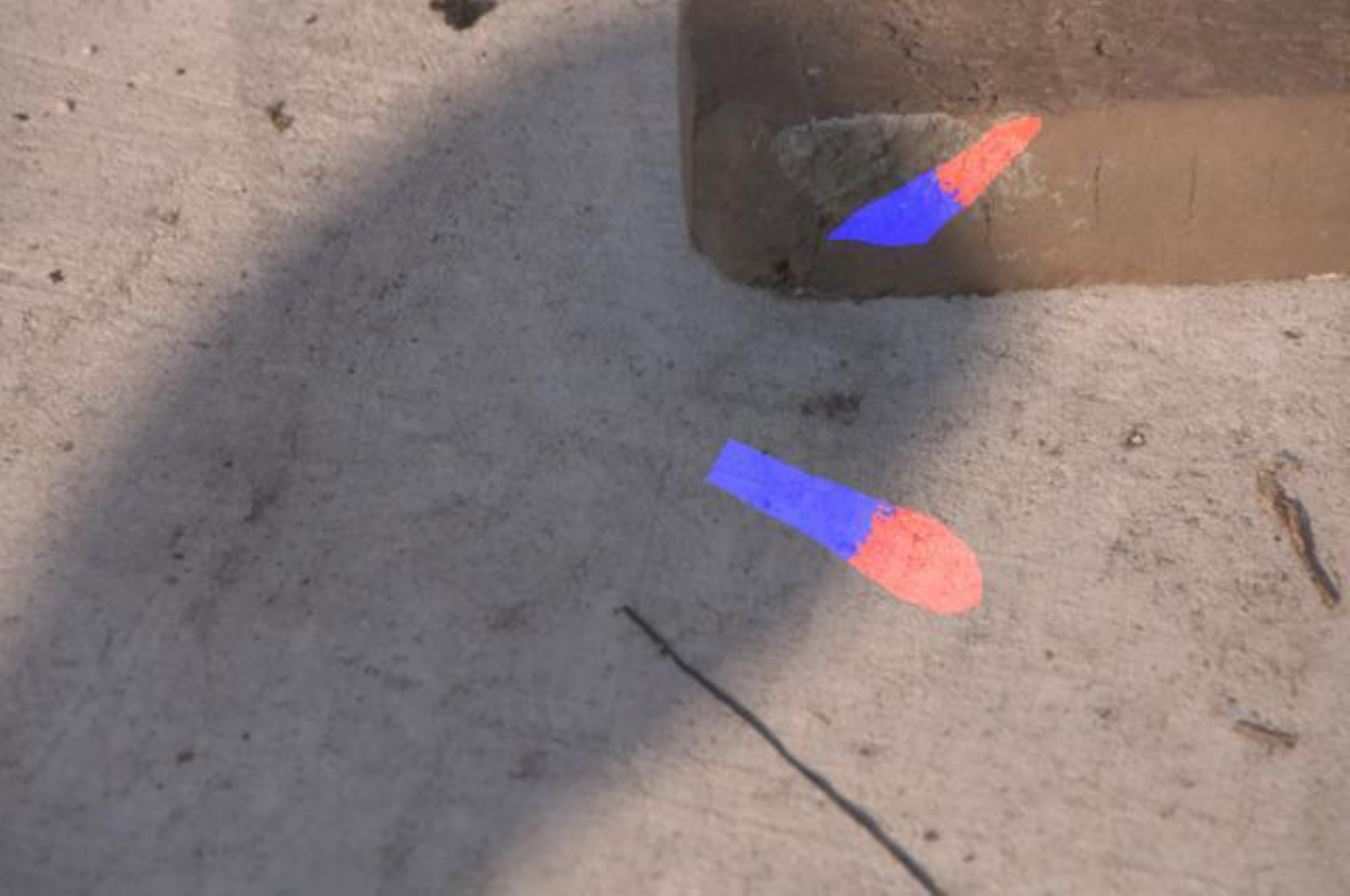}
    \includegraphics[width=0.32\linewidth]{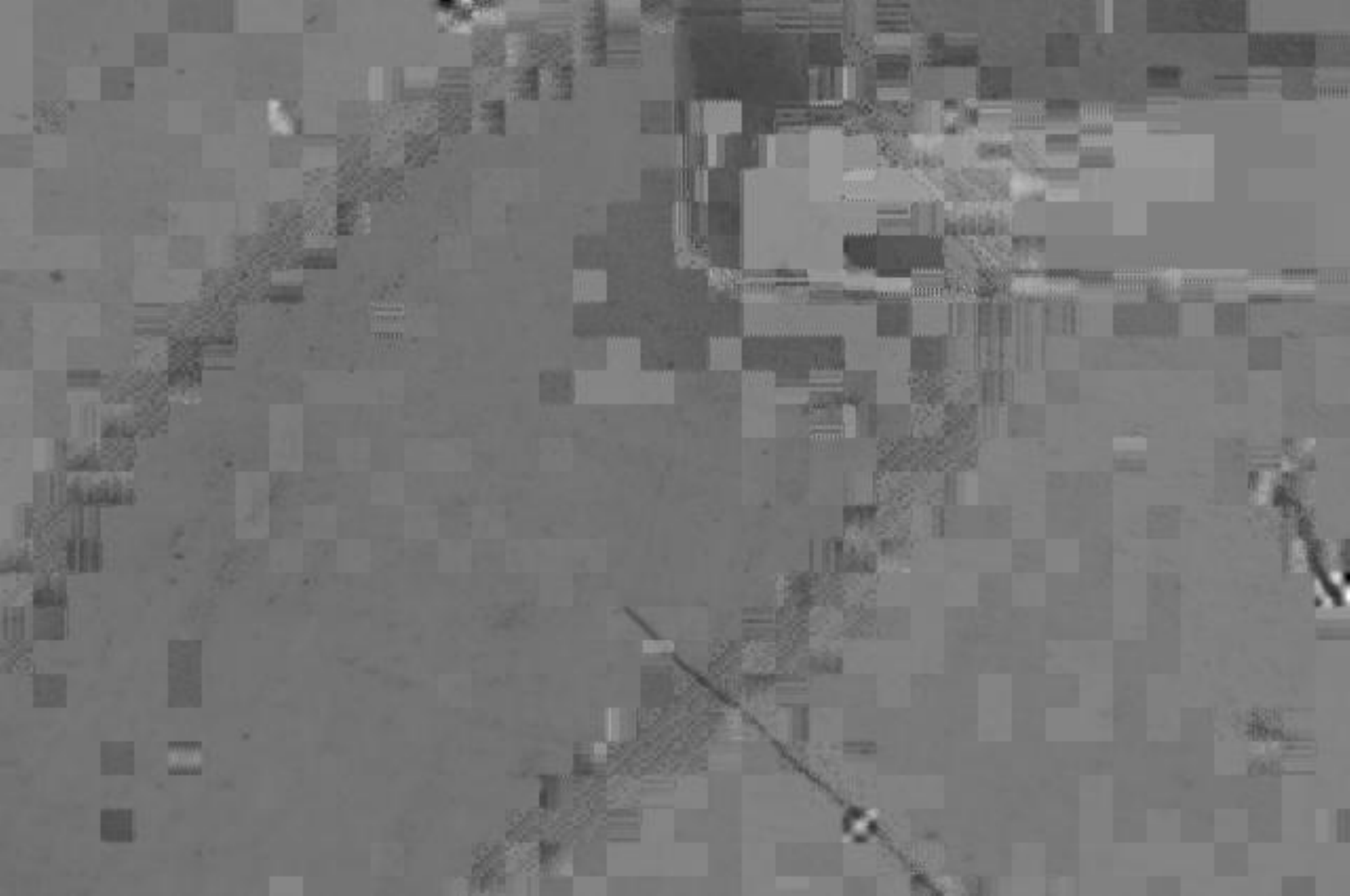}
    \includegraphics[width=0.32\linewidth]{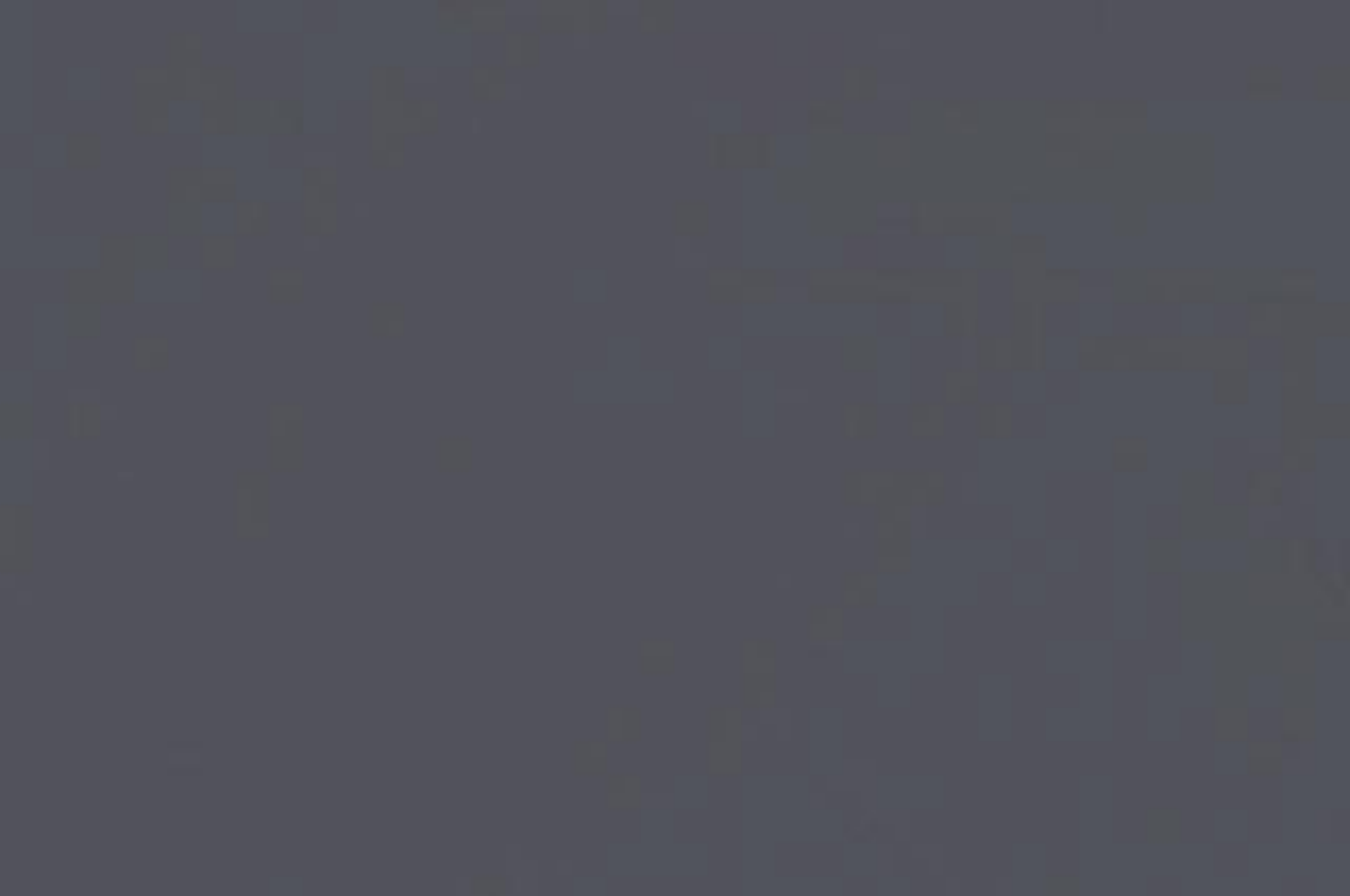}
    \caption{Failure cases (heavily JPEG compressed): left -- original image with user input; middle -- 1D illumination-invariant image; right -- L1 log chromaticity illumination-invariant image.}
    \label{fig:fail}
\end{figure}

\begin{figure}
    \centering
    \includegraphics[width=0.9\linewidth]{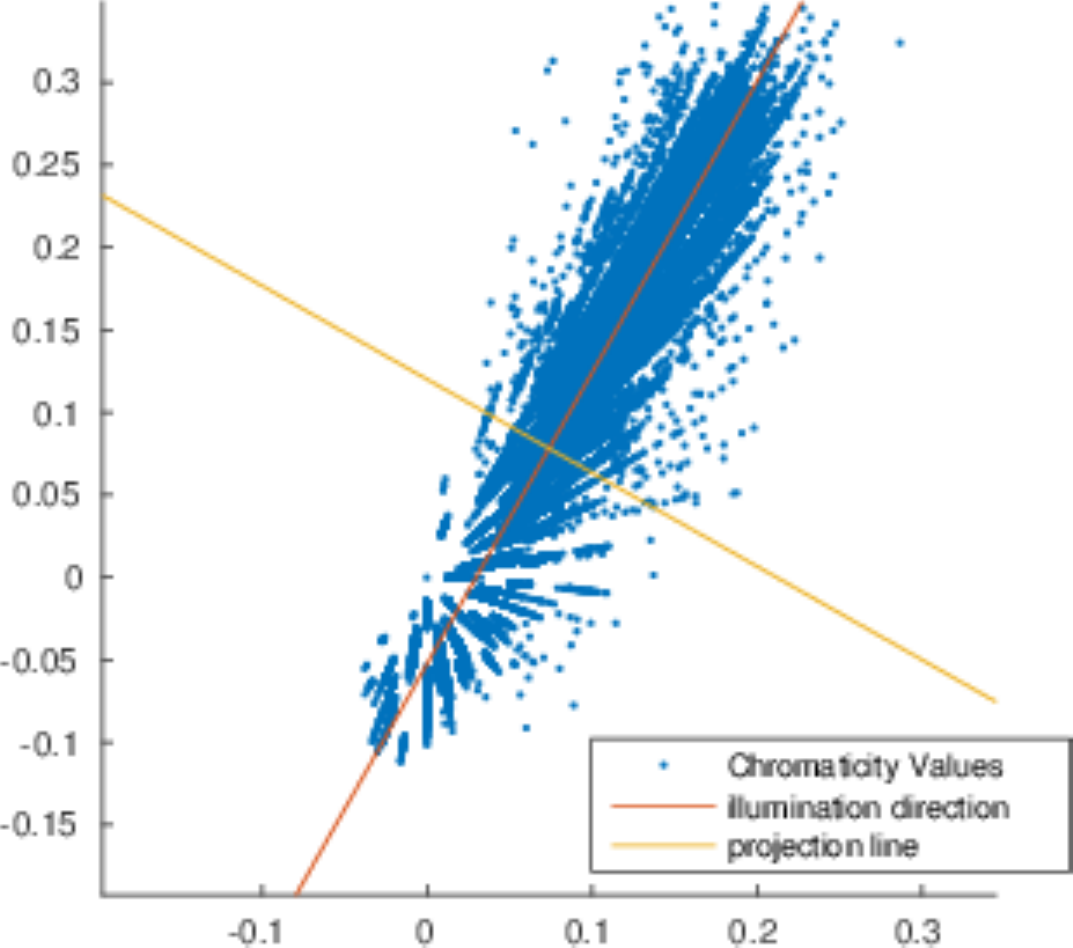}
    \caption{2D log chromaticity of a JPEG compressed image: The lines of chromaticity points are not parallel but spread in different directions. This is due to the non-linear post-processing operations.}
    \label{fig:2D_fail}
\end{figure}

Since we still assume the near linearity of pixel intensities, our approach fails when the image has undergone a heavy non-linear rendering process, e.g., low quality JPEG compression. A failure cases is shown in Fig.~\ref{fig:fail}. In Fig.~\ref{fig:fail}, the shadow and the JPEG blocks are clearly visible in the 1D illumination-invariant image and its L1 log chromaticity image almost contains no information. Fig.~\ref{fig:2D_fail} shows the 2D log chromaticity plot of Fig.~\ref{fig:fail}.

\section{Conclusion}
\label{sec:con}
We have presented a novel approach for interactive illumination-invariant image derivation. Using our system, users are only required to supply a simple stroke defining an area in an image where the illumination change is significant. Compared with other automated methods, this additional hint significantly enhances the robustness of finding the illumination-invariant direction and makes the illumination-invariant image derivation steerable. Future work includes: (1) A more noise-resistant user input analysis; (2) A more adaptive algorithm to deal with the non-linearity of rendered images.




\small

\bibliographystyle{plain}      
\bibliography{cic}   

\end{document}